\documentclass{article}

\usepackage{authblk}
\usepackage[sort&compress,comma,authoryear]{natbib}
\usepackage{setspace}
\usepackage{subcaption}
\usepackage{svg}
\usepackage{booktabs}
\usepackage{balance}
\usepackage{xcolor}
\usepackage{tabularx}
\usepackage{graphicx}
\usepackage{amsmath}
\usepackage{amssymb}
\usepackage{listings}
\usepackage{textcomp}
\usepackage{url}
\usepackage{multirow}
\usepackage{hyperref}
\usepackage{cleveref}
\usepackage{xspace}
\usepackage{wrapfig}
\crefformat{footnote}{#2\footnotemark[#1]#3}
\usepackage{ragged2e} 
\usepackage[subnum]{cases}
\newcolumntype{Y}{>{\RaggedRight\arraybackslash}X} 
\usepackage{algorithm}
\usepackage{bm}
\usepackage[noend]{algpseudocode}
\usepackage{hyperref}
\usepackage[toc,page]{appendix}
\usepackage{float}

\newcolumntype{L}[1]{>{\raggedleft\arraybackslash}p{#1}}
\newcolumntype{R}[1]{>{\raggedright\arraybackslash}p{#1}}
\newcolumntype{K}[1]{>{\centering\arraybackslash}p{#1}}

\graphicspath{ {figures/}}

\newcommand{\degree}{\ensuremath{^{\circ}}\xspace}
\newcommand{\coo}{\ensuremath{\mathrm{CO_2}}\xspace}
\newcommand{\secref}[1]{Section \ref{#1}\xspace}
\newcommand{\figref}[1]{Figure \ref{#1}\xspace}

\usepackage{todonotes}

\newcommand{\alloc}[1]{}

\renewcommand{\vec}[1]{\boldsymbol{#1}}

\title{Heterogeneous robot teams with unified perception and autonomy: How Team CSIRO Data61 tied for the top score at the DARPA Subterranean Challenge }

\author[1]{Navinda Kottege}
\author[1]{Jason Williams}
\author[1]{Brendan Tidd}
\author[1]{Fletcher Talbot}
\author[1]{Ryan Steindl}
\author[1]{Mark Cox}
\author[1]{Dennis Frousheger}
\author[1]{Thomas Hines}
\author[1]{Alex Pitt}
\author[1]{Benjamin Tam}
\author[1]{Brett Wood}
\author[1]{Lauren Hanson}
\author[1]{Katrina Lo Surdo}
\author[1]{Thomas Molnar}
\author[1]{Matt Wildie}
\author[1]{Kazys Stepanas}
\author[1]{Gavin Catt}
\author[1]{Lachlan Tychsen-Smith}
\author[1]{Dean Penfold}
\author[1]{Les Overs}
\author[1]{Milad Ramezani}
\author[1]{Kasra Khosoussi}
\author[2]{Farid Kendoul}
\author[2]{Glenn Wagner}
\author[2]{Duncan Palmer}
\author[2]{Jack Manderson}
\author[2]{Corey Medek}
\author[3]{Matthew O'Brien}
\author[3]{Shengkang Chen}
\author[3]{Ronald C. Arkin}

\affil[1]{CSIRO, Australia (At the time of this work being done)}
\affil[2]{Emesent, Milton, Queensland, Australia}
\affil[3]{Georgia Tech, Atlanta, Georgia, USA}
\date{}

\begin{document}

\maketitle

\begin{abstract}
The DARPA Subterranean Challenge was designed for competitors to develop and deploy teams of autonomous robots to explore difficult unknown underground environments. Categorised in to human-made tunnels, underground urban infrastructure and natural caves, each of these subdomains had many challenging elements for robot perception, locomotion, navigation and autonomy. These included degraded wireless communication, poor visibility due to smoke, narrow passages and doorways, clutter, uneven ground, slippery and loose terrain, stairs, ledges, overhangs, dripping water, and dynamic obstacles that move to block paths among others. In the Final Event of this challenge held in September 2021, the course consisted of all three subdomains. The task was for the robot team to perform a scavenger hunt for a number of pre-defined artefacts within a limited time frame. Only one human supervisor was allowed to communicate with the robots once they were in the course. Points were scored when accurate detections and their locations were communicated back to the scoring server. A total of 8 teams competed in the finals held at the Mega Cavern in Louisville, KY, USA. This article describes the systems deployed by Team CSIRO Data61 that tied for the top score and won second place at the event.
\end{abstract}

\section{Introduction}
\label{sec:introduction}

There have been significant advancements in field robotics in the past decade in terms of maturity of commercially available platforms, advanced sensor technology, navigation capability, power systems and compute systems. However, fully autonomous deployment of robots in real world field applications are still not commonplace. This is especially so in unknown, dangerous and difficult environments. There seem to be some technology gaps that prevent widespread use of robotic systems in such environments--an application area that can have significant benefits in removing humans from dull, dirty and dangerous environments.

Subterranean environments such as mining tunnels, underground urban infrastructure and natural caves are especially challenging for robots due to lack of GPS for localisation, degraded wireless communication, poor perception due to obscurants, locomotion and navigation difficulties due to narrow passages and doorways, clutter, uneven ground, slippery and loose terrain, stairs, ledges, overhangs, dripping water, and dynamic obstacles that can move to block paths etc. The DARPA Subterranean Challenge (SubT Challenge) was motivated by this gap in technology that was preventing effective deployment of autonomous robots in to this type of environments. As described in \cite{Orekhov_2022}, the primary scenario centred around providing advanced situational awareness to rescuers following a collapsed mine, earthquake or personnel lost or injured in a cave. The challenge was designed with extensive input from and in collaboration with first responders. The aims included spurring innovation, building communities of practice, setting new benchmarks for state-of-the-art, and creating societal impact. Point scoring in competition events was achieved by detecting, identifying and locating an artefact to an accuracy of within 5~m. Artefact classes included survivors (thermal mannequins), backpacks, ropes, helmets, fire extinguishers, power drills, vents, \coo concentrations, cell phones and LED-illuminated cubes. To make the scenarios as realistic as possible, the event courses included rough terrain, steep inclines, narrow openings, stairs, water, obscurants such as dust and smoke, and dynamic obstacles (e.g., representing further cave-ins during the mission). Due to the underground environment, GPS is unavailable and communications are severely restricted.

\begin{figure}[ht]
    \centering
    \includegraphics[width=120mm]{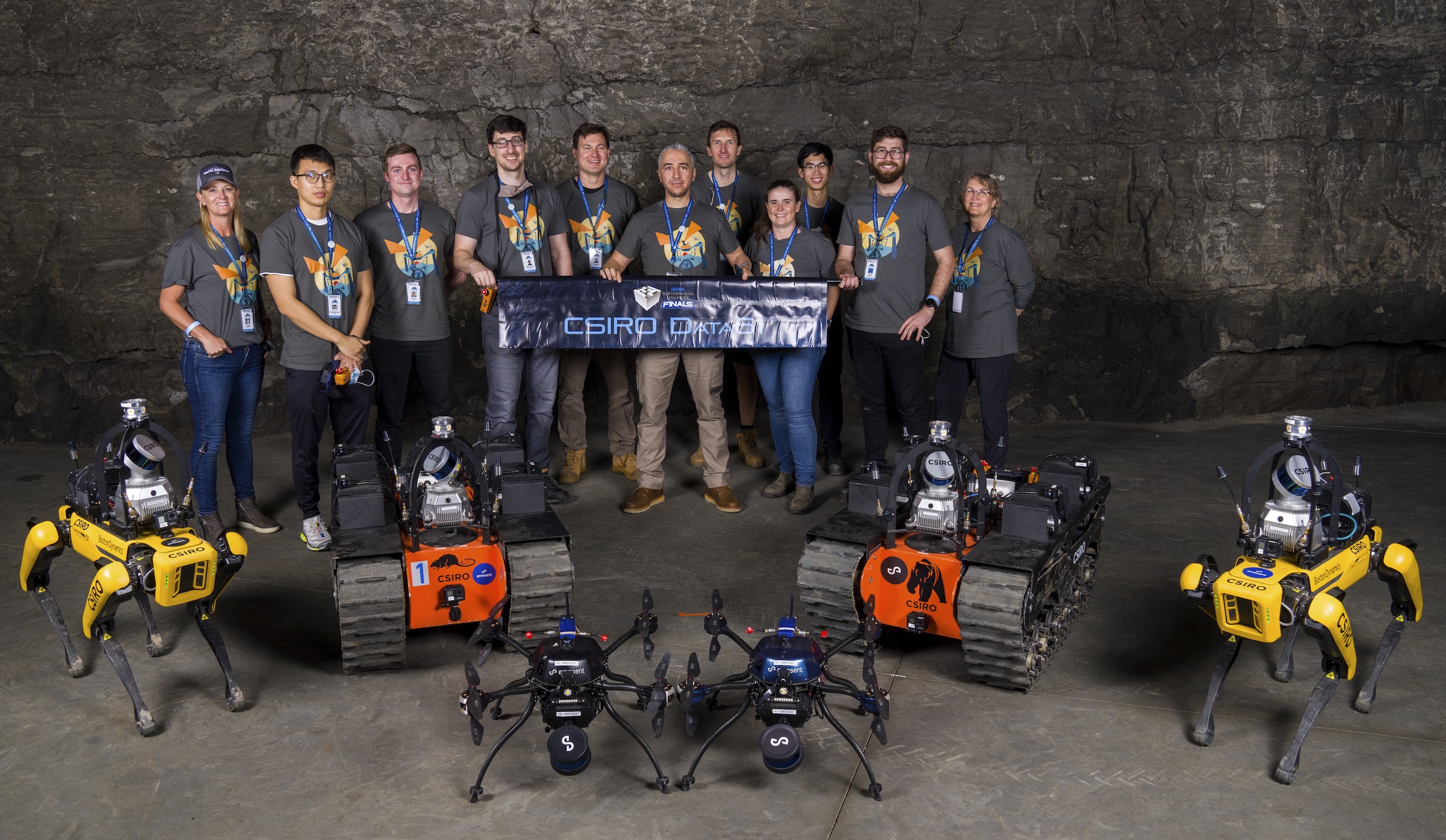}
    \caption{Team CSIRO Data61 members at the DARPA Subterranean Challenge Final Event along with the robot fleet.}
    \label{fig:CSIRO_Data61_team}
\end{figure}

Challenge events commenced with the SubT Integration Exercise (STIX) in the Edgar Experimental Mine, Idaho Springs, CO, USA in April 2019. The first competitive event was the Tunnel Circuit at the NIOSH Safety Research Coal Mine and Experimental Mine in Pittsburgh, PA, USA in August 2019. The Urban Circuit event was held in February 2020 at Satsop Business Park, Elma, WA, USA, in an un-commissioned nuclear power plant. The COVID-19 pandemic led to the cancellation of the Cave Circuit event that was scheduled to take place in August 2020. The teams were encouraged to conducted their own testing in local cave environments that they could access. Each year of the challenge represented a Phase, with Phase I including the Tunnel Circuit, Phase II the Urban and Cave Circuits, and Phase III concluding with the Final Event. After successfully competing in circuit events in Phase I and II of the DARPA SubT Challenge \citep{hudson2021heterogeneous}, Team CSIRO Data61 was selected as one of the eight finalist to compete in the Final Event that was held in September 2021 at the Louisville Mega Cavern, KY, USA (\figref{fig:CSIRO_Data61_team}). After winning the preliminary round at the Final Event, Team CSIRO Data61 tied for the top score of 23 points with Team CEREBRUS in the final prize run. After tie-breaker rules were invoked, Team CSIRO Data61 won the US\$1\,Million second place prize, with only 1 minute,
1\,cm, or 1 extra artefact report away from first place~\citep{chung_into_2023}. 
 
This paper summarises the heterogeneous robot system deployed by Team CSIRO Data61, utilising unified multi-agent mapping and autonomy. We also describe how the team of robots overcame various challenges in the Final Event course, and results and lessons learned from the program. The overall system of systems demonstrated remarkable resilience in the harsh environment even in the face of attrition of individual agents. We will focus on the advancements and changes implemented on our systems for Phase III of the competition, in comparison to the work presented in our prior work \cite{hudson2021heterogeneous} representing Phases I and II. 

\subsection{Related Work}
\label{sec:related_work}
System reviews from the various SubT teams from Phase I and II each contain detailed literature reviews, and can be found in \cite{agha_nebula_2021, tranzatto_cerberus_autonomous_2022, scherer_resilient_2022, ohradzansky_multi-agent_2021, roucek_system_2021, lu_heterogeneous_2022, isaacs_teleoperation_2022, hudson2021heterogeneous}. Here, we describe the systems employed by SubT teams as described in these papers, and subsequent examinations of particular system components. 

\cite{tranzatto_cerberus_2022} describes the system employed by Team CERBERUS (University of Nevada, Reno, ETH Z\"{u}rich, NTNU, University of California Berkeley, Oxford Robotics Institute, Flyability and Sierra Nevada Corporation) that won first place in the final prize round, with \cite{tranzatto_cerberus_autonomous_2022} outlining the system from the Tunnel and Urban Circuits. The robot roster for Team CERBERUS consisted of aerial scout Uncrewed Aerial Vehicles (UAVs, including small, medium and large platforms), a tethered wheeled ground vehicle to extend communications, and two variations of the ANYmal C quadruped: Carrier for deploying communication-extender modules, and Explorer for proceeding deep into the environment. Graph-based path planning was used with a bifurcated architecture for efficient local and global exploration \citep{dang_graph_2020}. Localisation and mapping was performed with complementary multimodal sensor fusion (CompSLAM) described by \cite{khattak_complementary_2020}. \cite{miki_learning_2022} details the learning-based perceptive locomotion utilised by the quadrupeds that balances the information from multi-modal perception sources to traverse complex terrain (including stairs) in the presence of sensor noise. 

Team CoSTAR (NASA Jet Propulsion Laboratory, California Institute of Technology, MIT, KAIST, Lulea University of Technology) used a combination of Boston Dynamics Spot robots, UAVs and wheeled robots; the system that was deployed in the Tunnel and Urban Circuits is described in \cite{agha_nebula_2021}. The focus of their development was NeBula (networked belief aware perceptual autonomy), which aimed to address challenging, degraded-sensing environments through a combination of sensor fusion, and uncertainty-aware planning. By incorporating map and location uncertainty into the planning problem, risk-aware plans were constructed that account for difficult phenomena such as dust and smoke. Planning under uncertainty is a very challenging problem; this was approached through a hierarchical system described in \cite{kim_plgrim_2021}. The primary modality for SLAM was lidar/inertial, but alternative solutions were maintained based on visual, thermal, radar and wheel odometry, fused based on the estimated confidence \citep{santamaria_hero_2022}, and used as hints for the lidar/inertial solution \citep{paleri_locus_2021}. Description on the integration of NeBula onto the Spot legged platforms to enable autonomous legged exploration is provided in \cite{bouman_autonomous_2020}. A semantic object mapping pipeline provides the operator with enhanced object detections as described in \cite{lei_early_recall_2022}.

Team CTU-CRAS-NORLAB (Czech Technological University, Université Laval) solution for the Tunnel and Urban Circuits utilised wheeled Husky UGVs, tracked Absolem UGVs, PhantomX hexapods and DJI quadrotors \citep{roucek_system_2021}. Platforms used various models of lidar, visual cameras and depth cameras: The Husky used a multi-beam lidar and six cameras; the Absolem used a rotating line lidar and a single omnicamera, the PhantomX used RGBD cameras, and the quadrotor used multi-beam lidar and cameras. Multi-band communications provides high-rate WiFi connection at short ranges (5\,GHz and 2.3\,GHz), and low bandwidth control information at longer ranges (900\,MHz). Object detection used YOLOv3 trained on 20,000 images, alongside \coo and WiFi detection capability. Wheeled and tracked robots used an extended Kalman filter (EKF) to provide an initialisation for an iterative closest point (ICP)-based odometry method, based on lidar odometry and mapping (LOAM), with no loop closure enabled (since the required accuracy could be met without it). Exploration utilised frontier methods~\citep{Bayer2019OnAS}, with coordination provided by the human supervisor.

Team Explorer's (Carnegie Mellon University, Oregon State University) system from the Tunnel and Urban Circuit events is described in \cite{scherer_resilient_2022}. Custom wheeled robots enabled coverage of difficult terrain, and marsupial launch of a UAV. Communications were built around Rajant DX2's \footnote{\url{https://rajant.com/}}, and ground robots carried a total of 24 nodes, which were dropped autonomously based on line of sight and signal strength. Data was selectively shared between robots based on a ledger system. Rather than using a tightly coupled approach, super odometry \citep{zhao_super_2021} uses a mixed (loosely and tightly coupled fusion) scheme where visual and lidar inertial odometry (VIO and LIO) estimates are fused with with IMU measurements asynchronously to estimate robot trajectory at a fast (200\,Hz) rate. Object detection used CNN-based detection pipelines trained on RGB and thermal images, trained using data sets augmented with synthetic data. WiFi and gas localisation relied on human interpretation of the noisy signal strengths. The reference coordinate system (i.e., the ``gate'') was localised using a Total Station-based calibration. Exploration utilised a hierarchical approach \citep{cao_tare_2021}, where a global planner maintains a course tour, and a local planner maintains a detailed path within a local region. Exploration was driven by camera coverage of observed surfaces rather than mapping of 3D space. 

Team MARBLE's (University of Colorado Boulder, University of Colorado Denver, Scientific Systems Company, University of California Santa Cruz) solution for the Tunnel, Urban and Cave Circuit (held virtually) events utilised map and goal-point sharing among agents \citep{riley_assessment_2021}, a metric-topological graph-based planner and a continuous frontier-based planner \citep{ohradzansky_multi-agent_2021}. In this system, the base station did not act as a central agent, but instead could relay information as any other agent. Additionally, the base station merged artefact reports from all robots, providing the human supervisor with a single submission for detections with a similar position and type. The robot fleet consisted of wheeled Husky and tracked Superdroid UGVs, and Lumenier UAVs. Ground platforms used Ouster multi-beam lidars and RGBD cameras with a high-power GPU workstation providing computation, whereas the UAVs used RGBD and ToF cameras. Mapping was performed using Google Cartographer, and map sharing was achieved by extending Octomap to permit transmission of map differences. Communications were built on custom beacons using a custom transport layer solution named \textit{udp\_mesh}.

Team Coordinated Robotics used a teleoperation strategy for the Urban Circuit event (\cite{isaacs_teleoperation_2022}). Due to time constraints in preparing for the event (four months), the team made the decision to focus on integrating multiple platforms with minimal autonomy over a single platform with semi-autonomy. 
The SLAM algorithm LeGO-LOAM (\cite{shan_lego_2018}) was used, with its output fused with IMU data from two Intel Realsense D435i using the ROS Extended Kalman Filter package `robot\_localization' (\cite{moore_generalized_2016}) for estimating the robot pose and artefact position.
As the robots are teleoperated via a video feed, only one robot could be controlled at a time. This led to the team's strategy to use the robots as communication nodes, with a robot teleoperated into the course until the communication bandwidth is unable to sustain the video feed. The robot was then driven back into full communications range using the map data. The next robot was then driven deeper into the course, past the previous robot resulting in a wireless communication backbone and data flow to the operator.  

Team NCTU (National Chiao Tung University) used a heterogeneous team of ground robots (Husky) and blimps to navigate the complex environments of the Tunnel and Urban Circuit events. An overview of the approach is provided in \cite{lu_heterogeneous_2022}. 
Navigation was achieved through deep reinforcement learning using a cross-modal contrastive learning of representations (CM-CLR) method, where mmWave radar and lidar data were used for training \citep{huang_cross-modal_2021}. Through this approach, the ground robots were able to navigate in smoke-filled environments using only mmWave radar and XBee for communications. The unique aspect of the team's approach was the use of lighter than air platforms. The motivation for a blimp were twofold, to enable collision-tolerant navigation and for long term autonomous flight \citep{huang_duckiefloat_2019}. With lift achieved by helium, the motors are only required for propulsion, enabling longer flight times. Although the blimp could collide and bounce off the environment, due to the required size of the blimp to generate adequate lift, it had difficulties passing through narrow passageways and was susceptible to airflow within the tunnel. The team attempted to learn a navigation policy for the blimp, but were unable to learn a robust policy due to the inability to model the complex blimp dynamics in the Gazebo simulator. 

In-depth analysis of the results from the SubT Challenge Finals event, along with insights, lessons learned and future work recommendations were provided in \cite{chung_into_2023}. The performance of the different teams were compared using the competition's singular scoring objective (artefacts detected) as well as describing hypothetical scenarios where the artefact error threshold values (set at 5\,m from ground truth for the competition) was adjusted. Alternate relevant evaluation metrics, such as lowest map deviation, greatest map coverage, largest map (points) and fastest successful report, were also introduced to compare the teams' solutions. Additionally, technical and operational insights based on the competition results and evolution of technology during the Challenge period were provided. The most successful solutions had a heterogeneous team (different platform mobility and function) of robots built upon reliable commercial-off-the-shelf (COTS) products. This allowed the solution to adapt to different scenarios and reduce the impact when attrition of robots occurred. Another insight provided was on the crucial role the human supervisor performs. While the human supervisor was a mission enabler, providing high-level commands and coordination, they were also the weak link, where the cognitive load of the human supervisor could limit performance.

\subsection{Contributions}
\label{sec:contributions}
The key differentiator in our solution is the homogeneous sensing capability, which enables shared maps between all agents. UGVs and UAVs both utilise spinning lidars which power both mapping and autonomy, with cameras providing object detection capability. The highly modular UGV solution centred around a common sensing pack and navigation stack has enabled rapid adaption to a wide range of platforms. All agents utilise the same SLAM system, which develops complete shared maps on each agent. These were exploited to provide shared global maps between all ground agents, and multi-robot task allocation.

This paper includes a brief description of our full system approach, with further detail on aspects that changed from our Phase I and II system described in \cite{hudson2021heterogeneous}. Further details of the earlier work can be found in \cite{hudson2021heterogeneous}; particular differences from this approach include:
\begin{itemize}
\item The UGV exploration system was revised to calculate traversability frontiers, utilising multi-agent global mapping data.
\item A new planner was developed that specifically targeted passing through narrow gaps.
\item The multi-robot task allocation reward function was reformulated to incorporate elements that consider the positioning of all agents in the tunnel network.
\item The tools available to the human supervisor for prioritising tasks and altering assignments were greatly improved.
\item The object tracker was redeveloped to associate new detections to full multi-robot histories (as opposed to a recent window on the local agent).
\item The Boston Dynamics Spot was incorporated in our robot fleet, and integrated with our common sensing and autonomy stacks.
\item The drop nodes and dropping mechanisms were redesigned based on lessons in prior events.
\item A new UAV platform was adopted to overcome the limitations of the previous platform, and hardware was integrated to permit use of the same object detection pipeline as ground agents.
\item SLAM was modified to selectively share frames, avoiding repeated sharing of identical frames when the robot is not moving significantly. 
\end{itemize}
Additionally, we describe in detail, the results of the approach at the Final Event of the DARPA Subterranean Challenge. In particular:
\begin{itemize}
\item For each run, details of operator's intentions, complicating events and results.
\item Communications performance, use of autonomy and human intervention.
\item Object detection performance and lessons learned.
\item Analysis of the mapping performance achieved in comparison with the ground truth scans provided by DARPA.

\item Experiences providing remote support due to severe limitations on the deployed team.
\end{itemize}

\section{System Description}
\label{sec:system_description}
\begin{figure}[ht]
    \centering
    \includegraphics[width=120mm]{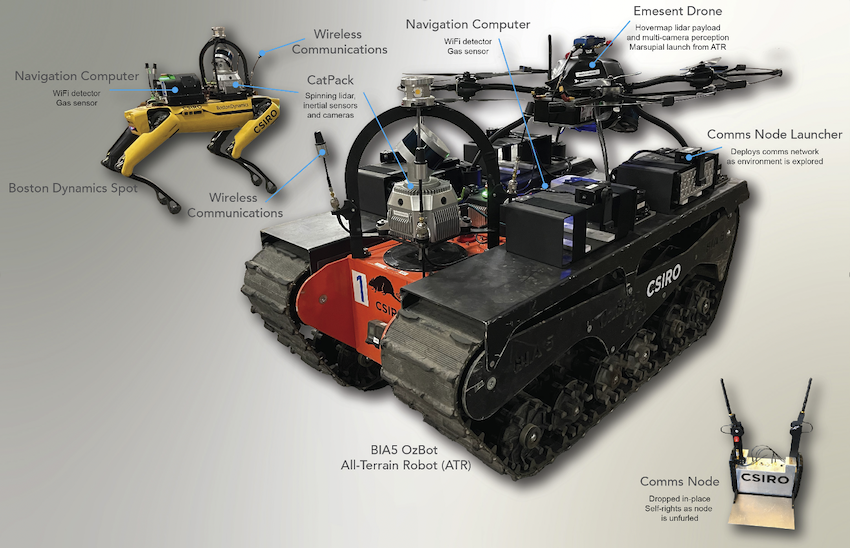}
    \caption{Team CSIRO Data61's robot platforms deployed during the final event.}
    \label{fig:annotated_robot_platforms}
\end{figure}

SubT involves a fleet of robots autonomously exploring an underground environment under the control of a single human supervisor. The Final Event combines the challenging elements from tunnel (e.g., large scale), urban (e.g., stairs) and cave (e.g., extreme terrain) environments. The solution fielded by Team CSIRO Data61 at the Final Event involved two Boston Dynamics Spot robots\footnote{\url{https://www.bostondynamics.com/products/spot}}, two BIA5 OzBot All Terrain Robots (ATRs)\footnote{\url{https://bia5.com/}}, and two Emesent UAVs carried by the ATRs. Communications nodes based on the Rajant Breadcrumb ES1 were carried on the ATRs and deployed through the course. The fleet employed is illustrated in \figref{fig:annotated_robot_platforms}.

In this section, we describe the hardware components of the system. We begin in \secref{sec:UGV_platforms} describing the UGV platforms developed for the Final Event, and subsequently the UAV system in \secref{sec:UAV_platform}. Finally, \secref{sec:comms_nodes} describes the design of the communications system.

\subsection{UGV Platforms}
\label{sec:UGV_platforms}
\begin{figure}[ht]
    \centering
    \includegraphics[width=120mm]{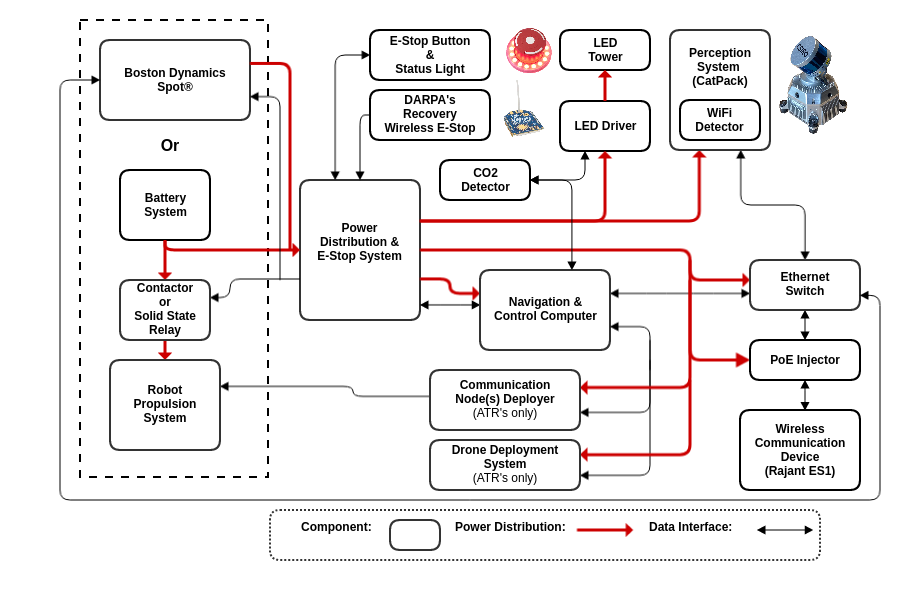}
    \caption{Hardware systems architecture for the Spot, DTR and ATR platforms.}
    \label{fig:ugv_hardware}
\end{figure}

The hardware systems architecture for the UGV platforms is summarised in \figref{fig:ugv_hardware}. A number of minor changes were made to the architecture deployed in the earlier phases of the SubT Challenge. The primary change to \figref{fig:ugv_hardware} is the addition of the Boston Dynamics Spot and the removal of a USB hub connecting external sensors. These changes and modifications are described in the following subsections.


\subsubsection{BIA5 OzBot All Terrain Robot}
\label{sss:atr}
The BIA5 OzBot All-Terrain Robot (ATR) was first utilised by Team CSIRO Data61 in the Tunnel Circuit event in August 2019. Subsequently, a light-weight (90\,kg vs 300\,kg) version was custom-built for CSIRO by BIA5 and employed in the Urban Circuit event and cave testing performed in lieu of the Cave Circuit event.  

The same robots were utilised in the Final Event, using the same LiFePO$_4$ batteries, power system and Cincoze DX-1100 ruggedised workstation (with Intel i7-8700T CPU). An evolved version of the ``CatPack'' perception pack was utilised, providing lidar, IMU and cameras with integrated compute performing SLAM and object detection. The new CatPack version had a full machined aluminium housing instead of the previous mixed aluminium and printed plastic housing, providing better dust and moisture protection, more effective cooling and better camera lens mounting giving improved image focus. A WiFi and Bluetooth module was also integrated into the CatPack and could be used for artefact detection instead of USB connected modules.

Extensive improvements were made to the robot's motors and motor controllers. The motivation for this was two-fold; firstly, to overcome issues with burn-out of motors in autonomous operation; and secondly, to provide finer control over the paths that the robot executed. The changes made are representative of those that have been found to be necessary to support autonomous operation on other platforms designed for teleoperation. The platform speed is 1.2\,ms$^{-1}$.

In relation to robustness, motors were burned out on a number of occasions. The first instance of this occurred during the Urban Circuit event, where a track became fixed on a large hook concreted into the ground. Subsequently, additional burn-outs occurred on a number of occasions where one track lost traction and spun repeatedly as autonomy attempted to recover from the condition. This was addressed firstly by introducing higher torque motors (a custom 108\,mm out-runner motor configuration coupled with BIA5's gearboxes) that were capable of handling more heat, and secondly by upgrading the motor control system and introducing a thermal model of the motor. The upgraded motor controller used an Elmo Motion Control (MC) system, providing feedback on current, temperature, velocity and acceleration at 200\,Hz. 

The thermal modelling introduced to this system was used to estimate the core temperature of the motor from an externally mounted thermistor on the motor casing. The thermal transfer from the motor core to the casing was based on a model provided by the motor manufacturer \cite{maxon_motor}. On each update of the motor communications loop, the motor current and thermistor readings would be fed into the model to get an internal temperature estimate. This allowed the system to apply significantly more power to the tracks in short bursts without risking a burnout, and consequently increasing the overall agility of the platform.


In relation to fine motor control, the stock motors and controllers had a range of difficulties, such as low-rate feedback (10\,Hz), simple PID-based control, fixed acceleration limits and an opaque interface. Most significant among these was the PID-based control, which resulted in significant overshoot due to a large integral wind-up to overcome stiction. This presented a significant barrier in tight environments such as when navigating through a narrow doorway. In contrast, the Elmo MC's built-in system identification was utilised to obtain precise velocity control. The higher-bandwidth 200\,Hz feedback also enabled improvements to the outer control loop, which was executed over a dedicated Ethercat interface. This higher rate feedback was utilised to implement a differential drive controller which balanced the desired level of aggressive control with smoothing when excessive velocity changes where commanded (e.g., when switching to a time-critical recovery behaviour).

Our efforts provided us with a highly robust platform capable of aggressive navigation, yet with exceptional fine control. This platform demonstrated robust control in adversarial terrain conditions in testing such as large rock piles and stairs, as well as challenging deformable cave environments in the self-led Cave Circuit. Around the time of the Final Event, further motor replacements were required, but these appeared to be related to deterioration over time rather than particular traumatic events. We plan to address this in future design iterations through an integrated gearbox, motor and controller solution.

\subsubsection{CSIRO Dynamic Tracked Robot}
\label{sss:dtr}
\begin{figure}[ht]
\centering
\begin{subfigure}{.5\textwidth}
  \centering
  \includegraphics[height=55mm]{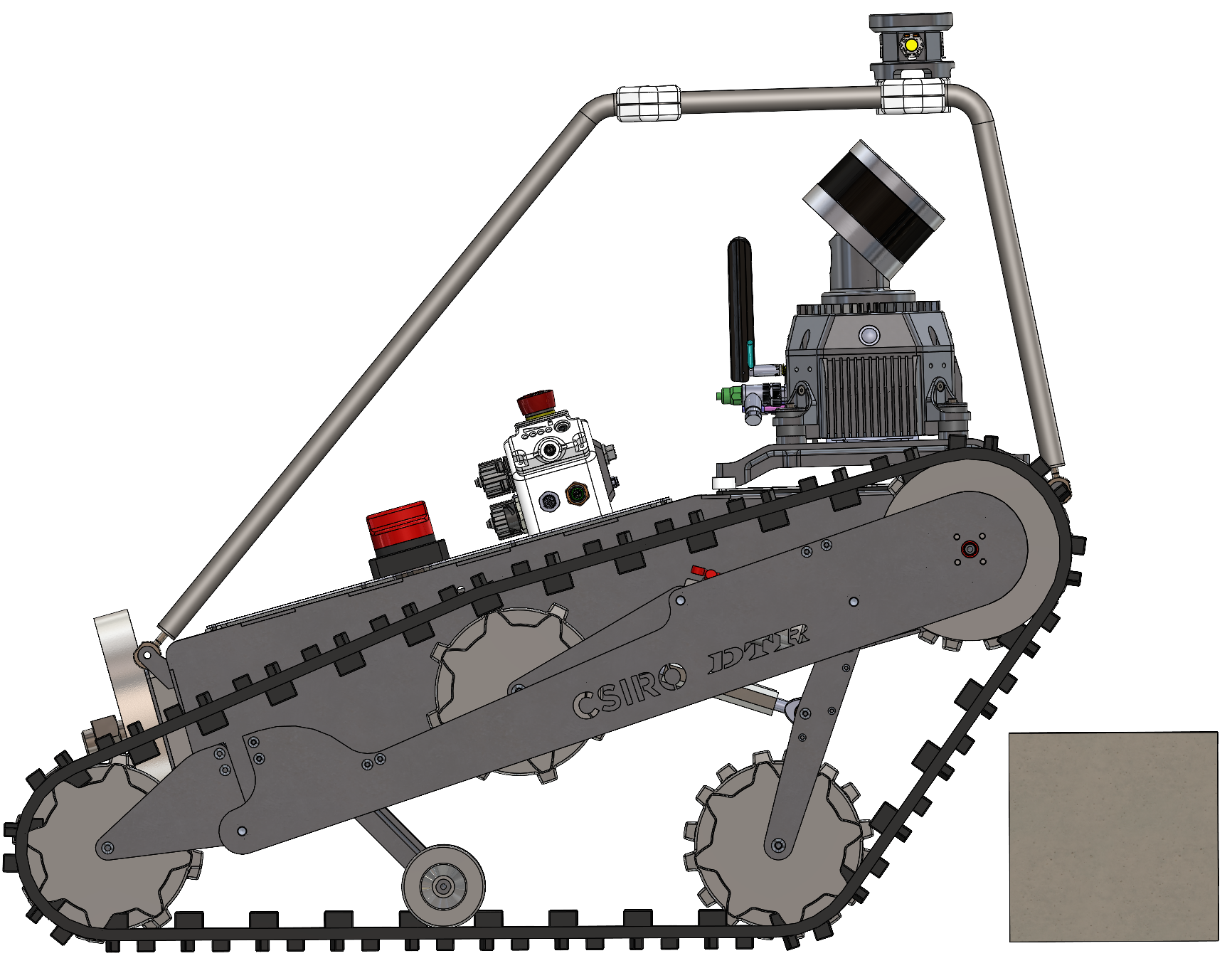}
  \caption{}
  \label{fig:dtr_cad}
\end{subfigure}%
\begin{subfigure}{.5\textwidth}
  \centering
  \includegraphics[height=55mm]{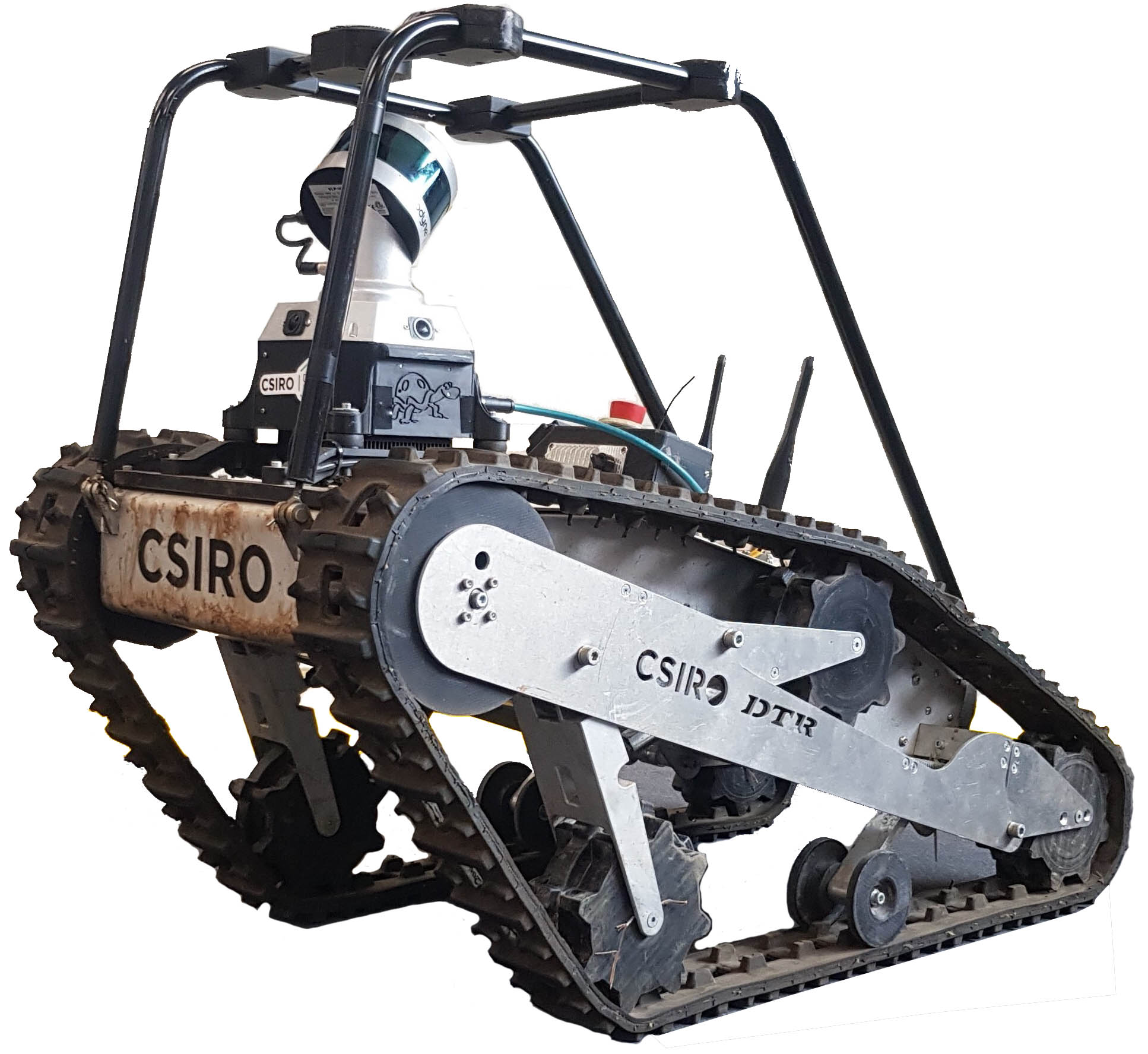}
  \caption{}
  \label{fig:dtr_iso}
\end{subfigure}

 \caption{The new DTR design showing its improved obstacle clearance ability next to a 180\,mm high step in a CAD diagram (a), and the real robot carrying an earlier version of the CatPack perception pack (b).}
\label{fig:dtr}
\end{figure}

The CSIRO Dynamic Tracked Robot (DTR) was developed to combine the strengths of the ATR with those of the smaller SuperDroid LT2-F, particularly around stairs and narrow doorways. Development of this platform continued from its initial design used at the self-led Cave Circuit up to the Final Event. Mechanically, the track tension mechanism was improved by coupling the position of the tensioner wheel and suspension arm. Their coupled motion maintain a constant tension in the track throughout the motion range of the suspension arm. This maintained soft suspension, but achieved the high track tension needed to provide robustness to small rocks in the tracks. The final design of the platform is shown in \figref{fig:dtr}; the platform is capable of operating at a speed of 3\,ms$^{-1}$, but is configured to use a similar speed to the ATR due to the tuning of autonomy.

The design maximised commonality with the ATR components, and served as a prototype of the Elmo MC system describe in \secref{sss:atr}. This commonality was an explicit design goal, to permit either platform to provide spare parts for the other. The same CatPack perception was utilised as the ATR, while the navigation computer was based on an Intel NUC NUC8i7BEH.

Tuning of the autonomy stack for the DTR was de-prioritised when the Spot was integrated into the robot team, but continued at a low rate of effort. It was intended that the platform would feature in the team at the Final Event, but this needed to be dropped due to the minimal team that was able to be sent due to COVID travel restrictions. An accurate simulation model of the platform was developed, and was utilised by Team CTU-CRAS in the DARPA SubT Challenge Virtual track. The platform is seeing continued use, particularly in natural environments where it has less impact than the larger and heavier ATR platform.

\subsubsection{Boston Dynamics Spot Quadruped Robot}
\label{sss:spot}
The Boston Dynamics Spot platform is a COTS quadrupedal legged platform. It features a top speed of 1.6\,ms$^{-1}$, max payload of 14\,kg and a typical operation time of 90\,min (unloaded; typical operation times were 40-45\,min with a payload). It has stereo cameras positioned around the body which allow it to generate a 360\degree map of the local terrain. This map, combined with state-of-the-art locomotion software, allows it to traverse a wide array of terrain. The Spot platform was used to navigate into areas that were unsuitable for the tracked ground platforms, including stairs, and narrow passageways \cite{tam2021deploying}.

Previously the Ghost Robotics Vision60 Platform had been fulfilling this role within the fleet. The decision to switch to the Spot platform was made based on testing its performance in these specific areas. Predominant capabilities of the Spot are understood to be due to its terrain sensing: Unlike the Spot platform, the version 4.2 Vision60 platform available at the time did not provide any local terrain mapping or footfall planning, which made it unsuitable for traversing stairs, a key area the legged platform was aimed at addressing. The Spot platform also provided a payload interface that was more robust, better documented, and supported for the user. This not only allowed the team to rapidly integrate Spot into the fleet, but also allowed the onboard sensor data to be integrated into the team's autonomy stack as described in \secref{sss:spot_local_nav}.

As illustrated in \figref{fig:annotated_robot_platforms}, the Spot was fitted with the same CatPack perception pack utilised in the other platforms, along with the same navigation computer utilised with the DTR and the earlier Ghost platform. Communications were provided by a Rajant ES1 Breadcrumb node as detailed in \secref{sec:comms_nodes}, however the platform was not equipped to drop additional nodes.

\subsection{UAV Platform}
\label{sec:UAV_platform}
\begin{figure}[t]
    \centering
    \begin{subfigure}{.48\linewidth}
    \includegraphics[height=40mm]{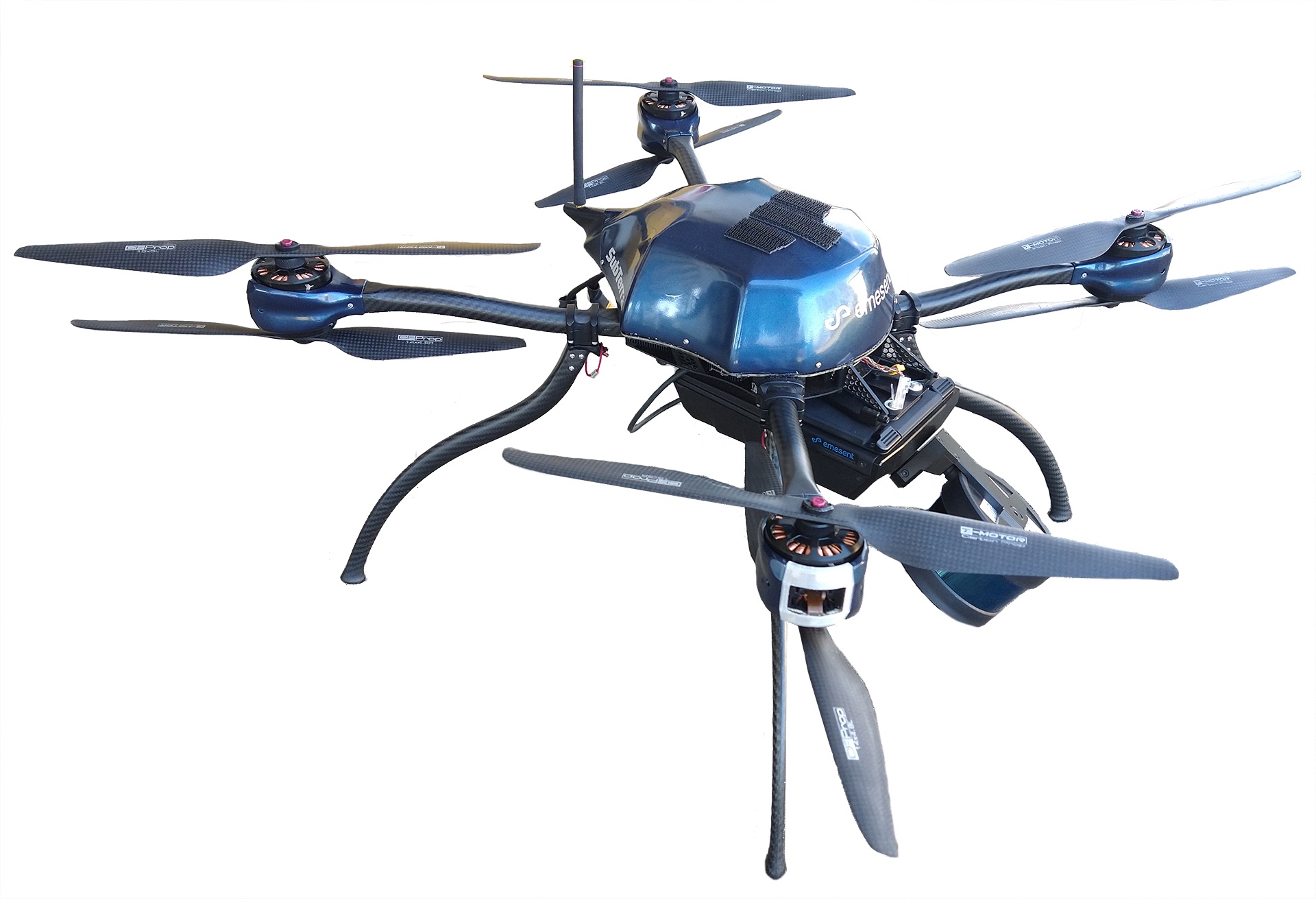}
    \caption{}
    \label{fig:titan_cave_rocks}
    \end{subfigure}
    \begin{subfigure}{.25\linewidth}
    \includegraphics[height=40mm]{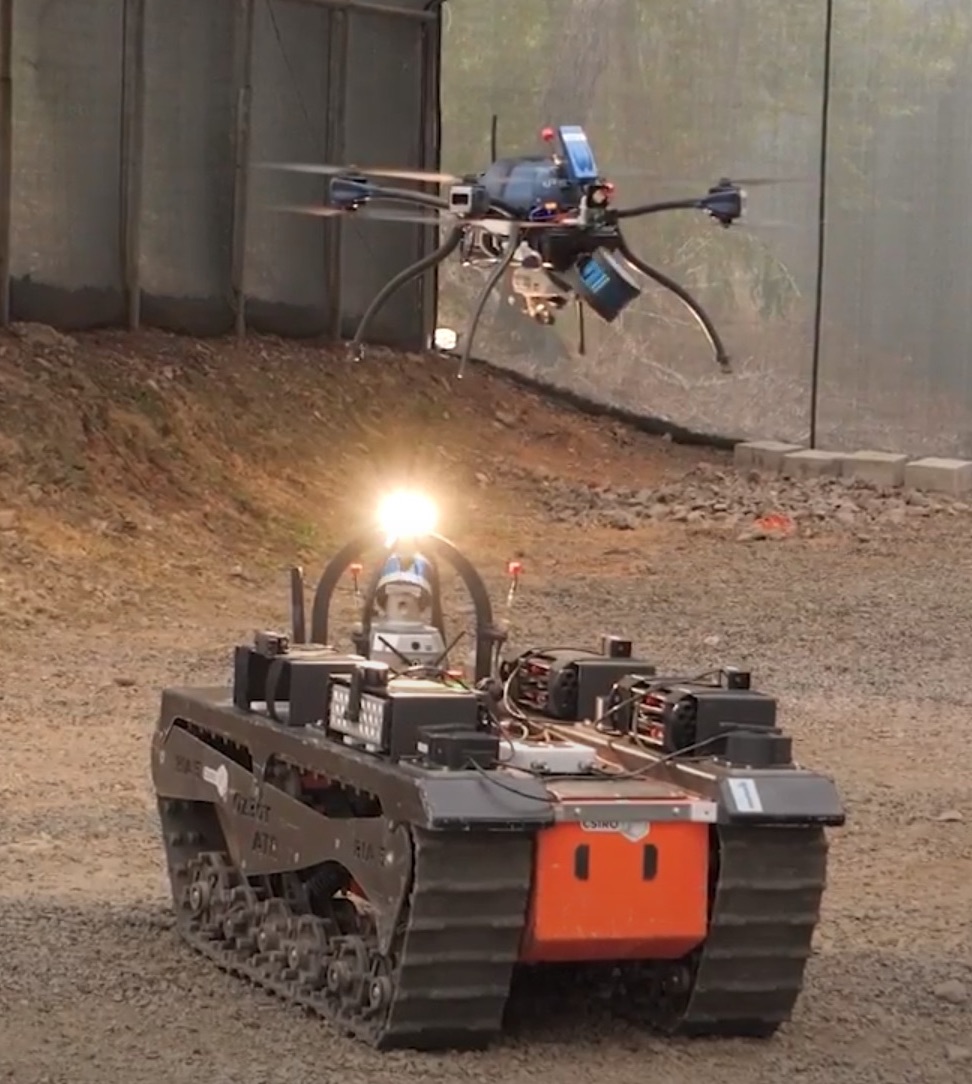}
    \caption{}
    \label{fig:titan_tunnel_node_dropped}
    \end{subfigure}
    \begin{subfigure}{.25\linewidth}
    \includegraphics[height=40mm]{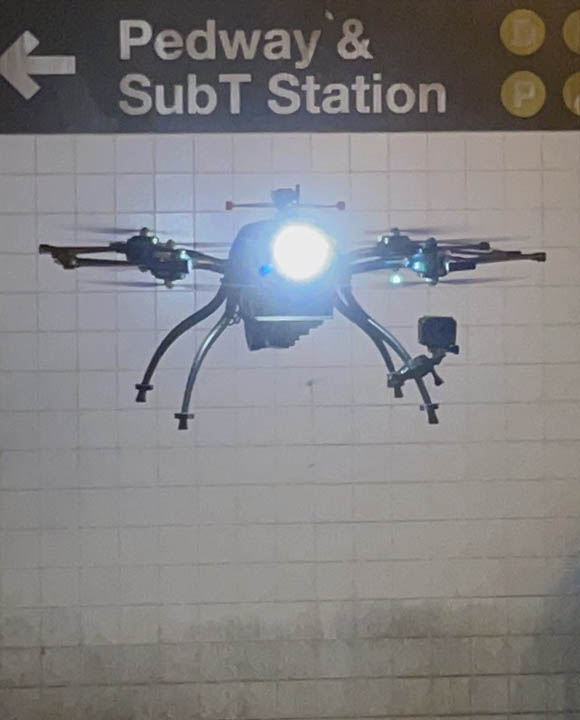}
    \caption{}
    \label{fig:titan_tunnel_drone}
    \end{subfigure}
    \caption{Emesent/Aeronavics SubTerra (``Navi'') UAV (a), being launched off the back of an ATR UGV (b), and in flight inside the DARPA SubT finals course (c).}
    \label{fig:navi_drone}
\end{figure}

The SubTerra ``Navi''\footnote{\url{https://aeronavics.com/models-of-drones/navi/}} UAV used in the SubT Finals is the first UAV specifically designed to work with the Emesent Hovermap payload (\figref{fig:navi_drone}). This platform was commissioned from the New Zealand-based company, Aeronavics, by Emesent.

Throughout all previous circuits, our UAV platform of choice was the DJI M210. Whilst this platform was robust and met our size and payload requirements, it provided some challenges when pushed beyond its manufacturer expected use cases. Most critical were random events in which the UAV would override take-off commands due to perceived magnetic interference. The act of moving an ATR through a circuit with an M210 on the back could cause sufficient interference with the UAVs in-built sensors, raising an error. A further problem was that, once fully loaded with the Hovermap, communications node and vision system, it only provided around 8\,min of flight time. With these problems in mind, a full configurable platform with a 20\,min battery life and high payload capacity was required to allow for full control over every aspect of the system and achieve our desired coverage goals.

With no commercially available UAVs on the market at the time, Emesent commissioned a custom solution from Aeronavics based off of their existing Navi UAV. The Navi provided an Ardupilot-based system that met our transparency, flight time and lift capacity requirements. Unfortunately, the development of this platform was problematic requiring significant effort to improve platform reliability. Such reliability issues included electronic speed controller (ESC) burnout, debris ingress through open motor housings, battery failure and short life-span, and communications issues with the ground station RC. Ultimately, most of these problems were solved or mitigated before the final circuit and the platform proved itself. In the second preliminary round, one Navi platform was downed by a sheet of foam in mid-air, however the system sustained no major damage to the air-frame, Hovermap or vision system, only requiring prop replacement to be back in the air within an hour of returning to the pit.

\begin{table}[b!]
  \begin{center}
    \caption{UAV Specifications - Loaded with Hovermap and Vision System.}
    \begin{tabular}{lr}
    \toprule
      \textbf{Property} & \textbf{Value}\\
      \midrule
      Length & 740\,mm\\
      Length (With propeller guards) & 900\,mm\\
      Width & 740\,mm\\
      Width (With propeller guards) & 900\,mm\\
      Height & 350\,mm\\
      Weight & 5.8\,kg\\
      Nominal clearance from robot center & 850\,mm \\
      Minimum passage width & 2000\,mm\\
      Minimum take-off clearance & 1000\,mm\\
      Flight time (Fully Loaded) & 20\,min\\
      Nominal movement speed & 1.5\,ms$^{-1}$\\
      Maximum movement speed & 2.0\,ms$^{-1}$\\
      \bottomrule
    \end{tabular}
  \end{center}
\end{table}

A feature of this platform were propeller guards. These were installed when needed as they traded increased survivability in the air for reduced ability to enter narrow passageways while riding the parent vehicle. Whilst on the ATR, the propeller guards extended beyond the UGV's 780\,mm width, preventing the ATR from entering as narrow passages. The guards also increased the UAV's width in flight, but the UAV could still navigate the same size tunnel, as light contact with the wall would no longer guarantee a crash. In the final circuit, propeller guards were installed for the first preliminary run to ensure that the UAVs would return intact. For the second and final runs, a more risky approach was taken and guards were removed on all UAV systems.

\subsubsection{UAV Vision System}
\begin{figure}[t]
\centering
\begin{subfigure}{.5\textwidth}
  \centering
  \includegraphics[height=40mm]{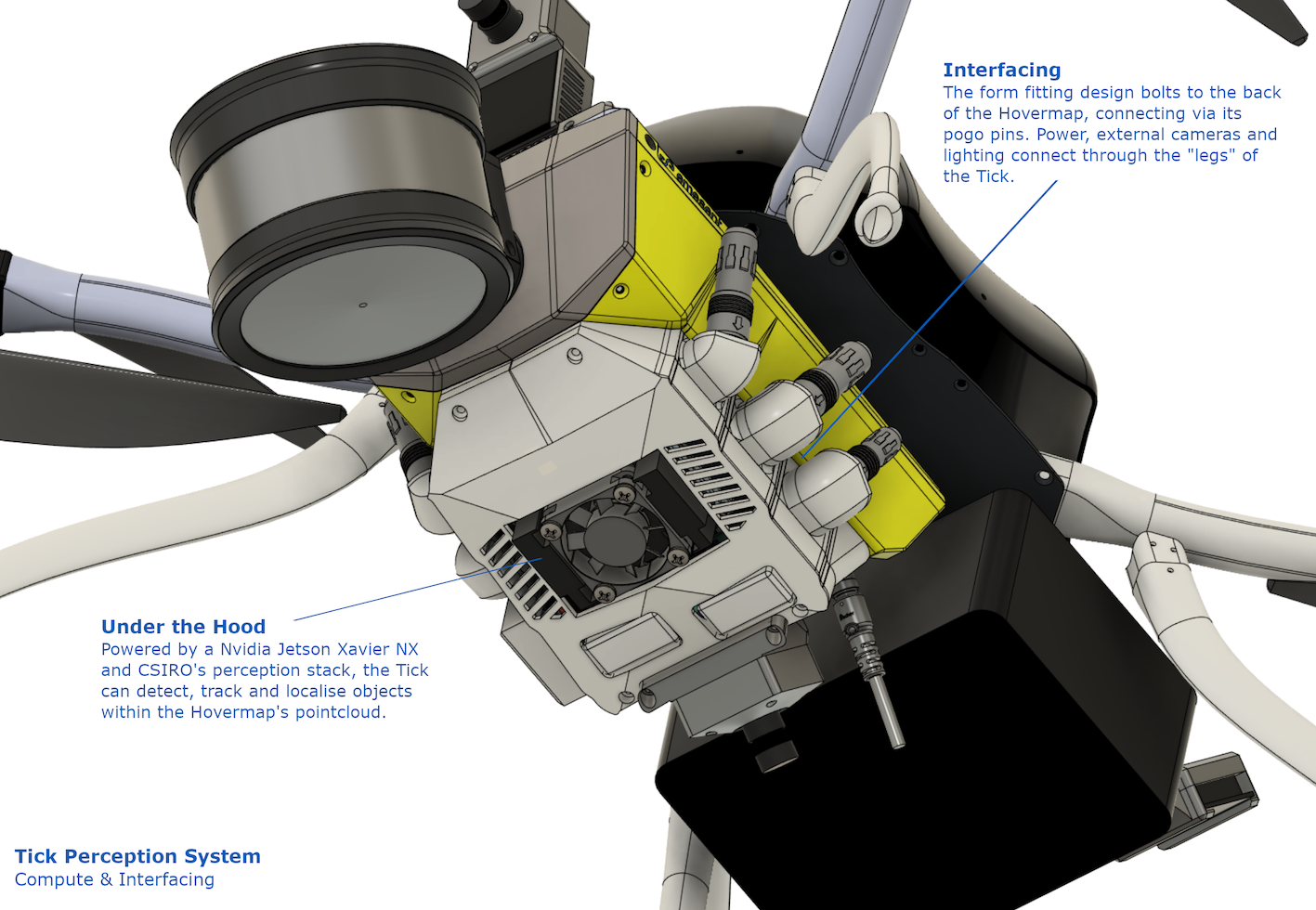}
  \caption{}
  \label{fig:annotated_tick_compute}
\end{subfigure}%
\begin{subfigure}{.5\textwidth}
  \centering
  \includegraphics[height=40mm]{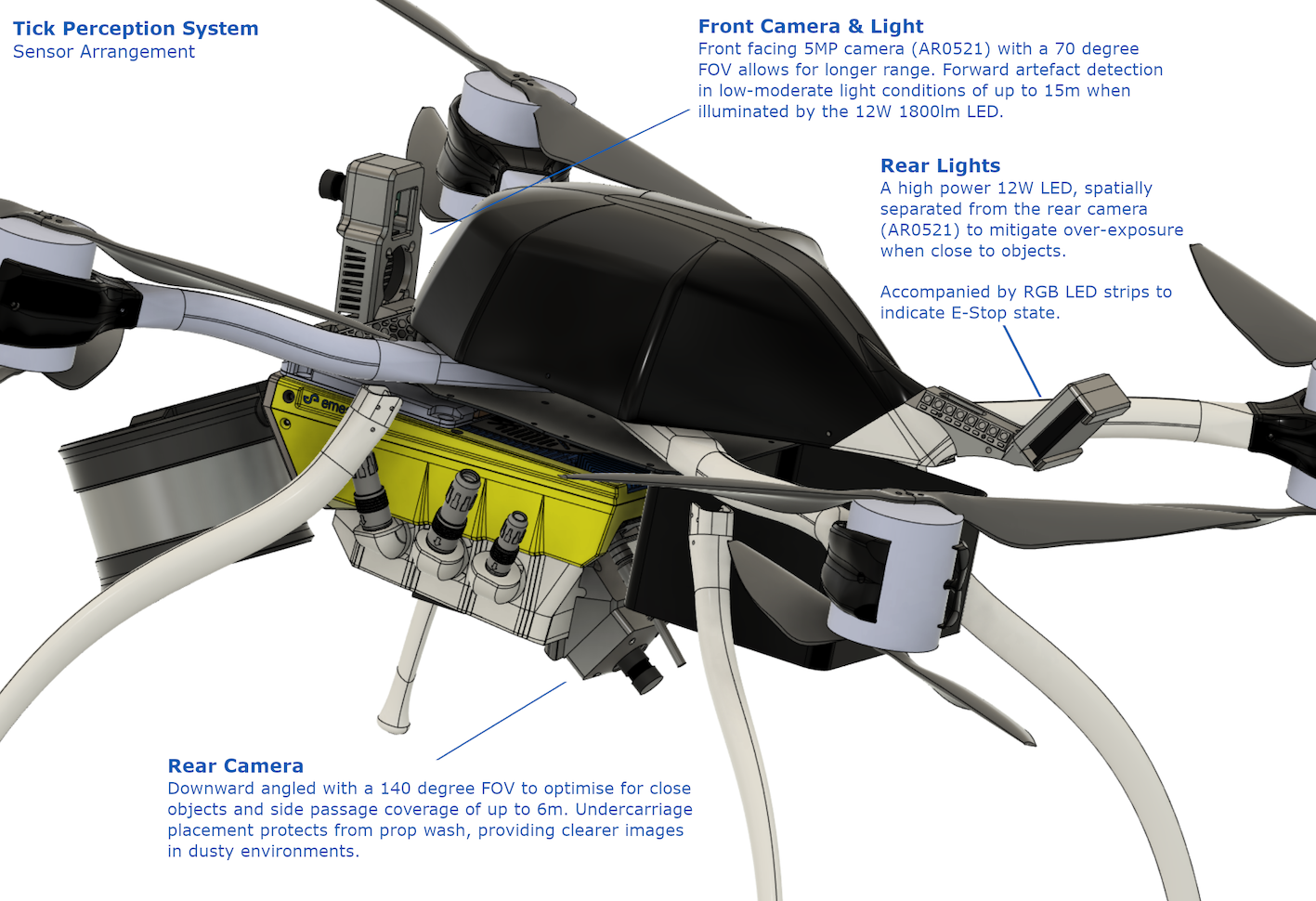}
  \caption{}
  \label{fig:sub2}
\end{subfigure}
\caption{Anatomy of the vision system (the Tick) on an Emesent Navi UAV showing compute and interfacing (a) and sensor arrangement (b).}
\label{fig:annotated_tick_sensors}
\end{figure}

Accompanying the Navi platform was a new vision system called the ``Tick,'' (\figref{fig:annotated_tick_sensors}), which was developed in parallel with improvements to the previous gimbal-based approach before being chosen as the UAV vision system for the final circuit. 

Prior to the Tick, a gimbal-based vision system was employed by our UAVs. This was an in-house design utilising off-the-shelf electronics to create a fully controllable and stabilized gimbal. A single FLIR BFS USB 3.0 camera would feed frames through the Hovermap to an Intel Neural Compute Stick 2 for image classification. This secondary compute unit was required as on-board image classification would compete with SLAM, trajectory planning and other core processes for CPU resources, leading to higher image classification latency and a lower output rate. 

The gimbaled camera system, along with our visual coverage tracking software, could scan the course independently from UAV orientation. Compared to a static system, this provided enhanced vertical coverage of the course and a higher chance to detect objects when the platform was stationary. In theory, the single moving camera with a narrow-moderate field of view (FOV) allowed for both highly detailed images at long ranges and large field coverage given enough time in one region. However, due to the limited flight time of small UAV platforms, it was not practical to loiter in one region for extended periods, thus in prior SubT circuits it suffered from limited coverage and missed target objects.

Due to these coverage limitations, we developed a static, multi-camera system with on-board processing utilising CSIRO Data61's perception stack. As the Hovermap host system was used to run the Wildcat SLAM solution, this package could leverage smaller NVIDIA Jetson hardware than the CatPack used on the UGV. This system was called the Tick due to its parasitic nature on the Hovermap. 

At its core, the Tick ran a NVIDIA Jetson Xavier-NX on a Connect-tech Quark carrier board, allowing for dual MIPI-CSI cameras. Whilst the Xavier-NX had the connectivity and processing power required to run more cameras, carrier board and weight restrictions limited this to two cameras. The specific camera arrangement and lens combination was the single largest factor in the performance of the Tick. Initially the Tick was configured with dual downward angled, forward/side facing, wide FOV (160-180\degree) cameras to maximise coverage, with particular focus on the space directly in front of the UAV and that of unexplored pathways/nooks to either side. Testing showed that this configuration suffered from problems including: 
\begin{itemize}
\item A short range of only 5-6\,m due to the large FOV of each camera.
\item High power consumption due to the lighting requirements for such a large FOV.
\item Over-exposure of the image due to reflections from UAV legs.
\item After transitioning from the M210 to the Navi, the new leg setup with four legs placed at the UAV corners blocked a significant part of the camera FOV.
\end{itemize}

Further testing led to the final configuration, consisting of one front-facing, 70\degree FOV camera and one rear, downward facing 140\degree FOV camera. This configuration was able to maximise both the systems coverage and range by relying on the tendency of the exploration code to stop and spin in place when passing an unexplored passageway.  The rotation of the platform allowed the fixed cameras to cover a larger FOV, with the board FOV (140\degree) sensor specializing in targets close to the UAV ($\sim$5-8\,m), while the narrow FOV camera could search for targets at ranges of up to 15\,m. Continued improvements to the system included aspects of the lighting configuration. Specifically, increasing vertical separation between the rear lights and camera significantly reduced image blow-out on close objects and increased the accuracy of the Tick.

\begin{table}[h!]
  \begin{center}
    \caption{Tick Specifications.}
    \begin{tabular}{lr}
    \toprule
      \textbf{Property} & \textbf{Value}\\
      \midrule
      Camera Sensor & AR0521\\
      Camera Resolution & 5.1\,MP\\
      Camera Frame-rate & 3\,FPS\\
      Front Camera FOV & 70\degree\\
      Front Camera Maximum Artefact Range & 15\,m\\
      Rear Camera FOV & 140\degree\\
      Rear Camera Maximum Artefact Range & 6\,m\\
      LED Power & 12\,W + 12\,W\\
      LED Luminous Flux & 1800\,lm\\
      Total System Power Consumption & 45\,W\\
      \bottomrule
    \end{tabular}
  \end{center}
\end{table}

\subsection{Communications System}
\label{sec:comms_nodes}
The communications system used in the Final Event closely matched the solution that was deployed at the Urban and Cave Events. COTS Rajant Breadcrumb nodes were used to provide a layer 2 (i.e., the data link layer in the OSI communications model~\cite{osi_zim_1980}) mesh network using wireless, dual-band (2.4\,GHz and 5.8\,GHz) 802.11 radio links. The base station and robot autonomy computers all used a standard Linux TCP/IP stack to communicate over this network. Layer 4 (transport layer) traffic was handled exclusively by a software component named Mule, described in \secref{sec:mule}.

\subsubsection{Base Station Communications}
The base station computer was connected to two Rajant Breadcrumb nodes: an ES1 and a Peregrine. Each node was mounted with antennae on dedicated masts that provided approximately 1.5\,m of ground clearance. The ES1 was connected to 4\,dBi (2.4\,GHz and 5.8\,GHz), multi-polarized, omni-directional, goose-neck style antennae whilst the Peregrine was connected to 9\,dBi (2.4\,GHz) and 12\,dBi (5.8\,GHz), directional sector antennae. The decision to use multiple node masts at the base station was made during the lead-up to the event and was motivated by the idea that the course entrance may have consisted of two or more narrow tunnels which would not be well handled by a single directional antennae. Though the actual course used a single entrance tunnel, the team decided to continue deploying both mast nodes in each run due to early results indicating a non-negligible degree of RF transparency through some of the course walls.

\subsubsection{Communications Node and Dropping Mechanism}
\begin{figure}
    \centering
    \subfloat[]{\includegraphics[width=0.55\textwidth]{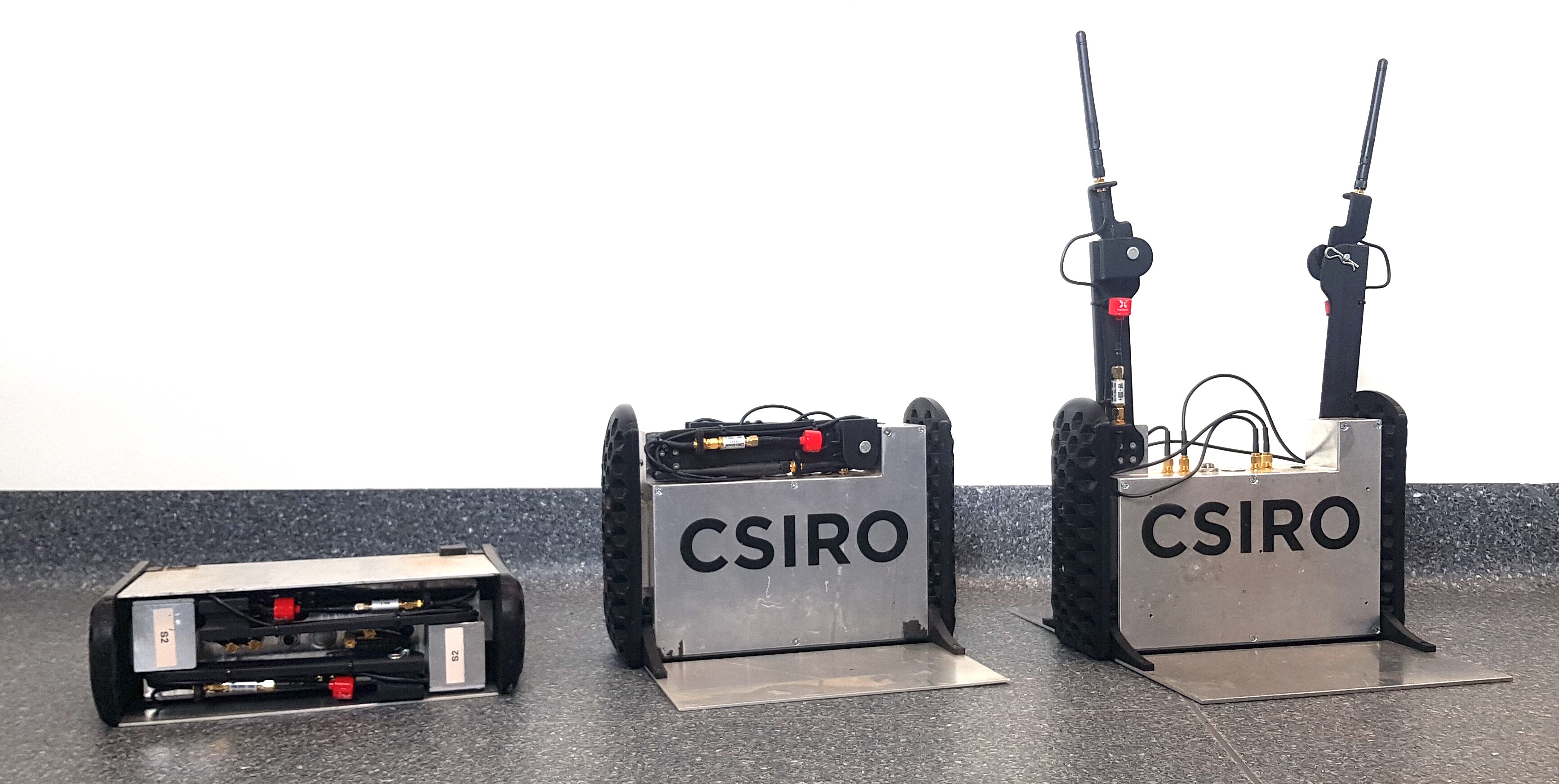}}~
    \subfloat[]{\includegraphics[width=0.4\textwidth]{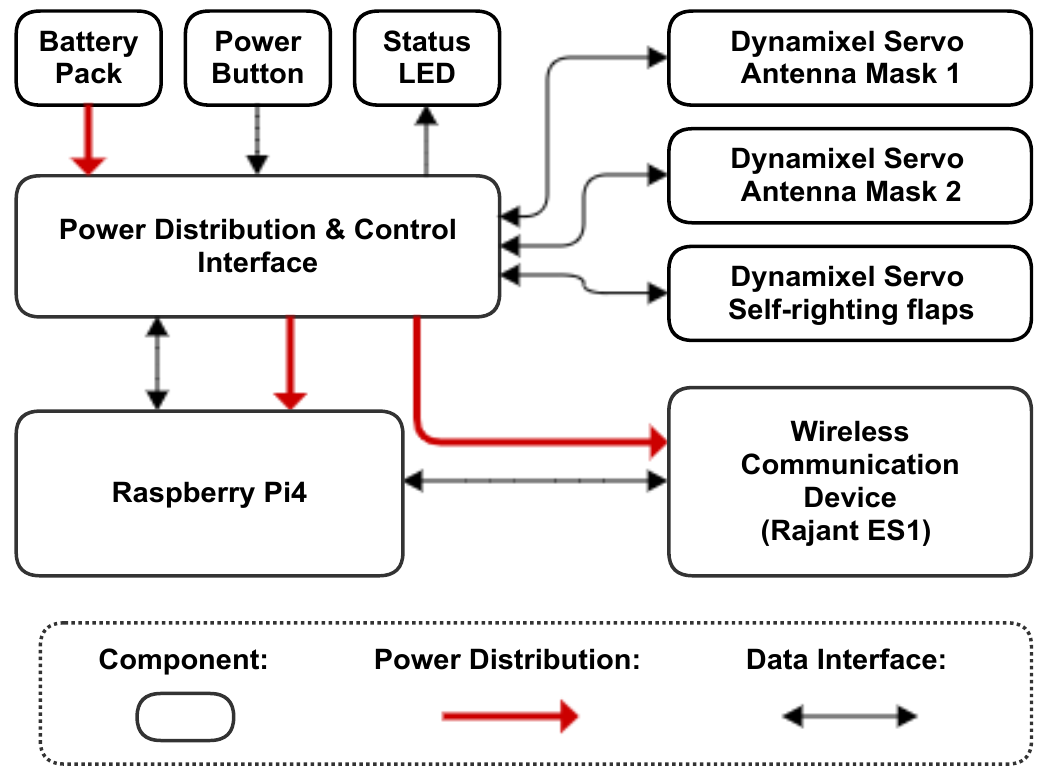}}
    \caption{(a) Communications node packing, showing stages of deployment: (left) as dropped from the robot, (middle) after side panels unfurl, self-righting the node, and (right) after antenna deployment. (b) Block diagram of communications node design.}
    \label{fig:comms_node}
\end{figure}

Whereas the earlier communications node design was a minor repackaging incorporating a battery into the Rajant Breadcrumb ES1, the node utilised in the Final Event underwent a major redesign. The core ES1 node was supplemented by a Raspberry Pi 4, which served two functions. Firstly, it controlled deployment of the node. Rather than being statically dropped, the revised node design actively unfurls two side panels, self-righting in the process, powered by Dynamixel motors. Subsequently, antennas unfurl to a configuration with sufficient height to provide improved communications quality, and finally the ES1 power is activated. The deployment process and design are shown in \figref{fig:comms_node}.

The Raspberry Pi 4 also provided the secondary capability of running an instance of the Mule communications software on the node. Due to re-prioritisation of tasks in the lead up to the final event (again, due to COVID-19 related travel restriction for the team), insufficient experimentation had been conducted with this concept prior to the Final Event, so the capability was not employed. However, it has potential to allow nodes at the boundary of communication to retain data from a robot as it goes deep beyond communication. 

The node dropper design also evolved for the Final Event. Lessons from testing in the cave environment showed that carrying node droppers which extruded beyond the vehicle footprint was highly undesirable. It was also found that dropping to the side of the platform was preferable over dropping in front or behind, as it inherently tended to result in node positions which did not block traversal of a passage. Accordingly, as shown in \figref{fig:annotated_robot_platforms}, four node droppers were installed on the track guards of each ATR (two on each side).

\subsubsection{UGV and UAV Communication Systems}

Each UGV was equipped with a Rajant ES1 node connected to 4\,dBi (2.4\,GHz and 5.8\,GHz), multi-polarized, omni-directional, goose-neck style antennae. Each UAV was equipped with the smaller Rajant DX2 node connected to 2.6\,dBi (5.8\,GHz only), omni-directional, lollipop style antennae due to size and weight constraints.

\subsubsection{Mule}
\label{sec:mule}
Mule was the layer 4 software component which bridged ROS topic messages between the independent ROS systems running at the base station and on each robot. Mule provided best-effort, end-to-end transport for ephemeral data such as robot status and teleoperation video, as well as disruption-tolerant, hop-by-hop transport for mission-critical data such as Wildcat frames, and object detection reports. These features enabled the operator and autonomy software to use data-muling as a strategy for improving exploration efficiency and overcoming robot attrition.

Improvements were made to the design of Mule for deployment in the final to expose more information to the human-operator, such as providing more detailed information on exactly what data was yet to be downloaded from a given peer, and a separation of
``synchronisation lag" into upload vs download metrics.

The capability provided by Mule was critical to the outcome achieved at the Final Event. For example, as described in \secref{sec:discussion}, in the Final Prize run, data from a fallen Spot robot was relayed to another platform that passed nearby, and later transmitted to the base. It was these data muling efforts that resulted in the final successful artefact report in the final critical seconds of the run.

\subsubsection{Communications System Improvements}
\label{sec:comms_improvements}
Despite the core design of the communications system remaining relatively unchanged between the Urban and Final Events, significant performance improvements were attained due to the resolution of previously undiscovered integration issues. 

The first of these issues was related to the active Bluetooth scanning being used for detection of the mobile phone artefact. At the Urban event, UGVs were equipped with a UD100 Bluetooth USB adapter and software that would repeatedly perform active Bluetooth ``inquiry'' scans to elicit responses from nearby Bluetooth devices. Each of these scans involved brief transmissions on a sequence of 1\,MHz Bluetooth channels spread across the 2.4\,GHz ISM band. These transmissions were unmanaged with respect to the 20\,MHz 802.11 channel that was used by Rajant Breadcrumbs in the same 2.4\,GHz ISM band. Integration testing conducted after the Urban event revealed that these Bluetooth transmissions were significantly degrading the performance of the Rajant mesh network. With consideration of overall system performance and time constraints, it was decided that the Bluetooth detection capability would be removed from UGVs altogether, relying on (passive) WiFi detection (for cell phones) and the visual signature of the cube artefact. A similar 2.4\,GHz interference problem was discovered with the remote control transceiver, however, this scenario was restricted to times when the handheld remote control unit was switched on and in close range of the robot (a situation that could not arise during actual competition). 

The second of these issues was related to unintentional mesh network traffic originating from devices other than the base station and autonomy computers. The hardware designs of the UGVs and UAVs both contained a single Ethernet network that was used to connect the robot autonomy computer with on-board sensors, controllers, and the Rajant Breadcrumb node. Early on in the development process, the Rajant Breadcrumb nodes were configured to use a bridging mode where layer 2 packets received via the local Ethernet connection could be transported via the mesh network regardless of the source address. This configuration simplified certain aspects of system development by allowing access to network-enabled robot hardware from any computer connected to a Rajant Breadcrumb. However, this configuration left the performance of the communications system vulnerable to unexpected sources of network traffic. Such sources were inadvertently introduced on multiple occasions during development. On each occasion, a new piece of hardware had been introduced that defaulted to a communication mode that generated a high-frequency stream of packets with a layer 2 broadcast destination address. This would result in a subtle but significant degradation of the communications system performance during subsequent system tests, requiring manual traffic analysis to detect and identify the offending hardware. Prior to the Final Event, it was decided that the Rajant Breadcrumbs would be configured to use a MAC address whitelist so that only layer 2 packets originating from the base station and robot autonomy computers could be bridged over the mesh network.

\section{Localisation and Mapping}
\label{sec:localisation_mapping}

Robust localisation and mapping are critical to downstream robotics tasks. For example, as described in the next section, robot autonomy requires the information of localisation and the environment map for path planning, multi-agent coordination requires knowledge of each robot's position on a shared map, and in SubT, detected artefacts must be reported with accurate locations. This section briefly describes Wildcat, our multi-agent lidar-inertial SLAM system used in the DARPA Subterranean Challenge. We refer the reader to 
\cite{ramezani2022wildcat} for additional technical and implementation details. A consolidated report on the experiences of the various SubT teams with localisation and mapping can be found in \cite{ebadi_present_2022}.

\subsection{Wildcat SLAM}
\label{sec:wildcat}
\begin{figure}[t]
    \centering
    \includegraphics[width=120mm]{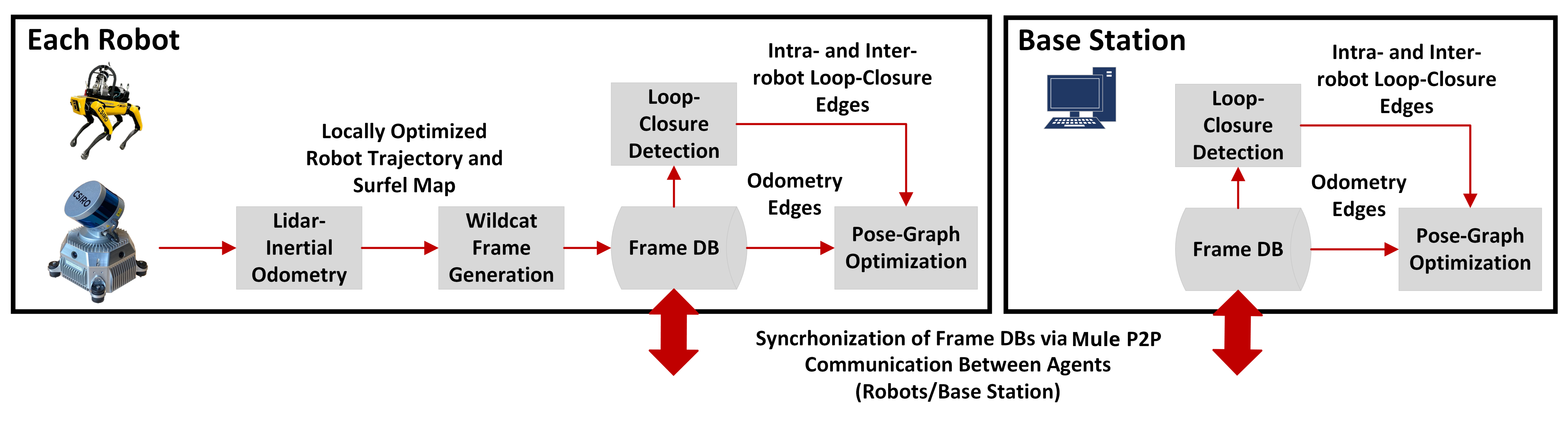}

    \caption{A diagram of Wildcat, our multi-agent lidar-inertial SLAM system. This diagram shows the components of Wildcat ran on each robot (left) and those that ran on the base station (right).}
    \label{fig:wildcat_diagram}
\end{figure}

A diagram of Wildcat is shown in Figure~\ref{fig:wildcat_diagram}. Wildcat has two major modules: (i) lidar-inertial odometry, and (ii) pose-graph optimisation. In the following, we briefly describe each module and present our SLAM results in the prize run of the DARPA Subterranean Challenge Final Event.

Wildcat odometry is a real-time sliding-window optimisation method that fuses IMU and lidar measurements collected within a three second window to estimate robot trajectory at a high rate. The odometry module runs on each robot independently.
First, Wildcat generates surfels from lidar points by voxelising them and fitting an ellipsoid to the points residing in each voxel. Only those surfels that are sufficiently planar are kept.

After establishing an initial set of correspondences between the surfels, Wildcat then estimates robot trajectory by minimising the point-to-plane distance between the matched surfel pairs and also the error between the predicted and collected IMU measurements (angular velocity and linear acceleration). Wildcat odometry alternates between matching surfels and estimating robot trajectory for a fixed number of iterations. Our method uses cubic B-spline interpolation to remove distortion from lidar points (caused by the motion of robot and lidar) and to efficiently fuse asynchronous measurements from lidar and IMU in the  previously mentioned optimisation problem; see \cite{ramezani2022wildcat}.

Every five seconds, the locally optimised robot trajectory is used to create a local surfel map. We call each rigid local map (together with additional information such as the corresponding segment of trajectory estimate) a \emph{frame}. Frames remain rigid after creation and thus their state can be represented by one of the underlying poses. Each robot stores its own frames, as well as frames received from other robots in a database. Frame databases are synchronised between the agents (robots and the base station) using Mule whenever peer-to-peer communication is possible (see \secref{sec:mule}); frames are suppressed from being shared if the overlap to the previously shared frame exceeds a threshold. Wildcat's pose-graph optimisation (PGO) module (also referred to as Atlas) runs separately on each agent and aims to produce an independent, globally consistent estimate of the team's map and trajectories. This module detects intra-robot and inter-robot loop closures using all available frames. Each agent then independently optimises the team's collective pose graph whose nodes correspond to (unknown) frames' representative poses, and whose edges correspond to odometry and loop-closure measurements.
In the prize run, the average total size of frames generated by our four UGVs (introduced in~\secref{sec:UGV_platforms}) was about 21.5\,MB per robot.

\begin{figure}[t!]
    \centering
    \includegraphics[width=120mm]{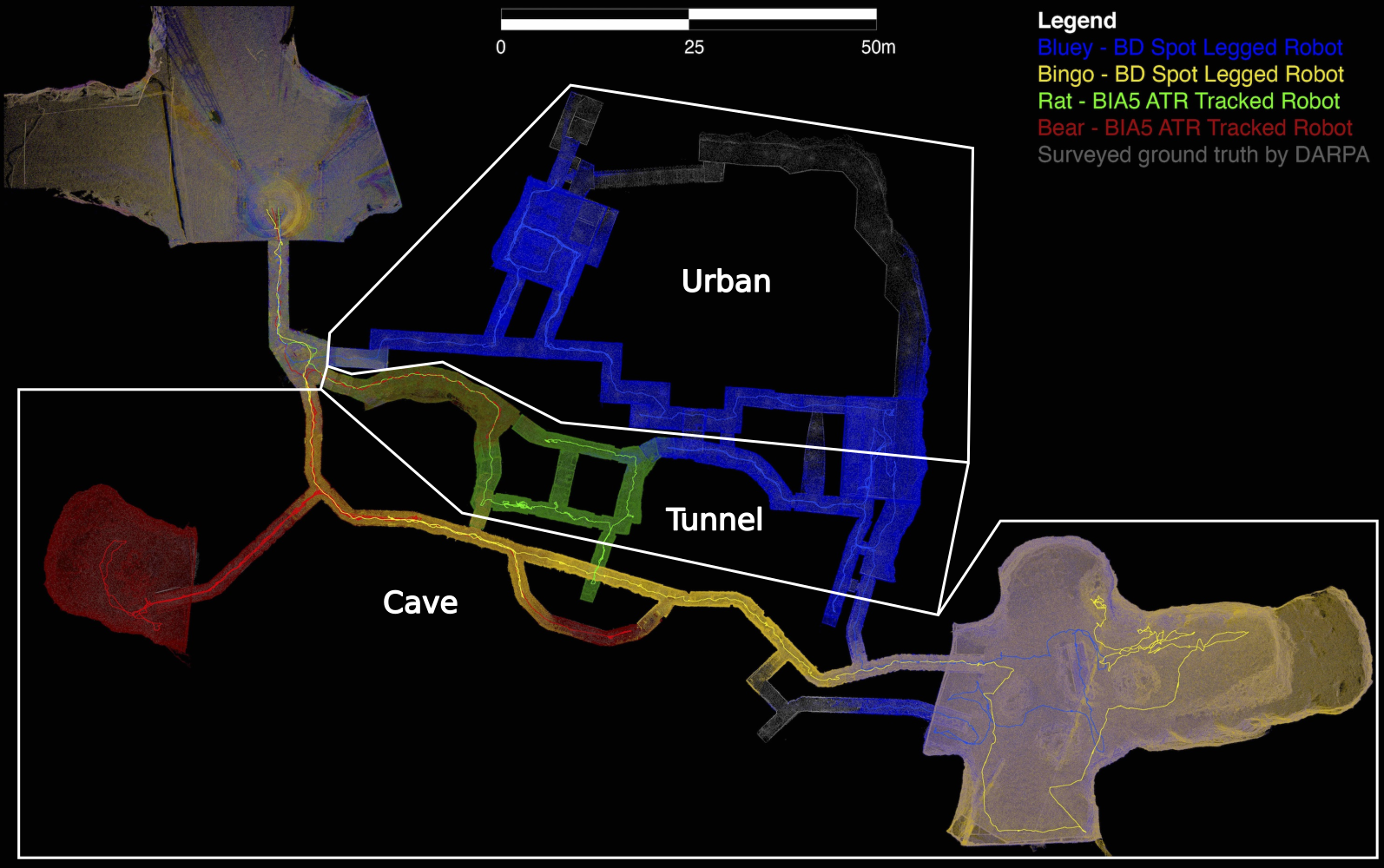}
    \caption{Multi-agent globally optimised Wildcat SLAM map from the robots deployed by team CSIRO Data61 during the 60\,min prize run. Point clouds collected by different robots are shown by colour, while white lines delineate the three course environments (i.e., urban, tunnel and cave).}
    \label{fig:prizerun_multiagent_offline}
\end{figure}

\begin{figure}[t!]
    \centering
    \includegraphics[width=120mm]{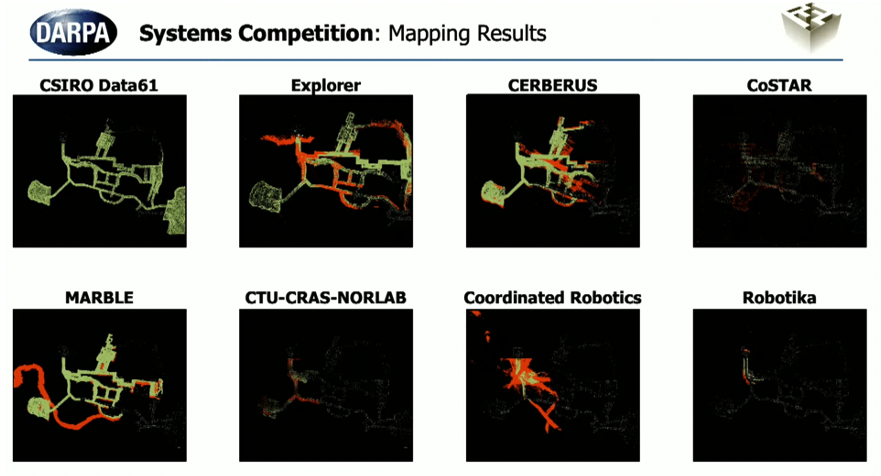}
    \caption{Visual comparison of online mapping data reported to DARPA during the competition runs for the eight finalist teams during the Final Prize Run,$^*$ with points matching ground truth shown in green, and non-matching points shown in orange. \\ $^*$\url{https://youtu.be/SyjeIGCHnrU?t=1676}}
    \label{fig:prizerun_mapping_comparison}
\end{figure}

Figure~\ref{fig:prizerun_multiagent_offline} shows the map created collaboratively by our robots by the end of the prize run. According to DARPA, this map has ``0\% deviation'' from the surveyed ground truth where ``deviation percentage'' is defined as the percentage of points that are further than one meter from the surveyed point cloud.\footnote{\url{https://www.youtube.com/watch?v=SyjeIGCHnrU&t=1932s}} \figref{fig:prizerun_mapping_comparison} shows the maps created by all teams by the end of the prize run. Green (resp., orange) points correspond to map points whose distance from the surveyed point cloud is less (resp., more) than one meter. Our team produced the most accurate map at the Final Event, while also having ``91\% coverage'' according to DARPA. We conducted our own quantitative analysis using the surveyed point cloud map provided by DARPA in \cite{ramezani2022wildcat}; the results show that the average distance between our map points and the nearest point in the reference map (after aligning the two maps) is about 3\,cm.

\begin{figure}[ht]
    \centering
    \includegraphics[width=120mm]{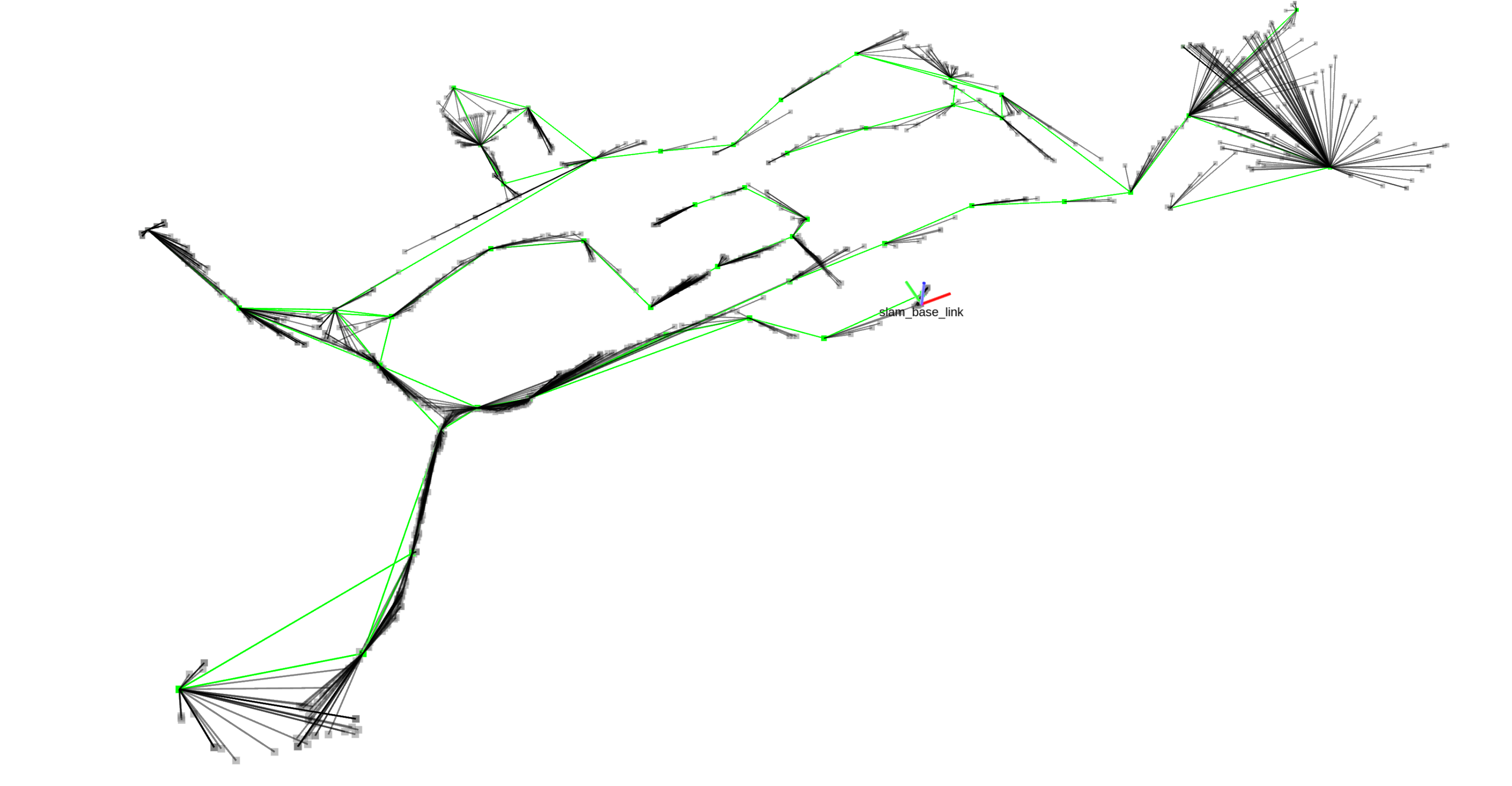}
    \caption{Pose-graph collected from four agents on one of the agents whose location is indicated by the red and green coordinate frame, during frame sharing. The white lines represent the connectivity between the children frames whereas the green lines indicate the connectivity between root nodes.}
    \label{fig:ma_pose_graph}
\end{figure}

Moreover,~\figref{fig:ma_pose_graph} depicts the collective pose graph based on one of the agents during the Prize Run. In total, 3950 frames (grey nodes) were shared between four agents out of which only 49 nodes were considered as root nodes (green nodes) whose poses were estimated through the pose graph optimisation. If frames sufficiently overlapped with a root node, they are considered as child nodes and they are rigid relative to the root node. This strategy decrease the number of root nodes as a result the number of parameters in the pose graph optimisation allowing Wildcat to deal with scalability in an efficient manner.

\section{Autonomy}
\label{sec:autonomy}
Autonomy is critical in SubT due to the dual limitations of a single human supervisor, and the communications challenges of underground environments. This section describes the solution utilised for this. For UGVs, we first describe the local autonomy in \secref{sec:ugv_local_autonomy}, then the global autonomy in \secref{sec:ugv_global_autonomy}, and finally the multi-robot task allocation in \secref{sec:MRTA}. Subsequently, we describe the UAV autonomy in section \secref{sec:uav_autonomy}.

\begin{figure}
    \centering
    \subfloat[]{\includegraphics[width=0.9\textwidth]{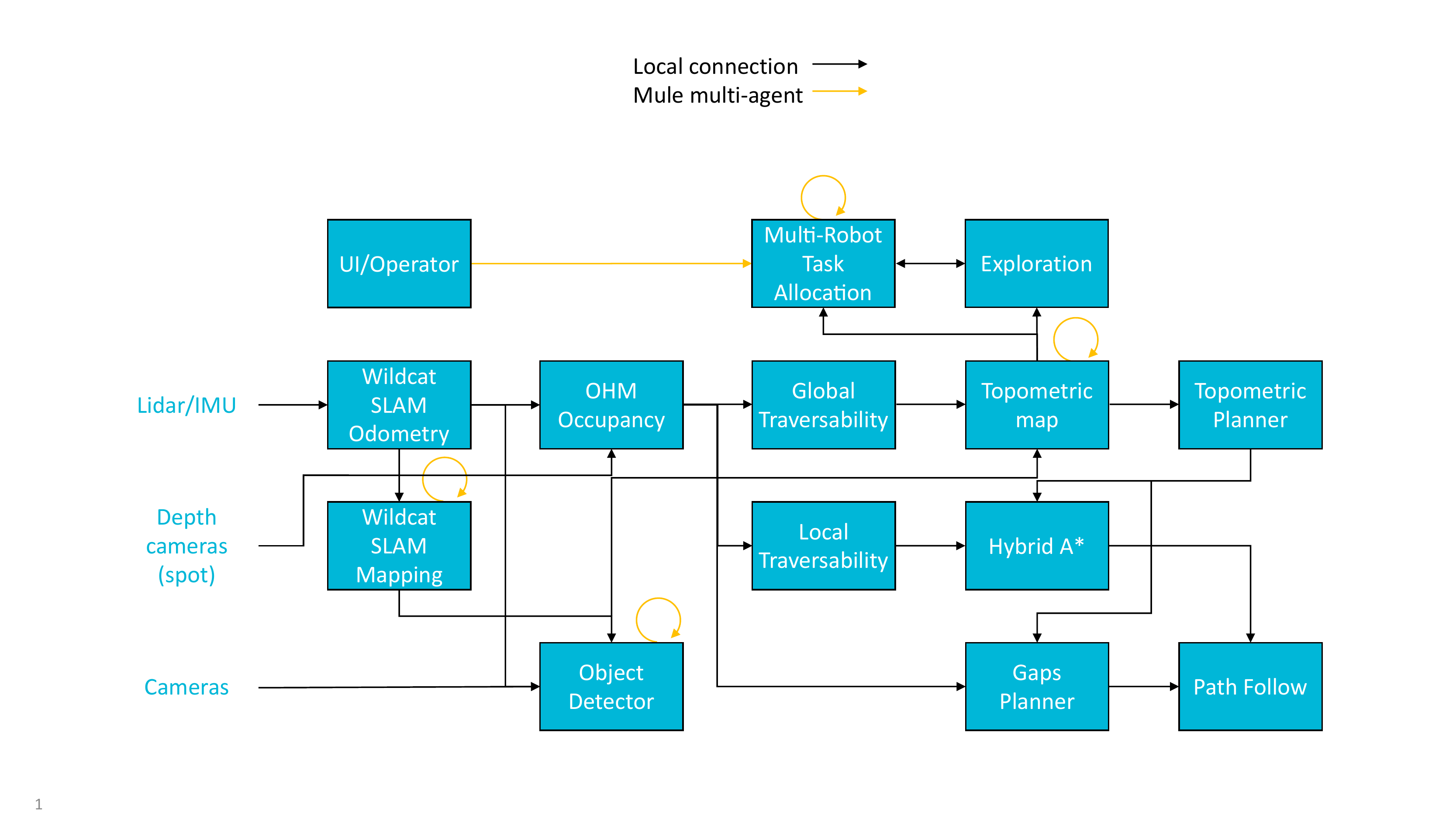}}
    \hfill
    \subfloat[]{\includegraphics[width=0.55\textwidth]{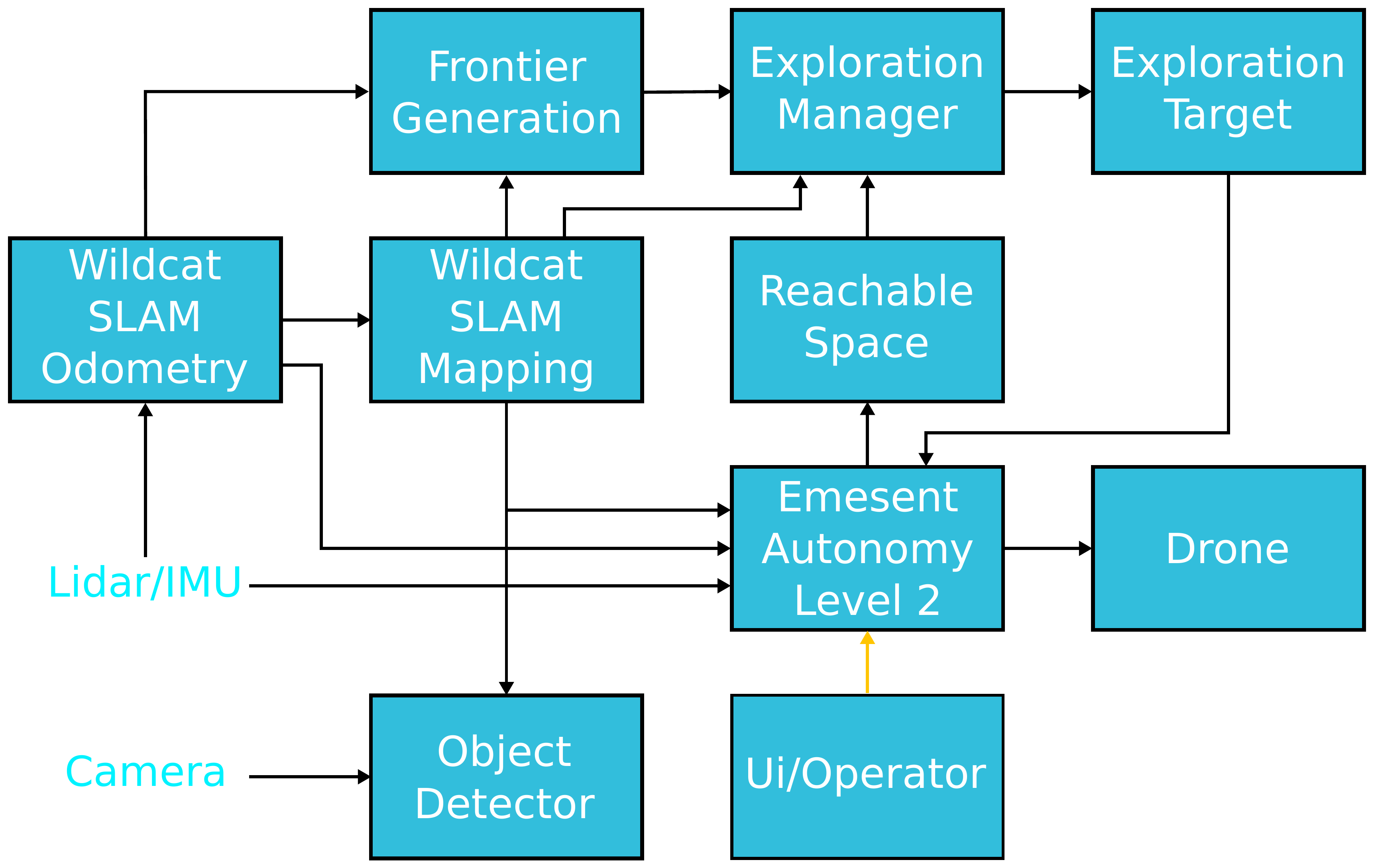}}

    \caption{Block diagram of autonomy system, (a) for the UGV, and (b) for the UAV.}
    \label{fig:autonomy_block_diagrams}
\end{figure}

Block diagrams of the respective UGV and UAV autonomy systems are show in \figref{fig:autonomy_block_diagrams}. The key component in common between the two is the Wildcat SLAM system, described in \secref{sec:wildcat}.

\subsection{UGV Local Autonomy}
\label{sec:ugv_local_autonomy}

The Wildcat SLAM system, described in \secref{sec:wildcat}, is a key underpinning technology to the solution. Local navigation primarily utilises the point cloud provided by the Wildcat odometry process. This is integrated into a 3D GPU-based occupancy map through the GPU-based Occupancy Homogeneous Map (OHM) system, detailed in \cite{stepanas_ohm_2022}. The occupancy grid is populated to a minimum range of 10\,m (with discrete jumps due to region-based addressing), and a resolution of 0.1\,m. Height maps are extracted from the OHM grid. Support was included for multiple vertical layers in height maps, e.g., to support navigation up a staircase in an area where the region beneath the staircase is also visible, but this remained to be fully utilised in the downstream parts of the stack at the time of the competition. The height maps include awareness of the clearance height required by the respective platform, and so will not output a ground level beneath an overhang with insufficient clearance. 

\begin{figure}[!tb]
		\centering
		\subfloat[]{\includegraphics[height=40mm]{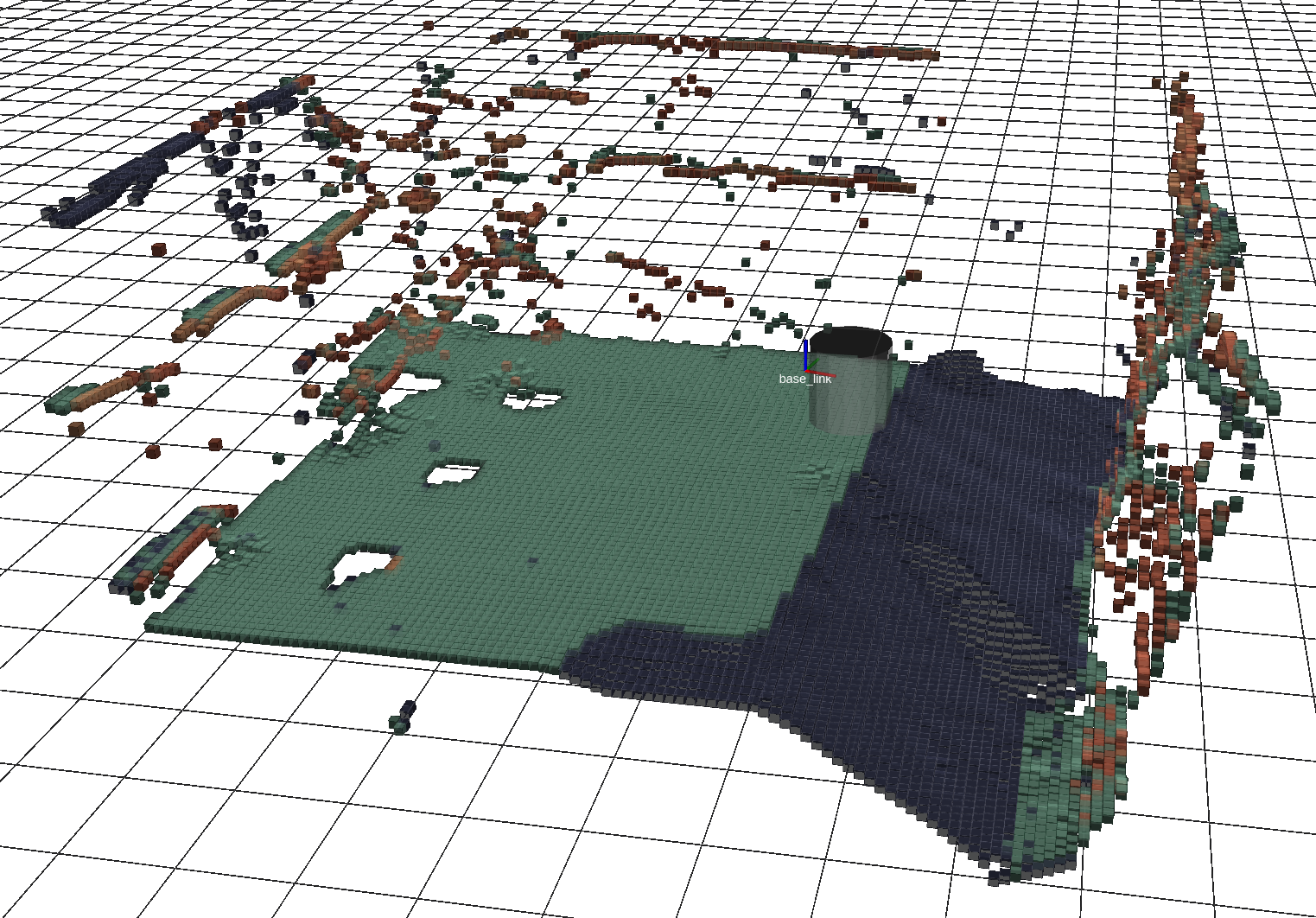}}~
		\subfloat[]{\includegraphics[height=40mm]{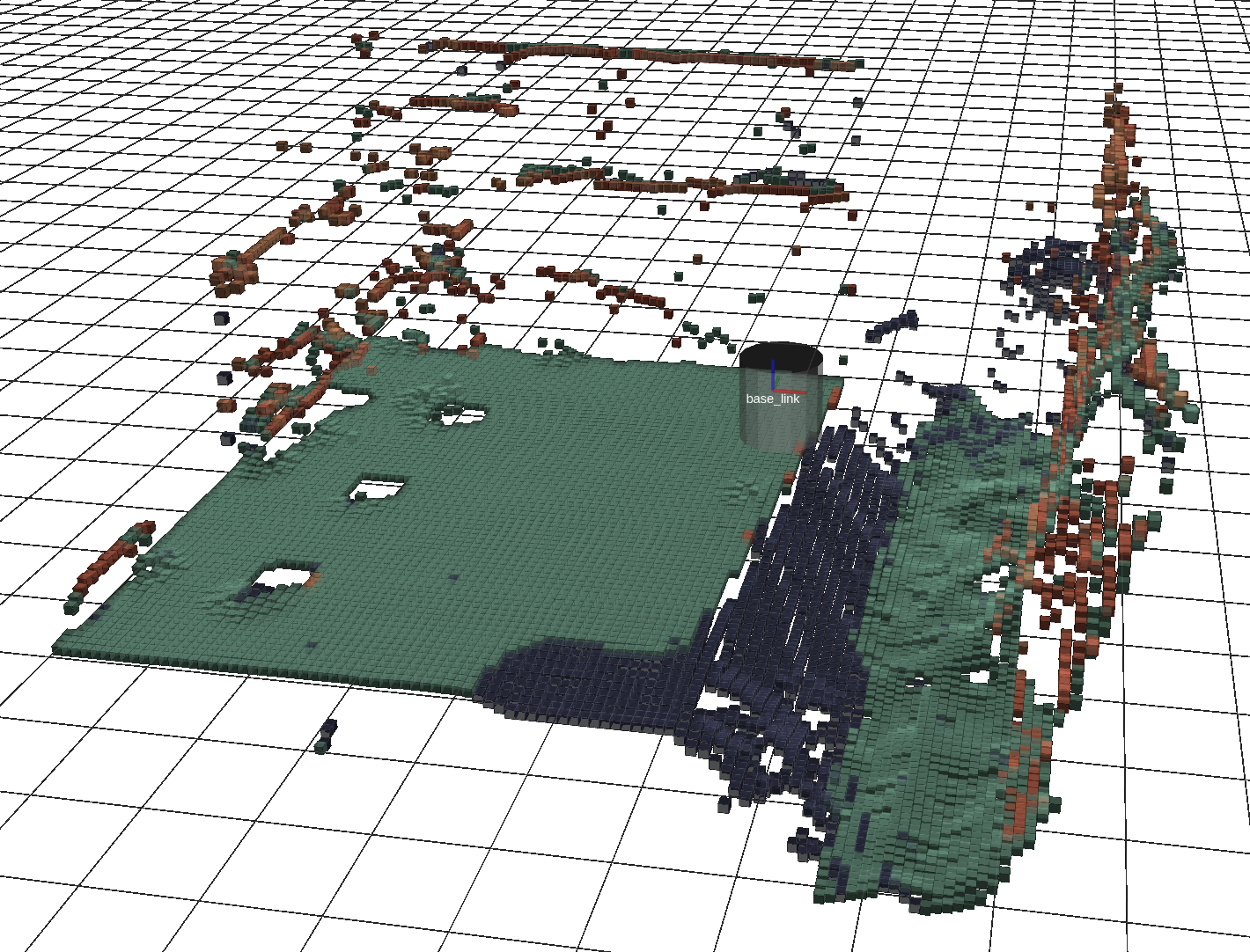}}\\
		\subfloat[]{\includegraphics[height=40mm]{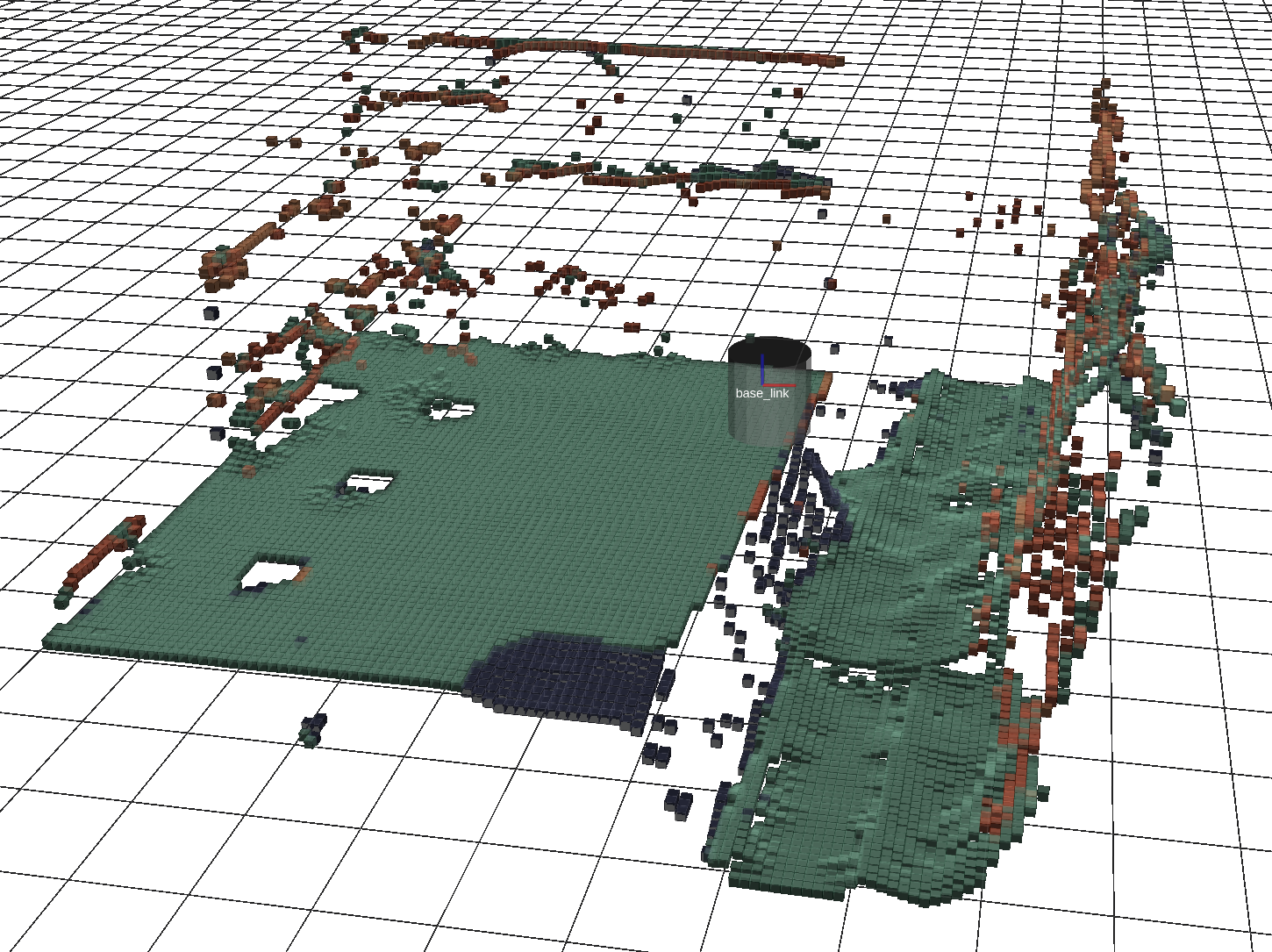}}~
		\subfloat[]{\includegraphics[height=40mm]{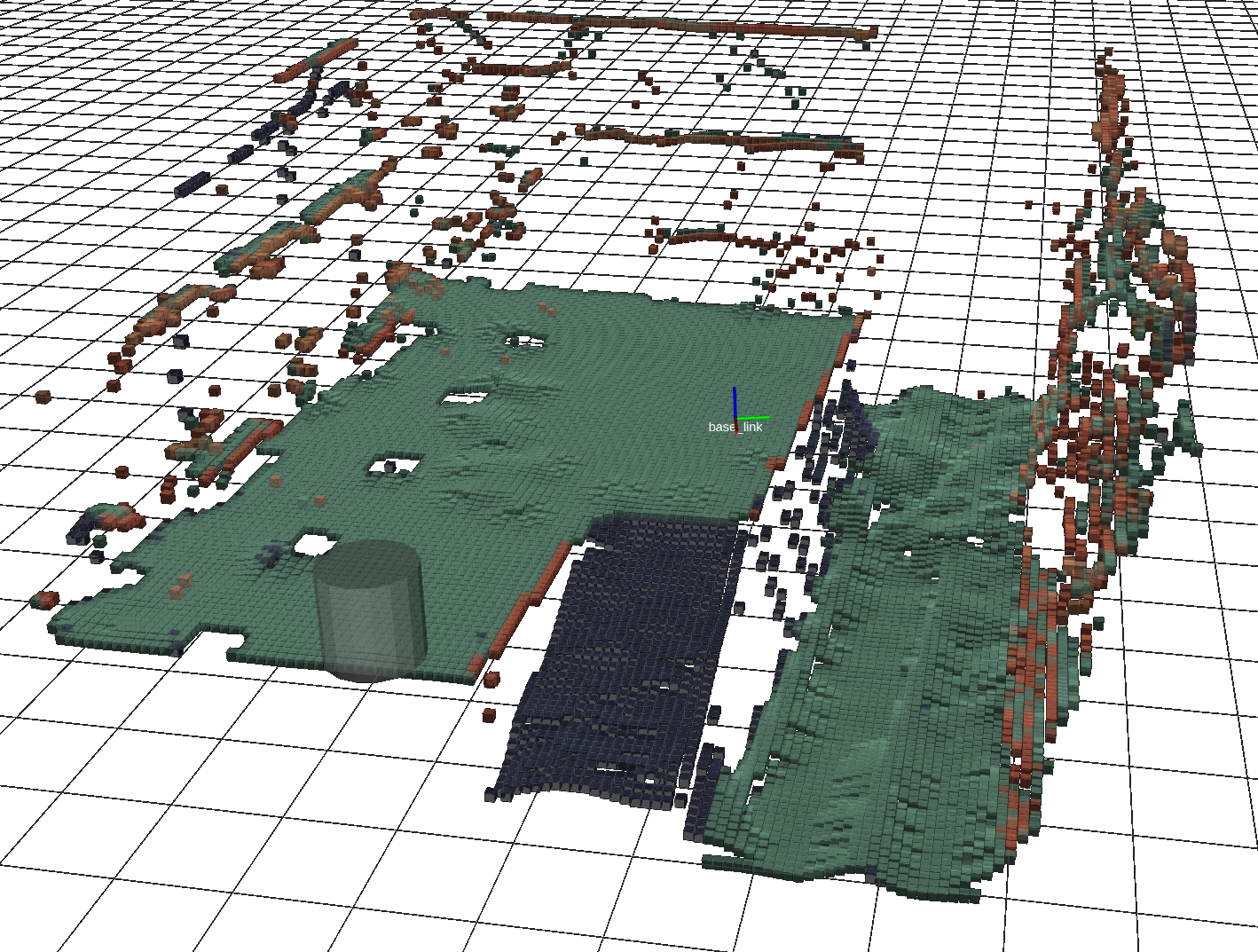}}\\
		\subfloat[]{\includegraphics[height=40mm]{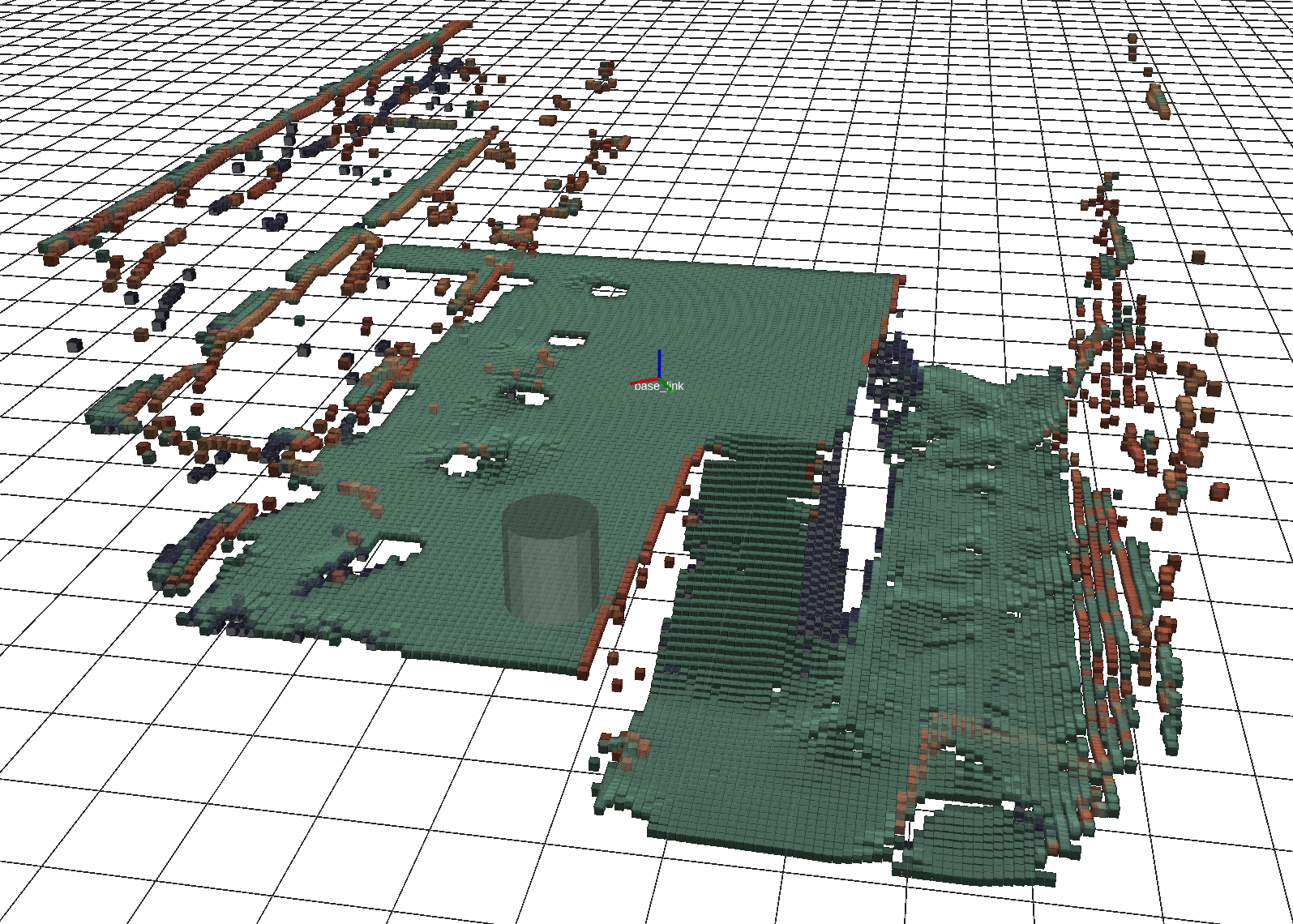}}
		\caption{Virtual surface processing during final circuit event: (a) A Spot robot autonomously traverses the subway platform, showing a virtual surface falling off the edge of the platform down to the tunnel. (b) As the edge is approached, the virtual surface becomes steeper. (c) This progresses until the steepness reveals the fatal cost at the edge of the obstacle. (d) The robot moves towards the front part of the platform, identifying a virtual surface above the stairs leading to the base of the tunnel. (e) The stairs are observed, revealing traversable terrain leading to the tunnel base.}
		\label{fig:virtual_surface_railway}
	\end{figure}

As described in \cite{hines_2020}, an important feature of the height map generation is identification of \textit{virtual surfaces}. These represent horizontal frontiers, where the space above has been identified as free space, but the space below is unknown. Consequently, they are surfaces that have not been directly observed, but observed data implies the potential presence of a surface beneath. The navigation stack utilises this by initially treating them as traversable, so that the agent will move toward them and either observe the surface itself, or identify the fatal incline descending from the top edge, as illustrated in \figref{fig:virtual_surface_railway}.

Traversability is assessed using the height map at two different ranges and rates. Local traversability analyses terrain at a shorter range (4\,m) but at a higher rate (5\,Hz), whereas global traversability analyses terrain at a longer range (6\,m) but at a lower rate (1\,Hz). The traversability analysis consists of tests on slope and steps. The slope analysis considers hypothetical (circular) robot footprints at each location in the map, and evaluates the resulting attitude. The cell is considered fatal if the slope exceeds a threshold. The step analysis searches for discrete steps exceeding a threshold in small local regions around each cell. The result of the traversability analysis is a classification of unknown (unobserved height), traversable, or fatal, along with the classification of observed or virtual provided with the height map.

The main planner utilised is hybrid A*, as detailed in \cite{hines_2020}. Planning is conducted on a graph where nodes are positioned on a 3D grid with the same 0.1\,m spacing utilised in the height map, and 30\degree in yaw. Edges are derived from motion primitives and costed dynamically as they are visited by A*. Costs were tuned to provide the desired balance between longer paths with gentler slopes and shorter paths with steeper slopes.

\subsubsection{Path follow}
\label{sss:path_follow}
In order to improve robot navigation performance and efficiency, and prolong the lifespan of robot hardware, improvements to the velocity command generation were deemed necessary, leading to the development of a new trajectory generation behaviour.

The new ``Path Follow'' behaviour was developed with three key requirements in mind:
\begin{itemize}
  \item Generated trajectories must strictly adhere to set velocity and acceleration limits
  \item Generated velocity commands must be continuous and smooth
  \item Overall performance must match or exceed previous systems
\end{itemize}

The new Path Follow behaviour meets these requirements through the use of actively updated 10th order B\'{e}zier curves to generate trajectories from the robot base link to a local goal pose. Trajectory generation is optimised such that the form of the trajectory fits the input path as close as possible whilst also adhering to the kinematic constraints of the robot and minimising trajectory completion time.

The Path Follow behaviour makes use of a dynamic short horizon envelope which reaches out from the base link of the robot platform out to the nearest obstacle, defined as at least one fatal cell in the cost map. The horizon defines the pose of the local goal, used as the target for all trajectory generation. The local goal is located on the local path at its intersection with the horizon. The horizon shrinks and grows as the robot moves closer to or further from obstacles and as it does, Path Follow dynamically adjusts the scaling of the robot kinematic limits thus slowing the robot as it approaches obstacles. This slowing around obstacles allows for more precise path tracking around obstacles, and is essential for narrow gap navigation.

A key component of Path Follow is the generation of smooth and continuous trajectories and velocity commands. Newly generated trajectories are generated with specifically set control nodes such that the initial state of the trajectory matches the last robot command. This ensures a smooth continuous kinematic command profile sent to the robot platform whilst constantly regenerating trajectories, and is essential for accurate path tracking.

Active trajectory regeneration is attempted at a rate of 25\,Hz. If a new active trajectory is generated, robot commands are set as $R=T(\Delta t)$ where $R$ is the generated set of robot velocity commands in a twist format and $T$ is the trajectory function with $t$ time as input. If a new trajectory cannot be generated which fits current kinematic constraints the existing active trajectory is iterated and robot commands set as $R=T(t+\Delta t)$. In the rare occurrence that trajectory regeneration fails continuously and a full trajectory is executed, a safety mechanism is built into every generated trajectory to ensure a safe stopping procedure is commanded at the end of each trajectory.

Path Follow works in conjunction with the rest of the behaviour stack and as such has been tuned to be quite aggressive. Path Follow will attempt to follow any given path as best it can regardless of whether the route of the path may result in collision with obstacles. Instead Path Follow will merely slow the robot down to a minimum speed, continue to follow the path and rely on other recovery behaviours such as the Decollide and Orientation Correction behaviours to recover the robot from any potentially dangerous collision state (described under Other behaviours below).

\subsubsection{Gaps planner}
\label{sss:gaps_planner}
The 0.1\,m discretisation resolution utilised in the hybrid A* planner is a significant limitation when seeking to pass through 800\,mm doors with an ATR that is 780\,mm wide.\footnote{The smaller width of the Spot robots was such that difficulties were not encountered and this behaviour was not required for that platform.} The gaps planner was developed to handle this type of case.

Initially, the development focused on reinforcement learning (RL) approaches. The method in \cite{tidd_passing_2021} was successful 93\% of the time in simulation, and 73\% of the time in on-robot tests. Despite these promising results, in the context of SubT where the a priori unseen environment is a large part of the challenge, a conventional planning approach was preferred. The finals course demonstrated the motivation for this decision: The course contained many tunnels that were far narrower than anticipated or previously encountered. The hand-engineered gaps planner adapted well to this type of environment, whereas the RL-based method that was not trained on data similar to this would not have been expected to generalise well.

The gaps planner can be seen as an extension of hybrid A*, which integrates a continuous optimisation step into the search to allow fine repositioning of search nodes within their respective discretisation cells. The search proceeds similarly to regular A*, but when cells are visited, the optimisation step seeks to improve the continuous position and yaw of the node based on the predecessor node and the nearby obstacles. Critically, this allows search nodes to be reliably found when passing through tunnels and doorways that were within one cell's dimensions of the vehicle size (as with hybrid A*, the position discretisation was 0.1\,m, and the yaw discretisation was 30\degree).

In addition to the improved planner, it was also critical to raise the platform's capability to accurately execute the plans. The combination of the hardware upgrades described in \secref{sss:atr} and 
the development in \secref{sss:path_follow} provided the required enhancement. Particularly important was Path Follow's adaptive speed control, which reduces velocity based on the distance to an obstacle, allowing for the most accurate control in critical circumstances.

\subsubsection{Other behaviours}
\label{sss:other_behaviours}

Specialised recovery behaviours were developed that have priority in scenarios that may place the robot in an unrecoverable state. The Orientation Correction behaviour activates if the pitch or roll of the robot may lead to the agent tipping over. The Decollide behaviour moves the robot to a nearby non-fatal region, recovering from the case where fatal cost appears within the footprint of the first search node preventing a valid path from being generated. 

For the Urban Circuit, a dedicated stair climb behaviour was developed for the Superdroid tracked platforms. This behaviour utilised extension of the robot flipper arms to extend the functional base of the robot, lengthening the lever arm required to tip the agent and preventing toppling down the stairs during ascension or descent. This behaviour could be manually activated by the operator or set to automatically activate when the agent orientation exceeded a pitch threshold. This allowed the operator to teleoperate the agent onto stairs at which point the behaviour would take over and complete the rest of the traversal in a safe manner (demonstrated during the Urban Circuit). Work on autonomous stair detection was deprioritised along with the Superdroids after the addition of the Boston Dynamics Spot robots to the team. Not only were the Spot's much more agile over a wider range of terrain conditions, their ability to handle stairs was superior to the tracked platforms that are unable to perform complex maneuvers on stairs (for example stopping and turning). Details on stair climbing with Spot are discussed in the Spot integration section below.

\subsubsection{Spot integration}
\label{sss:spot_local_nav}
The Spot platform was a late inclusion in the team, with integration work commencing in April 2021. The platform quickly evolved into a highly capable team member which provided unique capabilities traversing stairs and tight passages.

We found it critical to integrate Spot's internal cameras in order to address stairs and negative obstacles, since the location of our pack (see \figref{fig:annotated_robot_platforms}) provided limited visibility at steep elevation angles in front of the robot, and no visibility behind (note that the Spot robot must reverse down stairs due to its leg design). Spot's internal navigation capabilities presented many choices for integration with the autonomy stack. Spot internally generates a high resolution (0.03\,m) height map to a range of 1.9\,m based on its five depth cameras. Initially, this height map was blended with the lidar height map generated by OHM, and the plan produced by hybrid A* was truncated at the point where it departed the local map. However, this resulted in two problems, firstly relating to the plans themselves, and secondly relating to the quality of the height map.

The difficulty encountered with the first approach for path planning was that Spot's internal planner appeared to generate plans that approximated a slerp (Spherical Linear Interpolation) of the current pose and the provided goal pose, rather than constructing plans that are logically consistent with the environment. This was acceptable in more open environments, but sometimes prevented navigation in narrow corridors. Consequently, the local plan produced by hybrid A* was passed to Spot as a timed trajectory; this improved performance significantly and the trajectory was generally well-followed.

The difficulty with the height map was found when developing the capability to autonomously navigate stairs. Specifically, the difficulty occurred when the Spot commenced from the top of a platform from which stairs descended, but was positioned such that the stairs were not visible, but part of the ground plane at the bottom of the stairs was visible. In this circumstance, rather than marking the unobserved region as unknown, the processing hallucinated the continuation of the ground plane up to the edge of the top platform, without stairs. This was indistinguishable from an instance where there was a genuine negative obstacle, and so would prevent the platform from moving toward the edge to look for stairs. 

In contrast, the OHM-based virtual surface processing would continue to approach the edge until either the stairs become visible, or the free space observed implies a non-traversable slope. To exploit this capability with the Spot, the native depth cameras were integrated into OHM. The front-left, front-right and rear cameras were integrated into OHM at 5\,Hz; the lidar provided adequate visibility on the sides and the side cameras were not found to be required. Due to the focus on supplementing lidar coverage at close ranges, camera rays were truncated to 2\,m, and the resolution was decimated to match the height map resolution at that range. Online processing of this data stream and generation of virtual surfaces was possible with the GPU-based OHM implementation on the Intel NUC's integrated graphics.

Spot provides gaits for walking, crawling and stairs. The stair gait is documented as slowing speed and pitching the body to observe stairs; testing appeared to indicate that it also conditions the robot to expect flat foot holds. Difficulties with autonomous identification of stairs due to the poor visibility from common viewing angles led to an approach which utilised the walk gait exclusively. This was partly motivated by testing which showed that stair traversal in the walk gait was generally acceptable, whereas traversal of slopes or rough terrain of similar pitch in the stair gait was usually catastrophic. Because the spot leg design necessitates backward descent of stairs, the path cost was tuned to penalise forward motion down greater than a given angle, resulting in the desired effect. As discussed in \secref{ss:FinalPrizeRun}, this approach was successful, and stairs were autonomously ascended and descended by Spot robots out of communications range during the Final Event. It is possible that improved stability could be obtained by switching to the crawl gait on rough terrain; experimentation with this concept is the subject of future work.

Typical battery life for the Spot was 40-45\,min, which is insufficient to last an entire run. For this reason, a ``battery return'' behaviour was incorporated, which forced the robot to navigate back to the base area when the battery percentage hit a critical threshold.

\subsection{UGV Global Autonomy}
\label{sec:ugv_global_autonomy}
The role of the global navigation system is to build an expanding map of the traversable terrain observed by all UGV agents, allowing any agent to navigate to any point observed by itself or any other agent. This is made possible by the PGO-based SLAM system described in \secref{sec:wildcat}. Separate traversability maps (submaps) are generated for each root node in the SLAM graph, and like the SLAM frames themselves, traversability submaps are shared between agents. 

As described previously, global autonomy utilises the same traversability analysis as local navigation, but with maps generated with a longer range and at a lower rate. Subsequent observations which fall within the time range of the same SLAM root node are merged into the same image representation. Data with mismatching heights are handled by incorporating additional layers in the submap. The maps incorporate data for observed traversable and lethal surfaces; virtual surfaces are not incorporated into the map. Height maps and cost maps are also shared between agents; the data rate involved with these representations was found to be an order of magnitude lower than the SLAM frames. Compact graph representations of submaps are obtained by applying superpixel methods to the images, incorporating channels for the fatal traversability signal and height. Connections between submaps and submap layers are identified by finding superpixels which overlap between them. Global path planning is conducted utilising A* on the graph with nodes corresponding to superpixels. Edge costs are obtained through distances, with additional penalties based on slope and roughness (averaged over the superpixel for each node), which are designed to match the penalties used in local path planning.

Dynamic obstacles present a particular challenge as paths previously observed as traversable must be updated to reflect the path that has been closed. The local navigation stack was capable of traversing extremely difficult terrain, but this sometimes took several attempts. It was important for the global map update process to be tuned accordingly. It was also made difficult because both local and global navigation were based on height maps, and thus were unable to distinguish between a changed environment and difference in observation perspective of the same, unchanged 3D structure. For example, from one perspective, we may observe the underside of a ramp crossing to an upper level, and declare an obstacle only where the clearance becomes insufficient. From another perspective, we may observe the top of the ramp, continuing up to the next level. 

Because of this difficulty, the approach operated directly on the global superpixel graph. Specifically, whenever navigation failures occurred (i.e., timeouts without significant progress towards the goal), the source and destination nodes locations in the global graph were stored, and edges between superpixels containing those two positions were subsequently suppressed. The exception to this was edges which had previously been traversed by an agent. Because the cost of falsely mistaking the path home was high (e.g., preventing robots from returning to synchronise data), rather than suppressing these edges altogether, instead a high traversal cost was applied to them. Accordingly, any path not utilising that edge would be preferred. In the case where the edge is still utilised, traversal failures will trigger task failures as described in \secref{sec:MRTA}, which will in turn trigger selection of exploration tasks that have potential to discover the necessary alternative routes. Overall, this approach was effective, but could take some time to resolve the correct map.

Whereas UGV exploration in earlier stages of the program was based on the 3D point cloud visibility work in \cite{williams2020exploration}, in finals we utilised traversability frontiers, exploring to the boundary between observed traversable and unknown space. Thus frontiers were attached to superpixels which bordered unknown space, performing a natural clustering of frontier pixels. Again, due to the multi-agent global navigation representation, a region will only be marked as a frontier if it has not been observed by any UGV.

Connections are only made between submaps that are within a local neighbourhood in the SLAM graph. Accordingly, if a region is revisited but loop closure has not occurred, the previously visited area will be explored as if it is being observed for the first time. This is a desirable behaviour, as this additional exploration provides the data required for loop closure to occur. The process could likely be made more rapid by explicitly reasoning over the exact data necessary for loop closure (e.g., active SLAM, \cite{placed2022survey}, executing actions aimed at collecting the data necessary for loop closure to occur).

\begin{figure}[t]
\centering
\includegraphics[width=0.8\columnwidth]{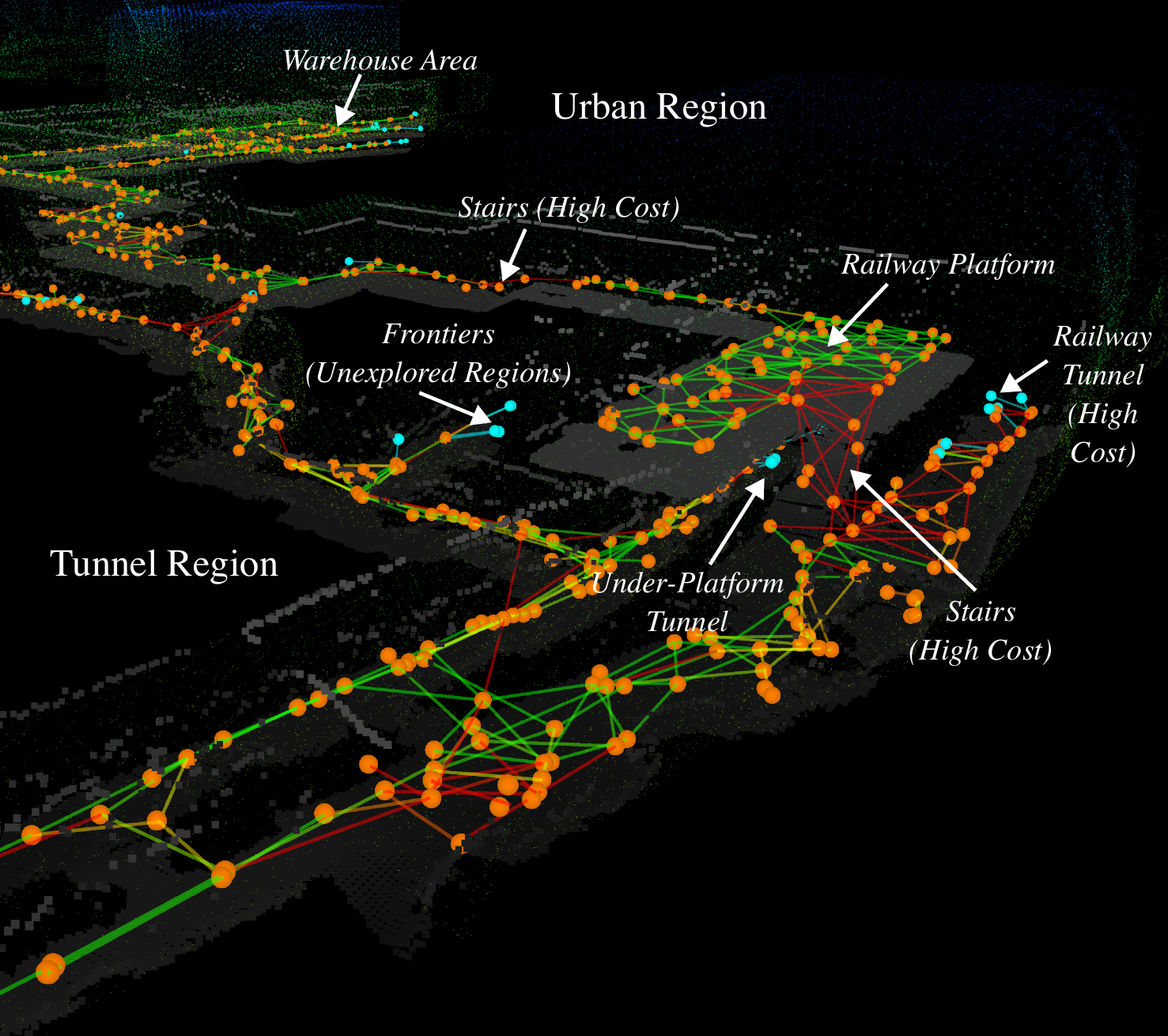}
\caption{Portion of the topometric (global) graph constructed during the Final Prize Run. The railway platform and stair portions correspond to the traversability maps illustrated in \figref{fig:virtual_surface_railway}.}
\label{fig:topo_graph_finals}
\end{figure}

Conceptually, the UAV could contribute to the UGV maps in the same way, but this was not exploited since the UAV does not run the same occupancy mapping, height mapping and traversability analysis pipeline, and the raw data is prohibitively large. %
Part of the global navigation graph from the Final Prize Run is shown in \figref{fig:topo_graph_finals}.

\subsection{Multi-Robot Task Allocation}
\label{sec:MRTA}
The multi-robot task allocation system provides the methodology for agents to autonomously and collaboratively agree on assignments of tasks to robots, based on a decentralised market-based process. This allows collaborative assignment of tasks to continue as the communications topology changes. For example, if two robots are deep in the course and have communication with each other but not with the base station, they can seamlessly continue negotiating task assignments.

Each agent is allocated a bundle of tasks. The metric which task allocation seeks to optimise is the total reward of the bundles, where rewards have an exponential time discount based on the expected completion time of the task. For example, if we denote the bundle for agent $a$ as $\vec{p}_a=(j_{a,1},...,j_{a,{n_a}})$, the reward of task $j$ as $c_j$, the traversal cost from task $j'$ to $j$ as $t(j',j)$, and the duration of task $j$ as $T(j)$, then the total reward is:
\begin{align}
S(\vec{p}) &= \sum_{a=1}^A\sum_{i=1}^{n_a} \lambda^{\tau^{a,i}(\vec{p}_a)}c_{j_{a,i}}
\intertext{where}
\tau^{a,i}(\vec{p}_a) &= \sum_{k=1}^i t(j_{a,k-1},j_{a,k}) + T(j_{a,k})
\end{align}
$\tau^{a,i}$ denotes the time when execution of the $i$-th task $j_{a,i}$ completes, and $t(j_{a,0},j_{a,1})$ denotes the cost of navigating from the agent's current position to the first task. The Consensus-Based Bundle Algorithm (CBBA) operates by building the bundle incrementally, bidding on the task which produces the largest increase in the bundle reward, where the reward increase is evaluated by inserting the task into the best position in the bundle.

The primary task utilised by the task allocator is exploration. For the exploration task, frontier nodes identified by global navigation are clustered to provide tasks consisting of frontiers that are nearby in terms of global path distance. Each agent independently generates its own set of frontiers and tasks based on traversability data received from all agents. This avoids the solution of a decentralised clustering problem (which would be greatly complicated by regular periods of disrupted communication). Duplicate assignments are avoided by treating nearby tasks belonging to different agents as equivalent for the purposes of bidding, so that assignment of an agent to a task requires outbidding another agent assigned to nearby tasks.

Tasks can be bid upon at any time, and new tasks continually arrive as agents move through the region. Each agent maintains its task bundle, which is limited to a maximum number of tasks, and a maximum expected duration. Because of the open-ended nature of exploration tasks, their duration is set such that two exploration tasks will not be added to the bundle at the same time. In order to allow the bundle to be adapted once it is full, we consider bidding actions which drop the final element from the bundle sequence and add a new element (into an optimised position, as is standard in the bidding process).

Methods were developed which estimated the reward of a frontier through bounds on the new volume that might be observed from the candidate pose. However, this was found to be counter-productive, since features such as narrow tunnels and tight doorways were of critical importance. Thus, without cognisance of this higher level semantic information, the standard reward estimates were found to be unhelpful, frequently triggering undesirable behaviour (e.g., stopping exploration of a region of interest in favour of a more open area). For this reason, exploration tasks have a fixed reward, such that selection is based purely on the path cost.

Motivated by \cite{brass2011multirobot} and by the tree-like structure of many of the environments of interest, we encourage the agents to separate by penalising the shared component of the path from each robot back to the base. In \cite{obrien_2022}, this is shown to improve performance in environments that are well approximated as trees.

As well as exploration tasks, the task allocator also supports ``return to synchronise'' tasks (i.e., return toward the base until all data is uploaded to and downloaded from the base), and ``drop node'' tasks. Synchronisation tasks benefit from multi-agent implementation so that if another agent is in a location that allows it to return more quickly and it has all of the data from the agent that needs to synchronise, it can bid on the task. Drop node tasks are manually generated, and can be executed by any agent that has a communications node available. Logic was developed to automatically deploy communications nodes based on signal strength, but it was not deployed in competition due to the risk of rendering a narrow tunnel impassable. Further details of the task allocation method can be found in \cite{obrien_2022}.

\subsection{UAV Autonomy}
\label{sec:uav_autonomy}
The UAV autonomy used in the final competition broadly follows the autonomy used in previous competitions, with improvements focused on exploration. The UAV utilises the navigation functionality commercially offered by Emesent as Autonomy Level 2, providing both local and global navigation solutions.\footnote{Emesent AL 2 (2021). https://www.emesent.io/autonomy-level-2.} The UAV features a manager node to coordinate local and global planning, and implement core behavioural primitives. The manager receives higher level tasks from the operator, such as move to these waypoints, explore, etc. It coordinates activities to achieve these tasks, and interfaces with the lower level autonomy. This modular architecture has simplified the process of continual improvement to our higher level autonomy functionality. 
To ease operator load, the UAV supports four major control modes: exploration, 3D waypoints, 2D waypoints, and planar waypoints. 3D waypoints specify an exact position in space where the UAV must move, useful to get the UAV into narrow openings when other alternatives are available (i.e., moving into a shed). 2D waypoints specify a location in the horizontal plane, but leave height free, which is useful for general purpose commands. Planar waypoints simply require that the UAV reach any point on a user-specified plane, but does not specify where. It is often selected as a vertical plane, in order to provide a direction of travel without the need for a precise goal. This is useful for sending the UAV to a general location (i.e., not just basic exploration) in a space whose approximate layout is very roughly known, e.g., go 100~m down a tunnel and then turn left at the branch. It can also serve as another form of more directed exploration.

\begin{figure}[t]
    \centering
    \includegraphics[width=100mm]{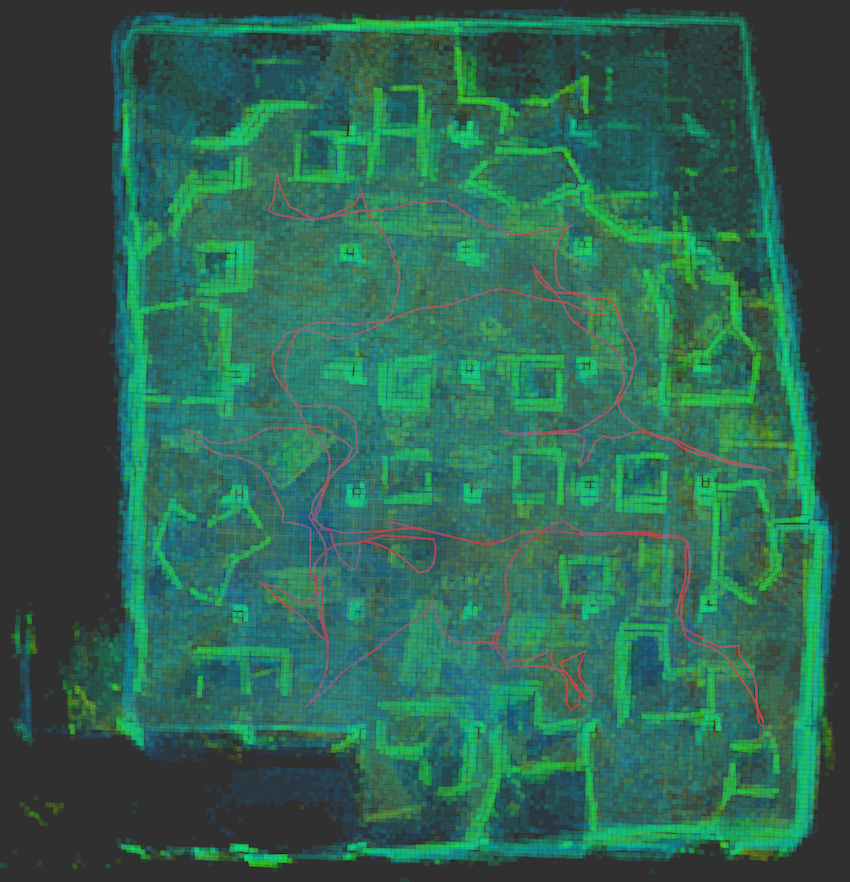}
    \caption{Trajectory of UAV (red line) performing exploration at an indoor paintball course prior to the final competition. Background colour shows height, illustrating the UAV's path through the maze-like course.}
    \label{fig:paintball_exploration}
\end{figure}

Frontier generation and selection broadly follows that described in \cite{williams2020exploration}, with frontier selection based on a scoring function balancing frontier size, proximity, and alignment with previous exploration. Improvements focused on dealing with invalid waypoints, motivated by experiments at an indoor paintball course. The paintball course featured many small windows, narrow doorways, and other non-traversable apertures which generated a large number of frontiers that were not reachable by the UAV, causing exploration to become stuck for significant periods of time. Mitigation efforts focused on utilising traversability information provided by the planner. The planner published its search tree after every planning iteration, which the exploration code used in two ways. The first was checking whether the planner was able to find a path sufficiently deep into a given frontier. If so, the frontier was marked as reachable, and its score for selection purposes was increased, strongly biasing exploration towards known-reachable frontiers. Furthermore, the closest point to the center of a reachable frontier was retained as a witness point. When a frontier was cleared, other frontiers that contained the witness point of the cleared frontier would be marked as reachable. This provided additional robustness when frontiers were modified at distances that exceeded the planning horizon, which was particularly important for large open spaces. An example exploration trajectory of the indoor paintball course is show in \figref{fig:paintball_exploration}.

Previously, a frontier was only marked as unreachable if the UAV had failed to make substantial progress towards a frontier for 20\,s. These delays caused by unreachable frontiers were reduced by identifying when the UAV was moving away from the target frontier, indicating that the frontier was unreachable. To do so, the exploration manager would compute the minimum distance from the search tree to the target frontier, and if below a threshold (3.2\,m) indicating that the UAV was in close proximity to the target frontier, it would compare the search tree distance to the distance from the end of the actual plan associated with the search tree  to the frontier. The path end distance being significantly larger than the search tree distance (2\,m) indicated that the path directly to the frontier was blocked, and the planner was attempting to find an alternate route. In this case, the exploration manager would mark the frontier as unreachable, and exclude it, and any other frontiers with centers inside the unreachable frontier, from consideration.

The previous iterations of the UAV exploration code ignored the global structure of the environment, using only the direct Euclidean distance between UAV and frontier when selecting frontiers, with obvious downsides in larger or more complex environments. For the final competition, the UAV exploration manager was enhanced to use the SLAM pose graph to reason about the global environment. When a frontier was created, it was associated with a frame in the pose graph, and the UAV was continuously associated with the currently active frame used by the SLAM module. Unfortunately, global SLAM optimisation had significant latency, often exceeding 10 sec, leading to poor associations. As a result, frontiers and the UAV were reassociated with the closest of their original frame and neighboring frames in the pose graph. When selecting a frontier, the UAV was limited to considering frontiers that were, after reassociation, associated with frames adjacent to the UAVs current frame in the pose graph. This limited consideration to frontiers that could be reasonably expected to be reachable by the local planner. If there were no frontiers in the adjacent frames, the UAV timed out in making progress towards a frontier, or the UAV marked its current frontier as unreachable, then the exploration manager would check if relocation was required. Relocation was required if there were no frontiers marked as reachable associated with the current UAV frame or adjacent frames. If relocation was required, the UAV chose the frame associated with the most, reachable frontiers, tiebreaking in favour of frames with more total associated frontiers. The exploration manager then used the SLAM pose graph to compute a path to the target frame, relocated there, then restarted exploration.

\section{Perception}
\label{sec:perception}
The Camera Based Perception framework used in the Final Event was similar to that used in the earlier circuit events, wherein artefacts detected in the camera stream by the DeNet object detector \citep{Tychsen_Smith_2017,Tychsen_Smith_2018} were localized in 3D using lidar depth measurements. Two major additions to the framework for the Final Event were the introduction of multi-agent artefact tracking and sending updated information of a previously detected artefact. 

The main purpose for the introduction of the multi-agent artefact tracking was to reduce the number of artefact reports seen by the operator. For the Urban Circuit, each agent tracked artefact locations in the odometry coordinate system and thus any drift in the odometry trajectory could possibly result in new detections being erroneously associated with known artefacts. These errors were mitigated at the Urban Circuit by removing known artefacts from the tracker memory when they had not been detected for 30\,s or the agent was more than 30m away from the artefact. Unfortunately, this meant that an agent would re-detect artefacts when visiting previously explored space. Maintaining the position of artefacts in the global map alleviates the issue of drift and thus artefacts can be remembered indefinitely. The similar problem of one or more agents visiting the same area and detecting the same artefacts can also be mitigated by using Wildcat's loop closure capability to establish correspondence between each agent's global map (see \secref{sec:wildcat} for more details).

The need to update information of a previously detected artefact was to address a common problem when using the artefact tracker to eliminate duplicate detections. In some instances, a true-positive would be correctly identified by the object detector but the operator was unable to confidently confirm the detection in the image because the artefact was either poorly illuminated or too small. For the Final Event, we introduced a policy where the operator GUI was updated with subsequent detections if the detection was at least 1\,m closer than a previously sent detection. The operator could then scroll through all of the artefact observations in order to more confidently confirm the detection. This functionality was easily implemented by assigning every tracked artefact a globally unique identifier and including the identifier in every sent artefact report.

The training dataset continued to expand after the Urban Circuit. The dataset used to train the DeNet detector model for the Final Event consisted of over 30,000 images of the nine artefacts at twelve different locations (21,000 annotated and 9,000 negative). The images were captured using a variety of mobile phones as well as the cameras used by the UGVs and UAVs.

WiFi and gas detections were presented to the user as a marker located at the position of the respective agent when detection occurred, and coloured by the strength of the detection (i.e., RSSI for WiFi, and detected density for gas). This provided the necessary information for the human supervisor to either detect and locate the respective artefact, or command a robot to collect additional information. As discussed in \secref{sec:comms_improvements}, the active scanning required for Bluetooth detection was found to cause significant communications disruption, so Bluetooth detection was disabled. Accordingly, we were reliant on  WiFi detection for cell phone artefacts (a visual detector was trained, but in practice due to the small size and indistinct appearance, it was rarely detected), and visual detection for the cube artefact.

\section{Human Robot Teaming}
\label{sec:human_robot_teaming}

The role of the human supervisor was to provide high level guidance to the robots while assessing incoming object detections and sending reports to the DARPA server. As the number of platforms increases, the bandwidth for a single human supervisor to manage individual agents becomes limited.

\begin{figure}
\includegraphics[trim=15cm 0cm 0cm 0cm, clip,width=\textwidth]{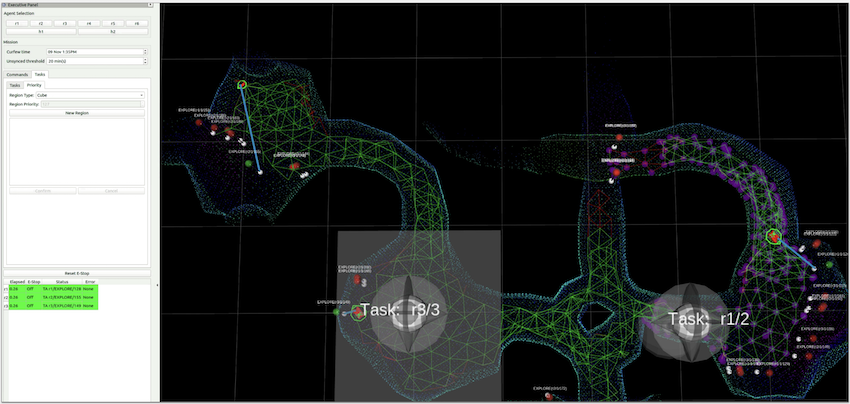}

\caption{Example of the map-based GUI in the operations user interface. Interactive ``lollipop'' markers enable the operator to manually assign or cancel tasks. The box in the lower left of the image defines a geometric prioritisation region ``Task: r3/3'' for robot \textbf{r3}, and ``Task: r1/2'' defines a graph-based priority region for robot \textbf{r1} (where the affected part of the graph is shown with purple shading). The octagon indicator surrounding the robot shows various aspects of status including SLAM, navigation, communications and tasking.}
\label{fig:gui}
\end{figure}

Previous experience showed the value of providing the human supervisor with full control range of the robots. While the dominant mode of operation was autonomous, fallback modes including waypoint navigation (or, more generally, missions consisting of scripted sequences of operations) and teleoperation proved useful in unexpected situations. With the focus on autonomous operation, interfaces were provided to permit entry of high-level guidance, in the form of prioritisation regions. These were specified geometrically, altering task priorities either within a region, or for any task downstream of the region in the shortest path tree commencing from the base location; examples of this are illustrated in \figref{fig:gui}. These latter graph-based priority regions were found to be a particularly valuable improvement, as they allowed prioritisation of an region of unknown shape and extent which lay beyond a junction.

A multimodal task-based graphical user interface (GUI) enabled the human supervisor to efficiently assess the status of each platform at a glance and provide mixed-level commands as needed (from teleoperation to fully autonomous exploration). \cite{chen_multimodal_2022} describes details of the user interface employed at the Final Event.

The operator interface was separated into two distinct windows: an artefact review window for efficiently assessing RGB images from detected objects and their localisation, and an operations window for interfacing with the robots. The example in \figref{fig:gui} shows the operations window for interfacing with three robots in autonomous exploration, illustrating prioritisation regions, and interactive task markers, which allow the operator to manually assign or cancel individual tasks. The human supervisor can assess several components of the robot health by glancing at the coloured octagonal ring around the robot markers, including the communication rate, percentage of data missing from the ground station, mission state, and any errors. The robot markers are a depiction of the robot type, and for the ATRs, display the number of remaining communications nodes and the UAV launch state.

The artefact review window presents the operator with both a list of objects and a map view showing their locations. The operator can quickly scan through new detections and either save or reject them. Gas and WiFi detections are illustrated by markers showing the detection location, with opacity indicating the concentration and signal strength respectively. In this case, the operator infers and indicates the source location on the map. 

\section{Results at the Final Circuit Event}
\label{sec:discussion}

The Final Event was held at the Louisville Mega Cavern in Kentucky, USA, on 21-24  September 2021, and consisted of two preliminary rounds (30\,min runs conducted on 21 and 22 September) and a final prize round (a 60\,min run conducted on 23 September). Team CSIRO Data61's objectives for the preliminary runs were to maximise information gained from the course and ensure each platform was fielded before the prize run while minimising risk to hardware. Without a full set of either human or robot team members (due to COVID travel restrictions), the robot roster for the three circuits faced in Louisville was carefully deliberated. Any post-run repairs could only be performed by a skeleton crew with over-the-shoulder support from experts in Australia using telepresence robots.

Based on extensive testing, the ATR robots (Rat and Bear) were trusted as exceptionally robust in a wide range of unstructured conditions, while the Spot platforms (Bluey and Bingo) had greater strengths in terrain coverage, and were particularly critical in urban areas (for example with narrow stairs). The aerial vehicles (H1 and H2) were reserved for open locations in the course with high ceilings. These preconceptions guided strategic planning for the preliminary runs. For each preliminary run, both ATRs would enter the course carrying deployable communications nodes, alongside a single Spot. For the first preliminary run a single UAV was carried by one of the ATRs, in the second preliminary run a UAV was attached to each ATR. This roster allowed for redundancy, in the event of misfortune with a Spot or UAV, there would be a set of platforms available for the Final Event. 

\begin{figure}[!b]
\centering
\subfloat[]{\includegraphics[height=3cm]{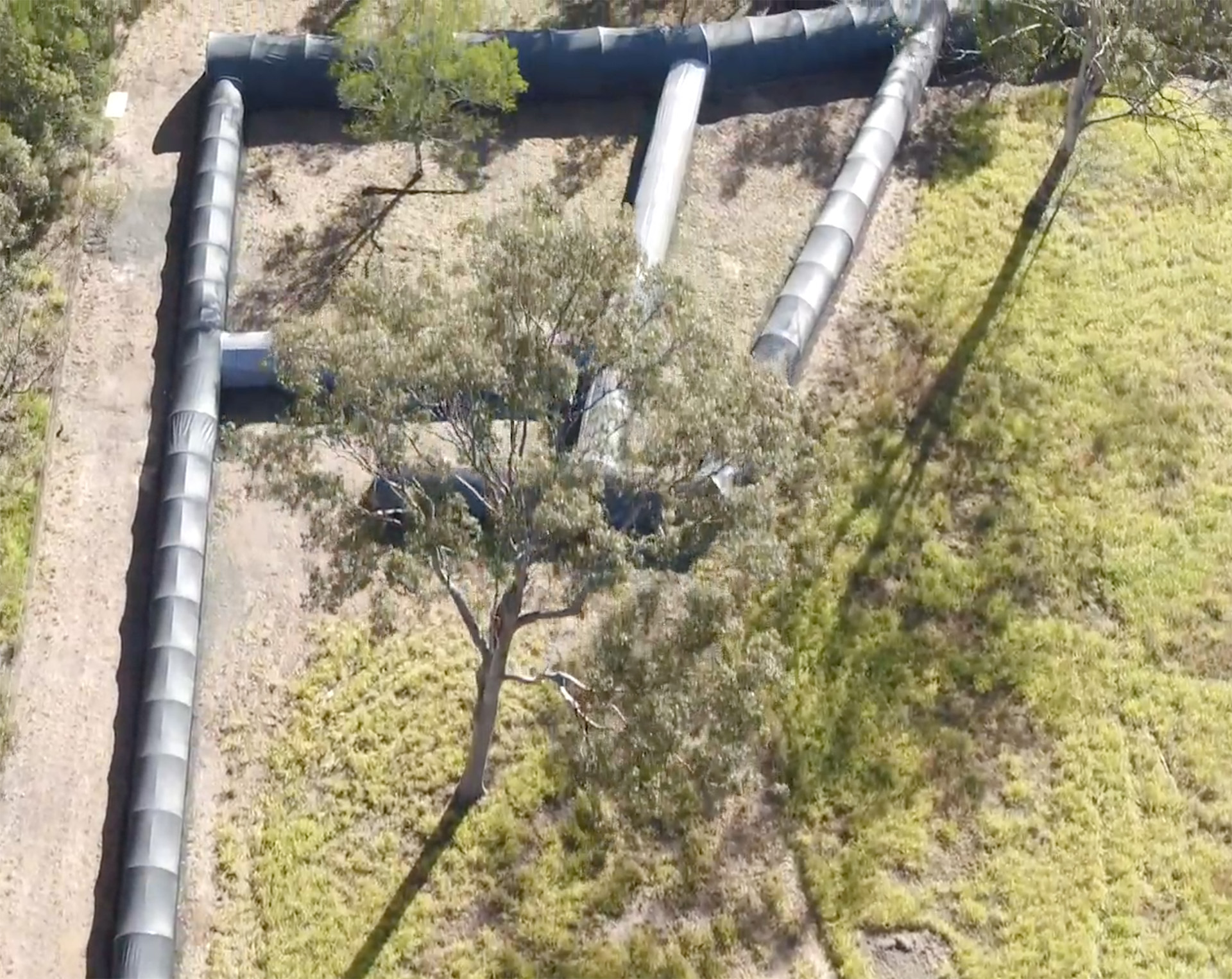}}~
\subfloat[]{\includegraphics[height=3cm]{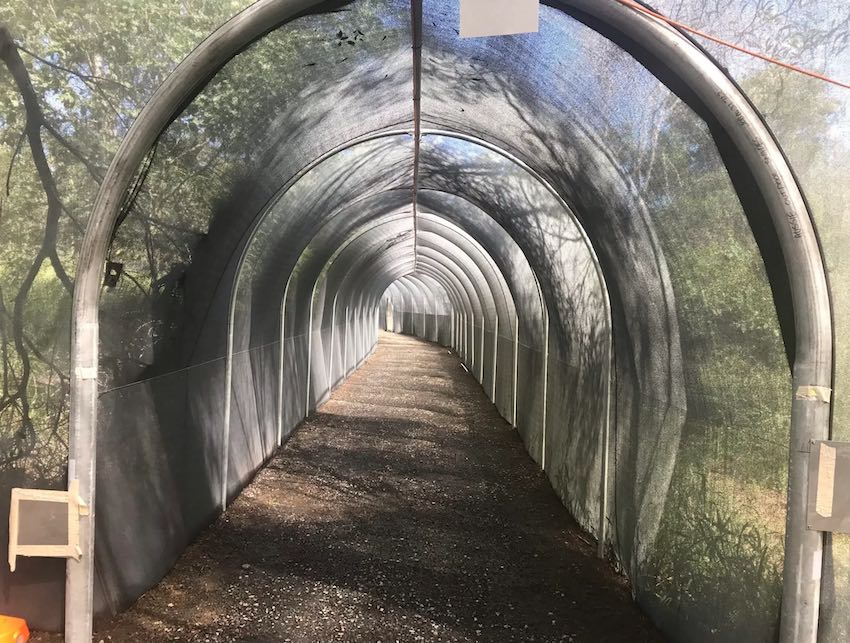}}~
\subfloat[]{\includegraphics[height=3cm]{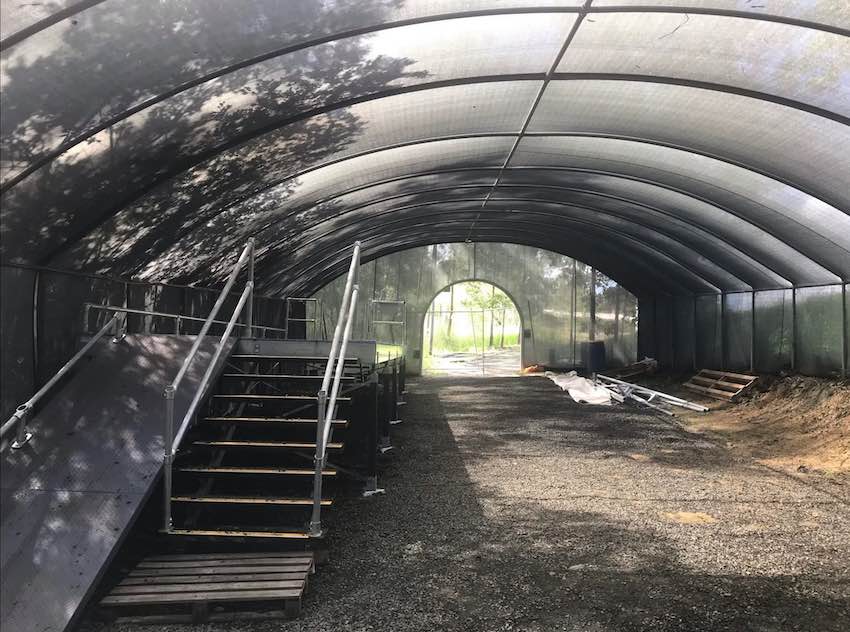}}

\caption{An aerial view of the shade-cloth tunnel with an overall length of over 300\,m built at CSIRO's QCAT site in Pullenvale, QLD, Australia (a), an inside view (b) and stairs and mezzanines built inside the tunnel (c).}
\label{fig:QCAT_tunnel}
\end{figure}

The Spot robots were generally more agile over a wider range of terrain conditions than the large ATRs. The strategy that evolved through testing was to send in a Spot robot first, followed by an ATR to bridge communications. Spot robots were able to autonomously traverse stairs, favouring urban settings, whereas the ATRs proved their robustness in challenging cave conditions. The confidence of the human supervisor in the abilities and limitations of each platform was cemented through extensive testing (based primarily on weekly test sessions at the CSIRO QCAT site, which incorporated urban industrial regions, a terrain park, and a synthetic tunnel environment shown in \figref{fig:QCAT_tunnel}).

\subsection{Preliminary Run 1}
\label{ss:Prelim1}

\begin{figure}[h!tb]
\centering
\subfloat[Robot Path]{\includegraphics[width=0.6\columnwidth]{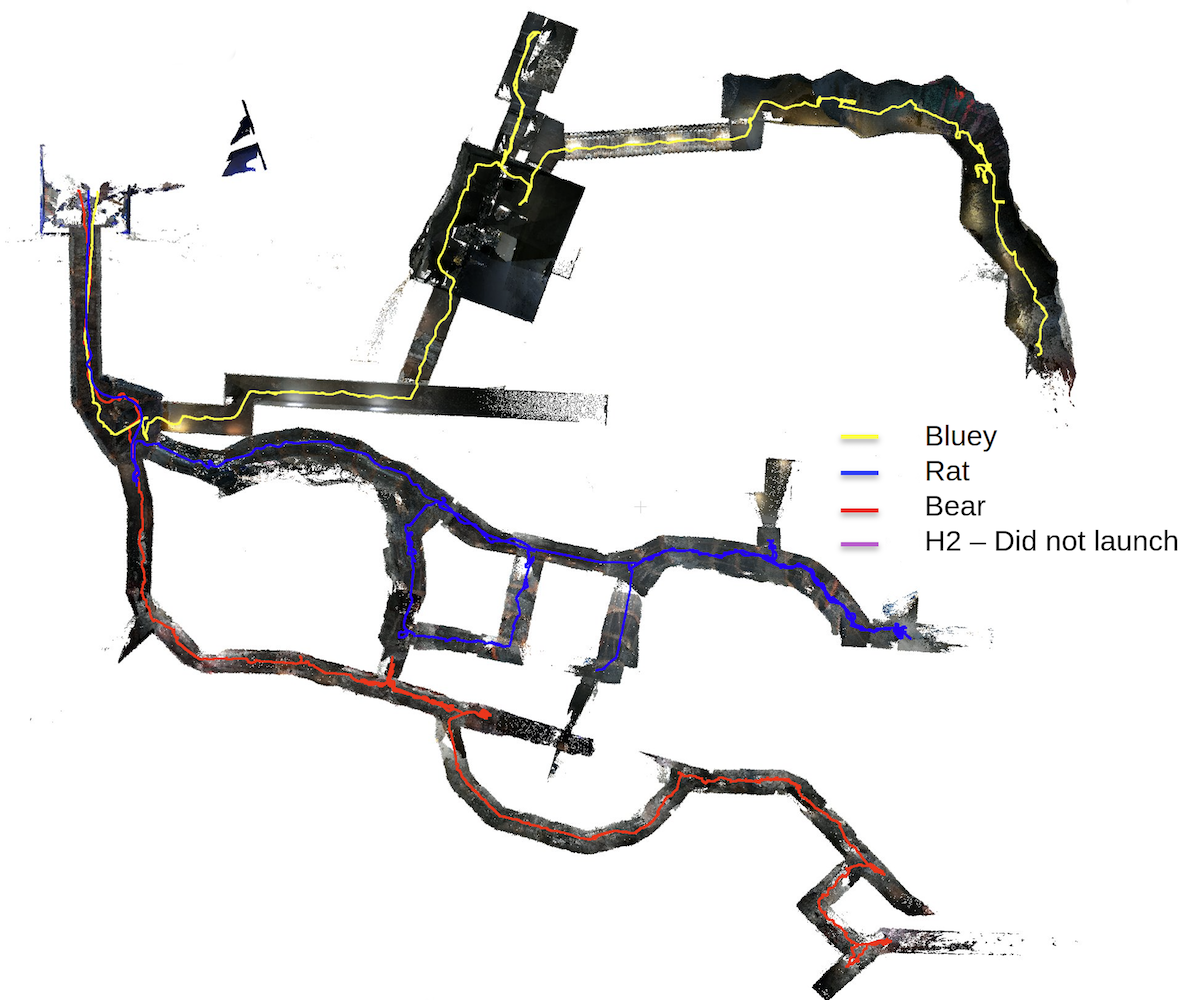}}
\subfloat[Object Reports]{\includegraphics[width=0.4\columnwidth]{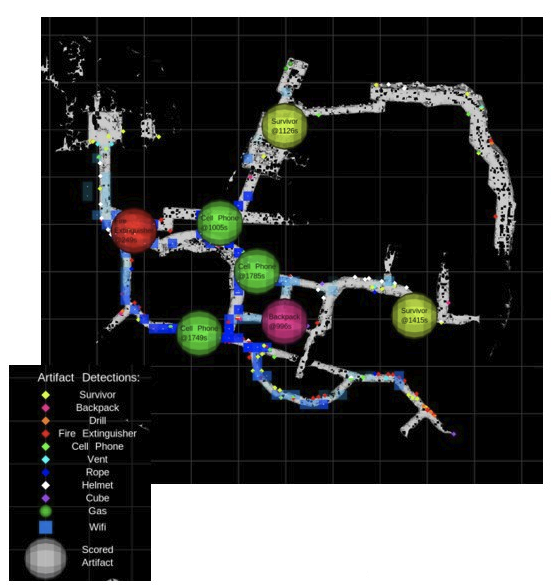}}
\caption{Course coverage and successful object reports for preliminary run 1. The starting area is located at the top left of each image. (a) shows robot paths by colour for Spot robot Bluey, ATRs Rat and Bear, and UAVs H2 (which was not launched). (b) shows the final map and object reports, based on information from the base station. The various artefact detections are shown as small dots in corresponding colours, while successfully scored artefacts are shown as large spheres, marked with the artefact time and scoring time (in seconds).
}
\label{fig:prelim1_results}
\end{figure}

The agents sent into the first preliminary run were Rat (ATR) with marsupial UAV platform H2, Bear (ATR), and Bluey (Spot). Bluey was the first robot sent into the course, and upon confirming three distinct environment types (urban, tunnel, and cave, see \figref{fig:prizerun_multiagent_offline}), the supervisor elected to send Bluey to explore the urban section. Rat was sent into the tunnel environment, and Bear into the cave section. \figref{fig:prelim1_results} shows the course traversal from each robot (\figref{fig:prelim1_results}a), and the object class and location of reports that were correct (\figref{fig:prelim1_results}b).

Communication nodes were placed at the entrance to the three sections, however, Bluey quickly lost communication with the ground station exploring the urban section. Connection was maintained between the ground station and the ATRs throughout the run. A connection was established between Rat and Bluey late in the run, enabling map information to be regained on the ground station. The recovered information revealed that Bluey had fallen traversing train tracks in the urban environment (within the first 10\,mins of the run). 

A total of seven objects were correctly detected and reported in the 30\,min run; five were detected visually and two were WiFi detections of cellphones. Another three were successfully detected, however, did not make their way back to the ground station, or were not efficiently displayed to the human supervisor before the end of the run. One artefact was detected but not reported as its probability was below the threshold for reporting. Post run analysis revealed a fault on the 2.4\,GHz channel of one of the communication nodes at the ground station, this was replaced before preliminary run 2.

The UGVs attempted to send a total of 188 visual artefact reports to the operator (6.26 reports per minute from all UGVs). Post analysis of the sent reports showed that 13 reports were true positives and the remaining 175 were false positives. Approximately one quarter of the false positives were due to artefacts being detected on the robot itself. This image mask configuration error was easily corrected in time for the second preliminary run. 

\figref{fig:prelim1_results} highlights that two or more robots explored the area near the start of the course as well as a passage which connects the cave and tunnel circuits. Analysis of the new artefact sharing capability of the artefact tracker showed that 13 artefacts (true and false positives) were detected by more than one robot. This feature reduced the number of artefact reports sent to the operator for inspection by 17 reports.

\subsection{Preliminary Run 2}
\label{ss:Prelim2}

The agents sent into the second preliminary run were Rat and Bear (ATRs) both with marsupial UAVs (H1 and H2), and Bingo (Spot). Bingo entered first, and was sent into the urban environment. The human supervisor was unable to prevent Bingo following the fate of Bluey from the first run. The robot quickly lost communications with the ground station, and slipped on the train tracks early in the run ($\sim$8\,min). Data from this robot was not recovered during the run. 

Rat was tasked to operate in the cave environment and a small cavern was identified for launching a UAV. The UAV (H2) was launched successfully and thoroughly explored the cavern. Rat become immobilised soon after launch with a suspected motor fault and remained in the cavern for the remainder of the run.

Bear was sent after Bingo into the urban section attempting to improve the communications backbone, however, a dynamic obstacle had activated as Bingo entered a warehouse area preventing Bear from pursuing Bingo directly. A communications node was dropped with Bear at a junction expected to help with communications to Bingo, but this was not successful. 

Bear explored several small office rooms before discovering an alternative entrance to the warehouse area. A second UAV (H1) was successfully launched in this area, however, the robot crashed spectacularly after ingesting debris soon after launch. Bear was the last active robot, and the final minutes were spent attempting to recover data from the lost Spot through the tunnel environment as in the previous run. Time ran out before this was achieved.

A total of eight artefacts were successfully reported during the 30\, min run. Seven of the artefact detections were from visual detections and the remaining artefact was detected using WiFi. Importantly, we detected every artefact we observed. The UGVs sent a total of 106 reports for the run (average of 3.53 reports per minute). Post analysis of the sent reports showed that 11 were true positives and the remaining 95 were false positives. Sharing artefact reports with all agents reduced the number of artefact reports sent to the operator for inspection by 5 reports.

\begin{figure}[ht]
\centering
\subfloat[Robot Paths]{\includegraphics[width=0.65\columnwidth]{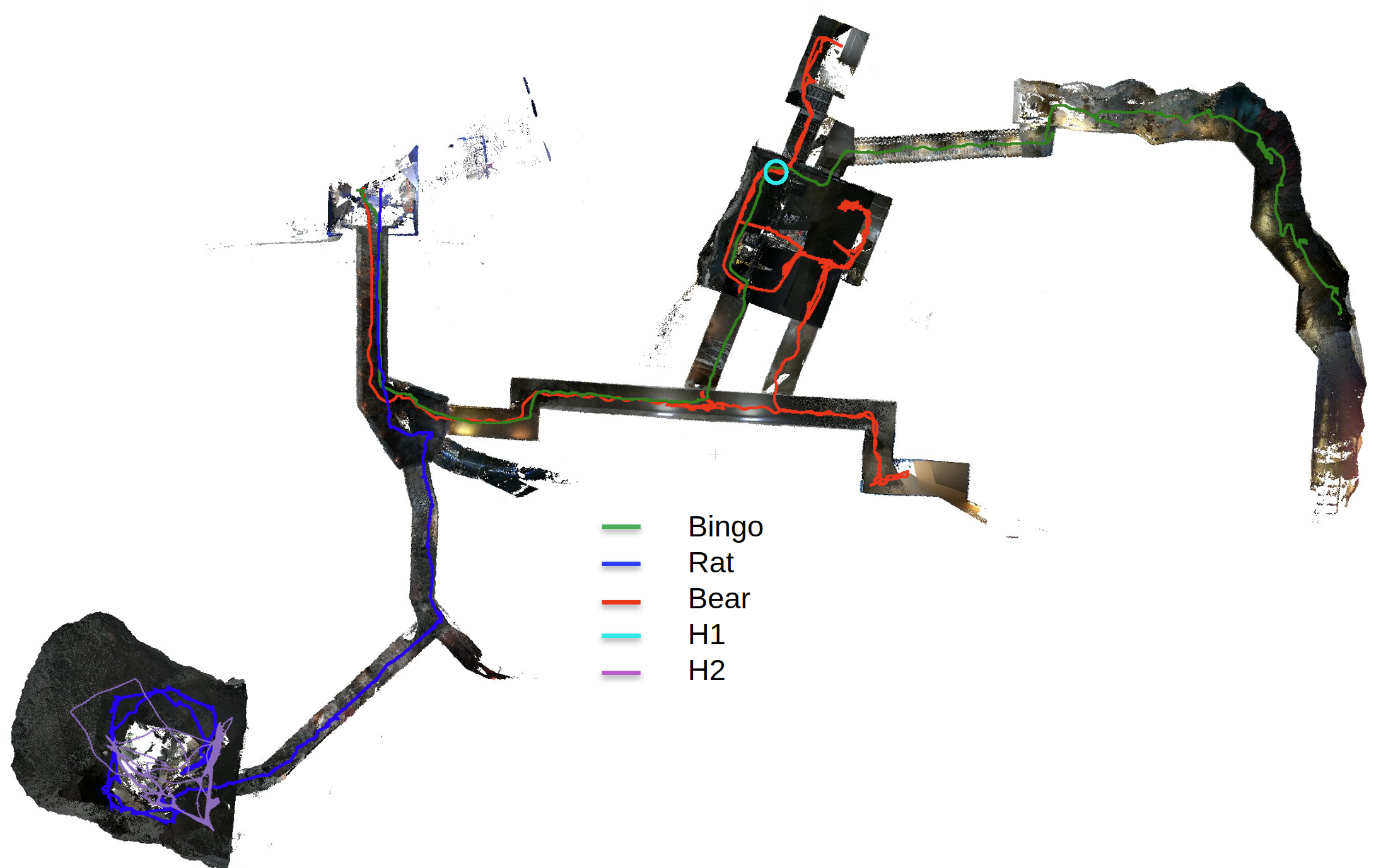}}
\subfloat[Object Reports]{\includegraphics[width=0.35\columnwidth]{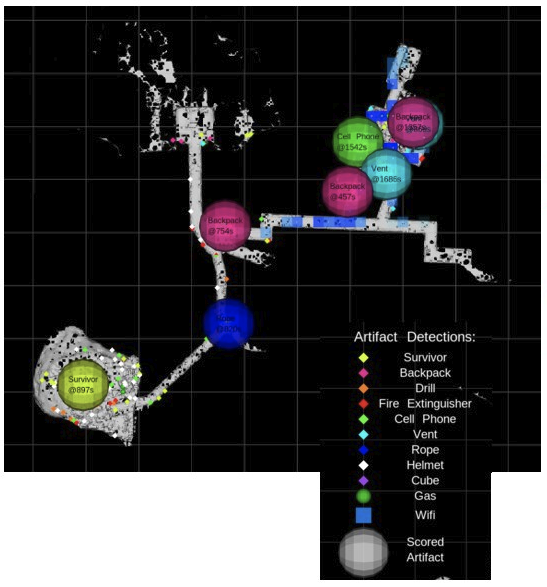}}
\caption{Course coverage and successful object reports for preliminary run 2. The starting area is located at the top left of each image. (a) shows robot paths by colour for Spot robot Bingo, ATRs Rat and Bear, and UAVs H1 and H2. (b) shows the final map and object reports, based on information from the base station. The various artefact detections are shown as small dots in corresponding colours, while successfully scored artefacts are shown as large spheres, marked with the artefact time and scoring time (in seconds).}
\label{fig:prelim2_results}
\end{figure}

\begin{figure}
\captionsetup[subfloat]{labelformat=empty}
\centering

\subfloat[]{\includegraphics[height=6cm]{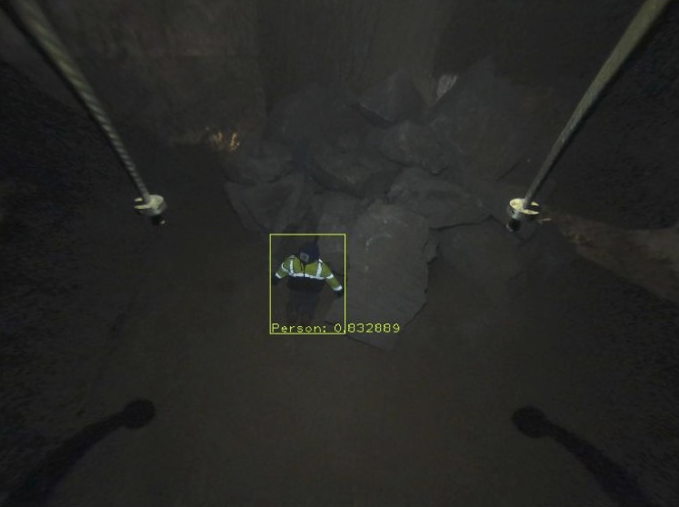}}
\hfill
\subfloat[]{\includegraphics[height=6cm]{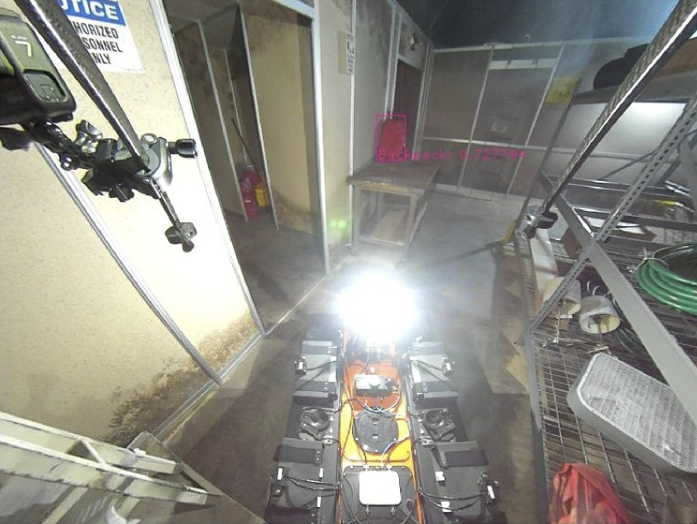}}
\caption{Example object detections from the UAVs.}
\label{fig:uav_detections}
\end{figure}

\figref{fig:prelim2_results} shows the course traversal and correct object reports from the second preliminary run. \figref{fig:uav_detections} shows examples of object detections from the UAVs. After the run several repairs were conducted including replacing an ATR motor and fixing antennas on damaged communication nodes. Repairs were performed with long distance support from Australia through ``Double 3'' telepresence robots from Double Robotics.

\begin{figure}
\captionsetup[subfloat]{labelformat=empty}
\centering
\subfloat[]{\includegraphics[trim=0cm 1.1cm 3cm 6cm, clip, height=2.5cm]{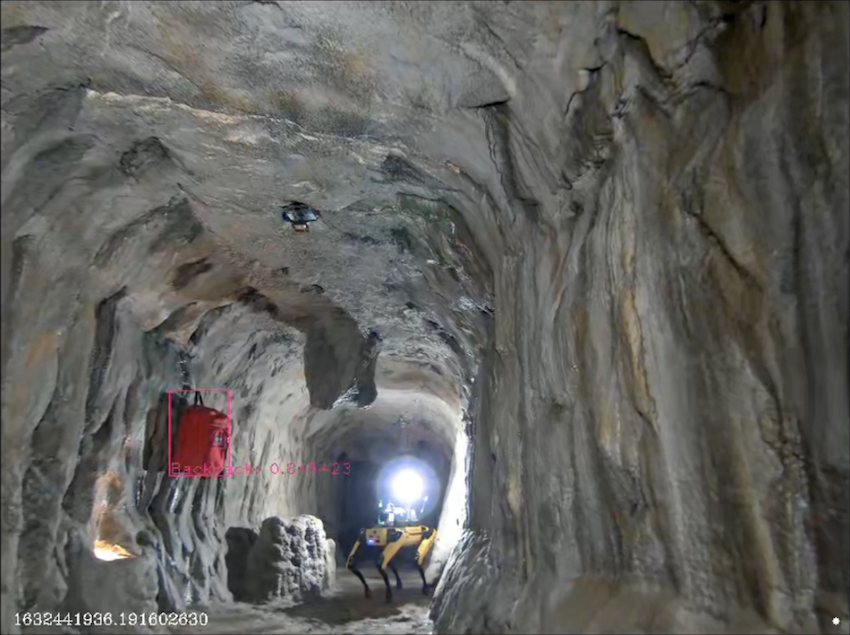}}
\subfloat[]{\includegraphics[trim=0cm 1.1cm 3cm 6cm, clip, height=2.5cm]{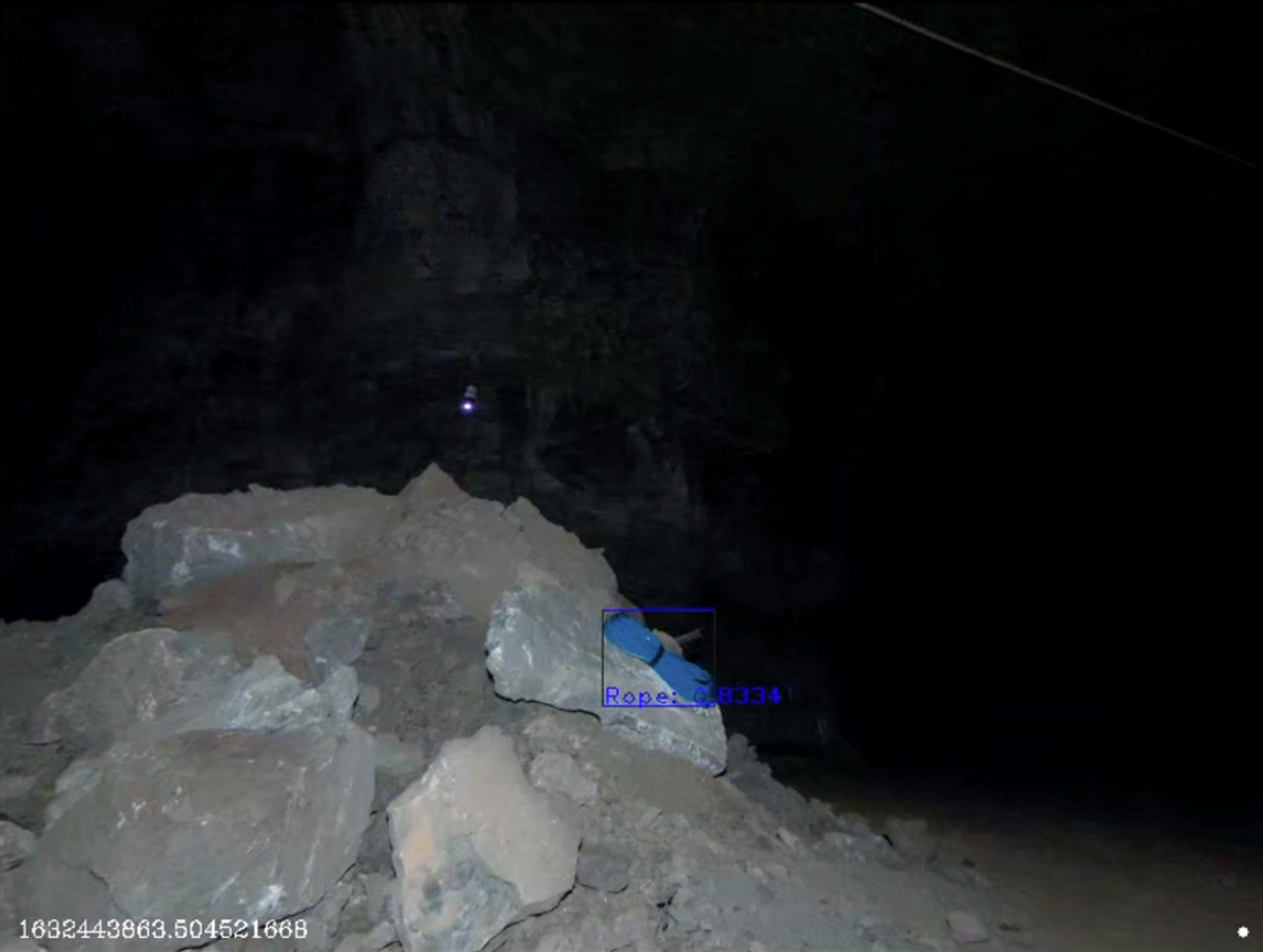}}
\subfloat[]{\includegraphics[trim=0cm 1.1cm 3cm 6cm, clip, height=2.5cm]{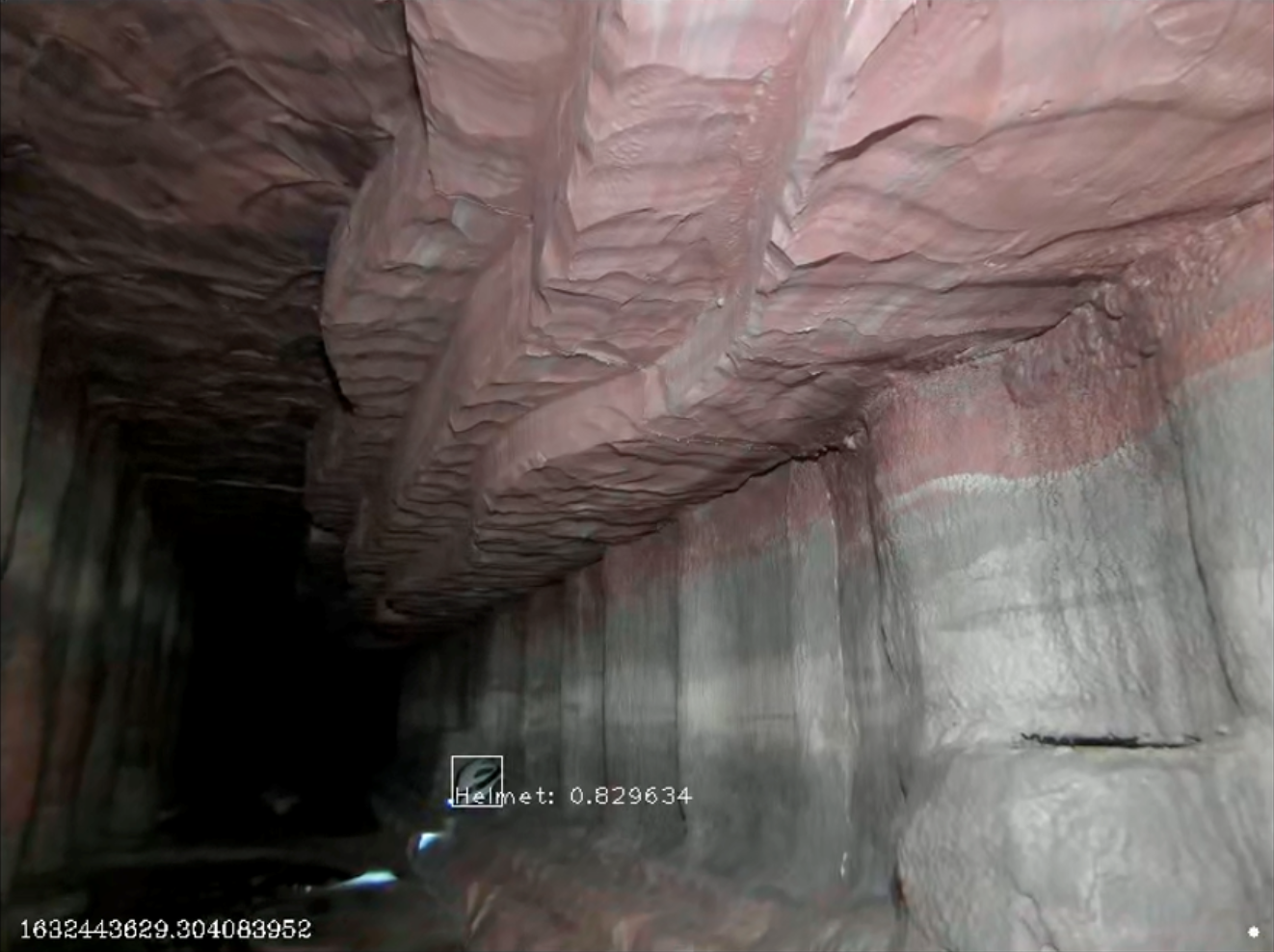}}
\hfill
\subfloat[]{\includegraphics[trim=0cm 1.1cm 3cm 6cm, clip, height=2.5cm]{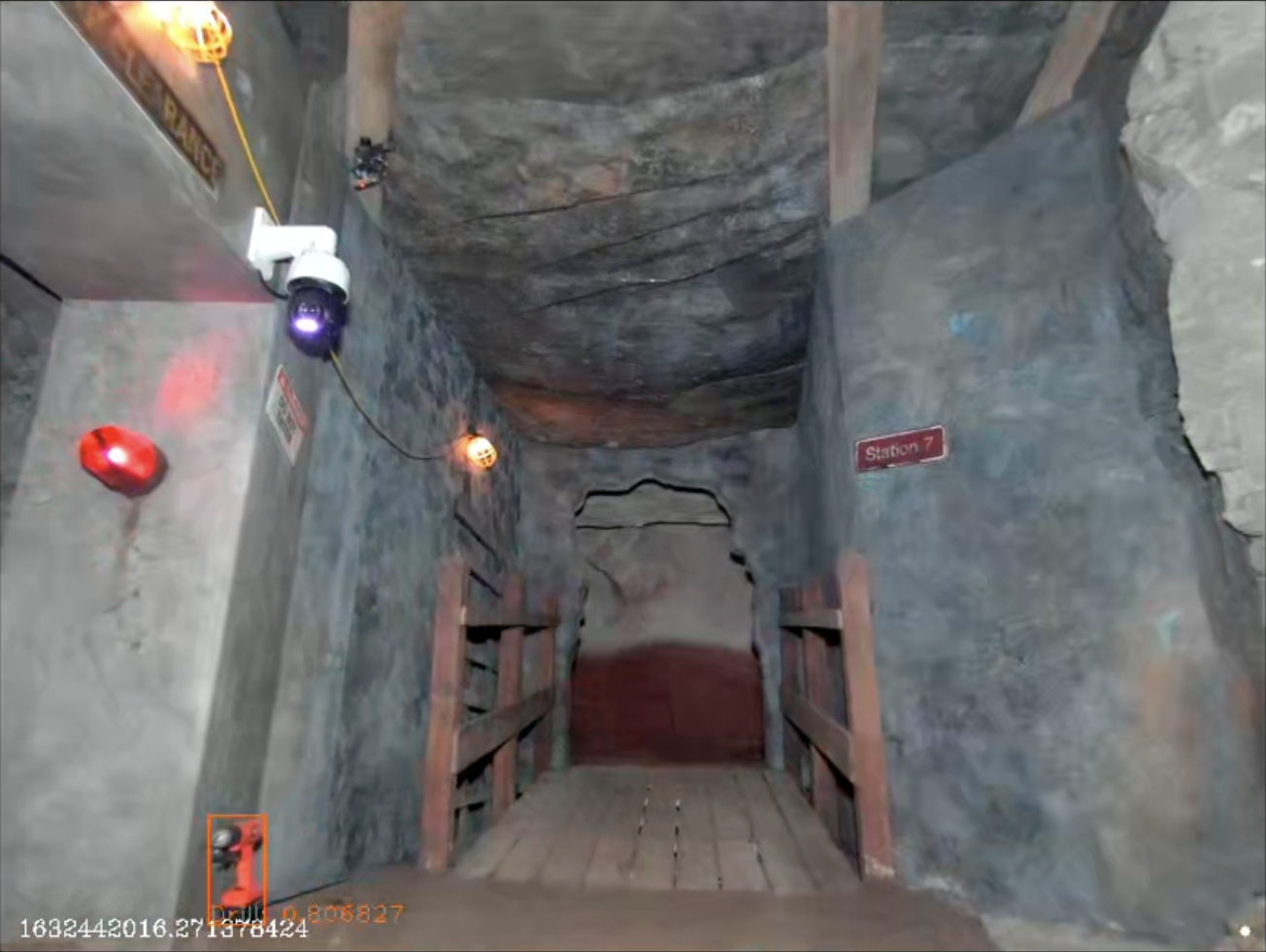}}
\subfloat[]{\includegraphics[trim=0cm 1.1cm 3cm 6cm, clip, height=2.5cm]{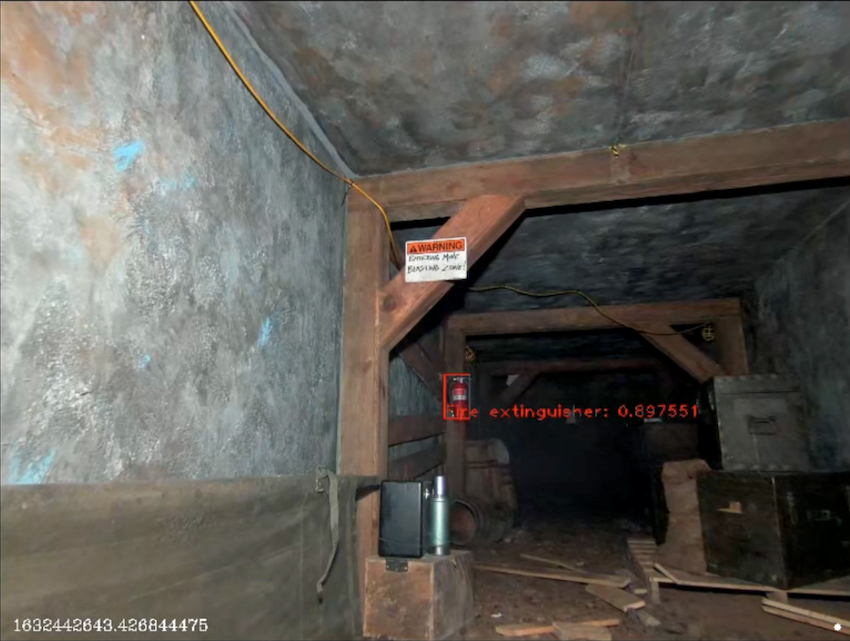}}
\subfloat[]{\includegraphics[trim=0cm 1.1cm 3cm 6cm, clip, height=2.5cm]{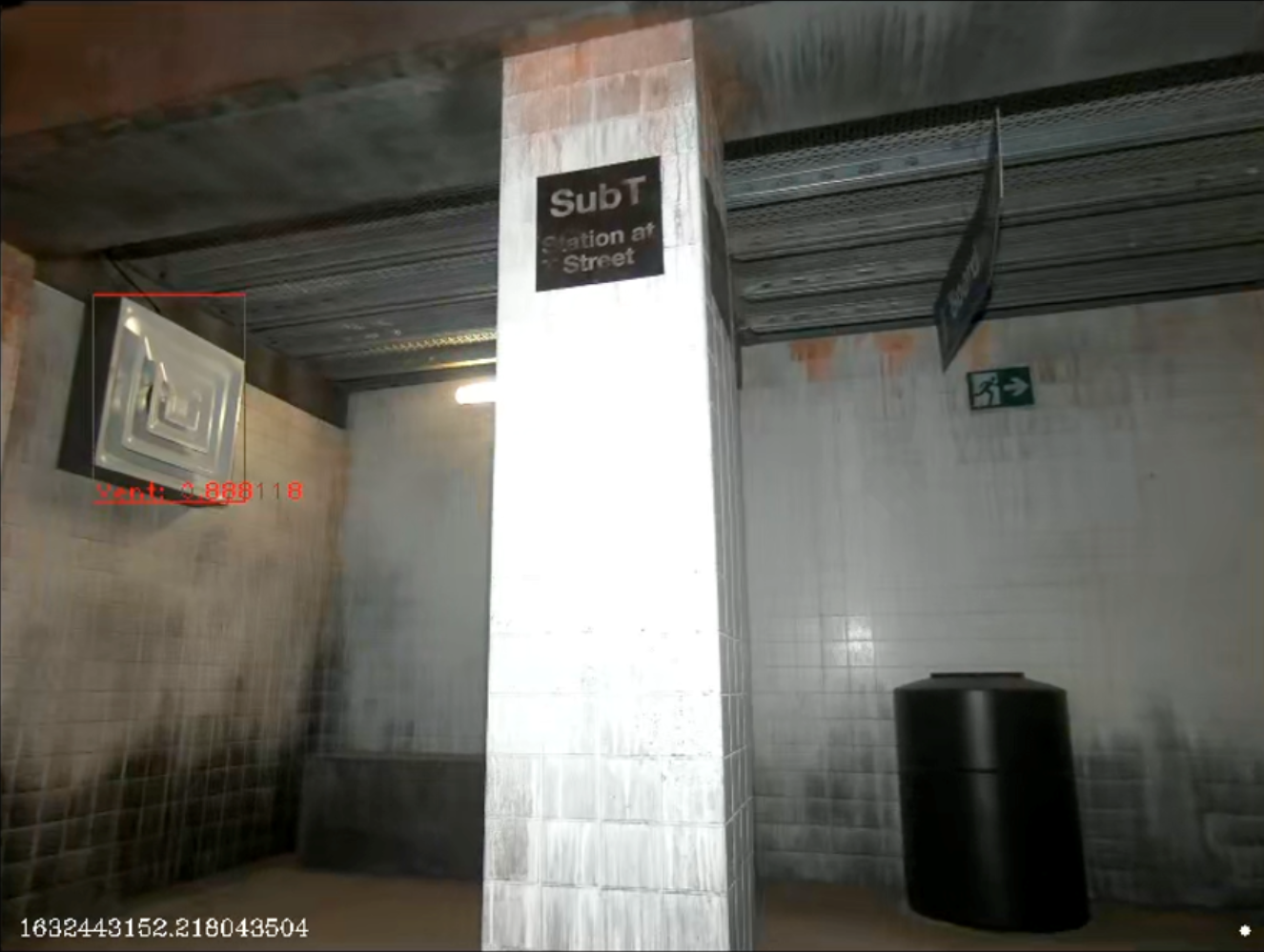}}
\caption{Example object detections from the UGVs.}
\label{fig:ugv_detections}
\end{figure}

\subsection{Final Prize Run}
\label{ss:FinalPrizeRun}
All available robots were taken into the final prize run, including two ATRs (Rat and Bear), two Spot robots (Bluey and Bingo), and two UAVs (H1 and H2). Both Spot robots were sent in first, one into the urban environment (Bluey), and one into the cave environment (Bingo). Both ATRs placed communication nodes at the intersection of the three environments, then Rat was sent into the tunnel section, and Bear into the cave.

Bingo rapidly progressed through the cave environment, and quickly disappeared from communications. The human supervisor carefully de-prioritised the region leading to the train tracks in the urban section, preventing Bluey from repeating the mistake of the two preliminary runs. Bluey lost communications as it headed towards the train platform at the end of the urban section.

Rat explored the tunnel section, making slow progress over tangled fire hose. The robot dropped a communications node at a junction, and made an unsafe turn over the rail of a mine track resulting in the de-tracking and immobilisation of the robot ($\sim$28\,min into the run). The tunnel was narrow and the ceiling height too low to launch the UAV without risking catastrophic failure of the map. Bear made slow progress through the cave environment, successfully detecting an object in the small cavern (previously explored in  preliminary run 2), then returning to the main channel of the cave to follow Bingo's path. 

Approximately 30\,min into the run, communications with Bluey were re-established. While out of communications, Bluey had climbed stairs to reach the subway platform, explored the top of the platform, descended stairs to the subway tunnel and exited towards the tunnel course. In the process, Bluey briefly reconnected with Bingo, and was subsequently able to mule part of its data back to the ground station. Bingo had located a large cavern at the end of the cave section, and had appeared to have fallen (post-run analysis confirmed Bingo fell at 22\,min). The human supervisor directed Bluey towards the large cavern, and prioritised getting Bear to this location with a UAV.

Bear was unable to make progress through the cave environment with the UAV mounted on the back due to the low clearance of the tunnel. There appeared to be a path leading to the large cavern through to the tunnel section, and the human supervisor directed Bear to backtrack to this location. However, a dynamic obstacle had closed the passage behind Rat, and the only alternative route was blocked by Rat, which was immobile. The human supervisor had to remove the UAV to enable the traversal of the narrow cave to recover data from the Spots. With time running out, the human supervisor relied heavily on teleoperation for faster traversal, ensuring communications were maintained (by dropping communication nodes along the way). Bear was controlled to the smaller cavern to launch the UAV, then made quick progress through the cave section towards the large cavern. In the final minutes, Bear established a connection with Bluey (fallen at $\sim$40\,min), and the final scored object detection came through to the ground station (reported at 25\,s remaining).

\begin{figure}[b!]
\centering
\subfloat[Robot Paths]{\includegraphics[width=0.85\columnwidth]{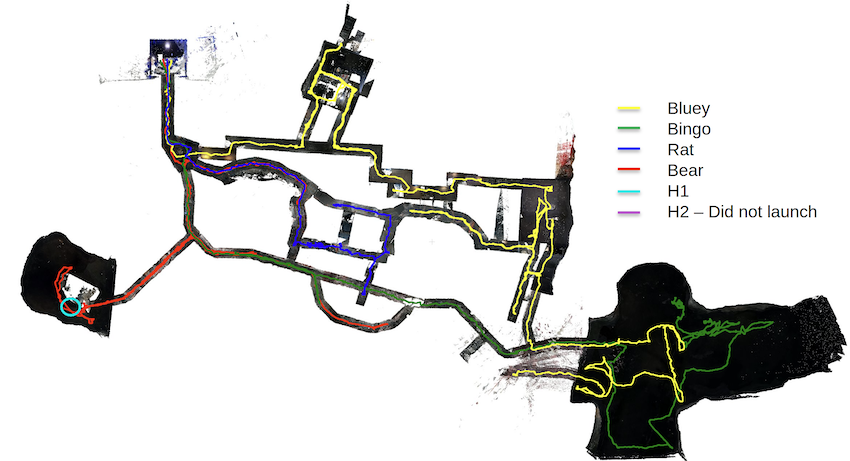}}
\hfill
\subfloat[Object Reports]{\includegraphics[width=0.85\columnwidth]{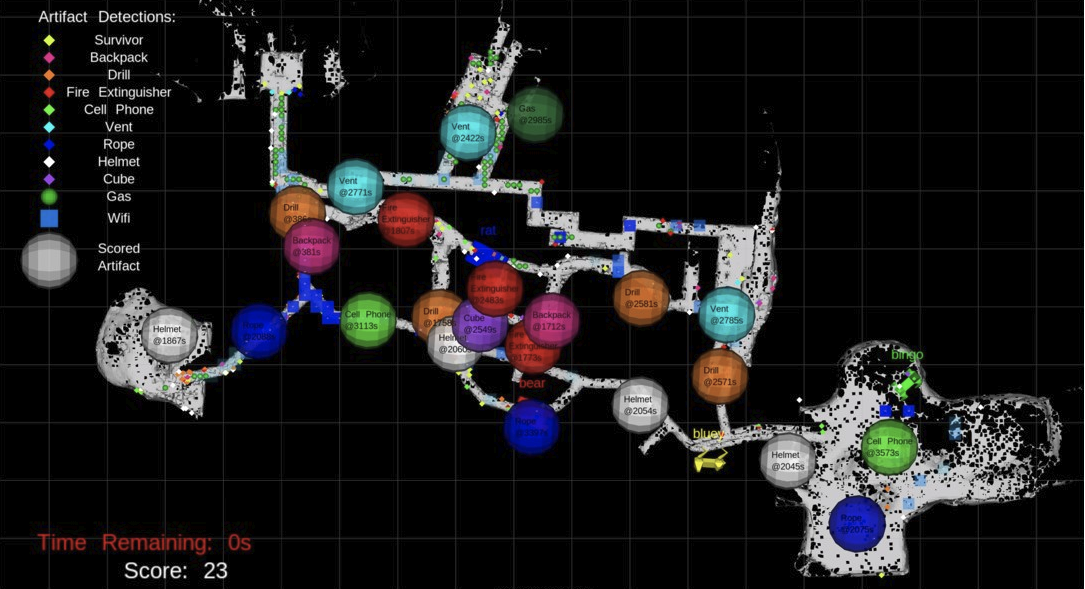}}
\caption{Course coverage and successful object reports for the final prize run. The starting area is located at the top left of each image. (a) shows robot paths by colour for Spot robots Bluey and Bingo, ATRs Rat and Bear, and UAVs H1 and H2 (the latter of which was not launched). (b) shows the final map and object reports, based on information from the base station. The various artefact detections are shown as small dots in corresponding colours, while successfully scored artefacts are shown as large spheres, marked with the artefact time and scoring time (in seconds).}
\label{fig:prize_run_results}
\end{figure}

A total of 23 objects were successfully detected. Another four objects were detected but not reported. Of these, one detection did not make it back to the ground station in time, and three non-visual detections (gas and two cellphones) were difficult to localise given the high load of the human supervisor at the end of the run (Table~\ref{tab:PrizeRunDetection}). \figref{fig:prize_run_results} shows the course traversal for each ground platform and the scored object reports. Table~\ref{tab:PrizeRunDetectionByPlatform} shows the percentage of object reports by platform type (ATR, Spot, UAV). These results indicate the Spot robots were slightly better at detecting objects than the ATR's (more detections for each meter of terrain covered). This can be explained by the increased field of view from an elevated perception pack and reduced occlusion compared with the UAV-carrying ATRs. The UAVs successfully detected objects in the preliminary runs but were not utilised for object detections in the final prize run, as the ideal launch locations were not able to be reached by the UAV-carrying ATRs. The UAV that was launched in the prize run did not move from its hover position and did not report any detections; as previously described, it was launched in an already-explored cavern for the purpose of providing additional clearance for the ATR.

\begin{figure}[tb]
\centering
\includegraphics[width=0.75\columnwidth]{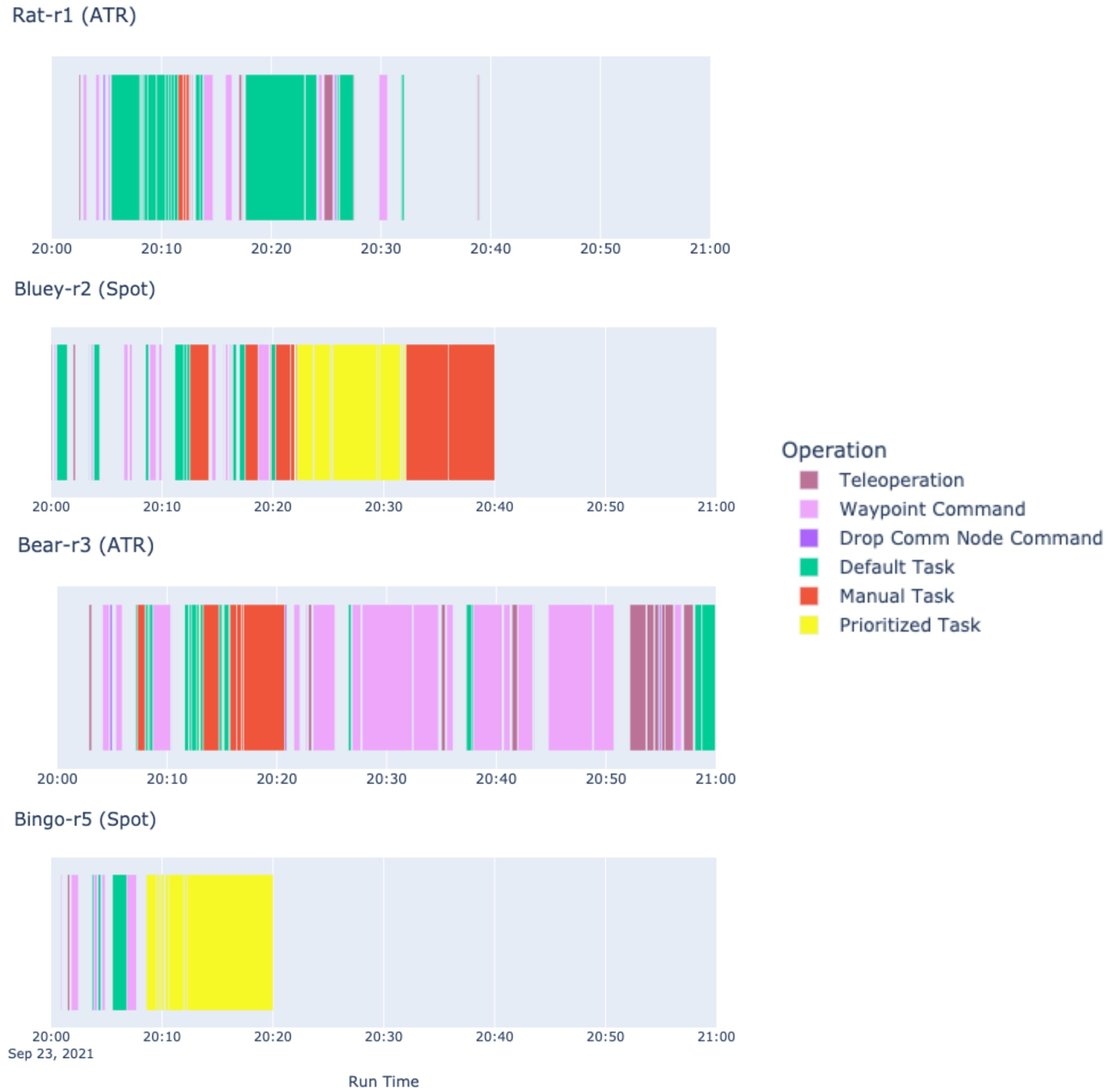}
\caption{Mode of operation of the autonomy system by time for each UGV during the prize run. Teleoperation denotes direct joystick control, whereas Waypoint Command denotes navigation to a specified waypoint. Drop Comm Node Command denotes an operator command to navigate to a specified location and drop a communications node. Default Task denotes the regular autonomous mode of task allocation (without priorization), whereas Manual Task denotes an operator override to execute a specific task, and Prioritized task denotes task allocation where the selected task was in a region that had been prioritized by the operator.}
\label{fig:autonomy_mode_prize_run}
\end{figure}

The mode of operation of each robot as a function of time is shown in \figref{fig:autonomy_mode_prize_run}, illustrating the use of the directed autonomy functionality to achieve the results previously described. A common pattern is the use of waypoints to position a robot in the desired area, followed by autonomous operation, either in the default mode (without prioritisation), or using prioritisation to ensure continued progress in the desired direction. For example, Bingo (r5) shows extensive use of prioritisation regions to achieve the desired result, while Bluey (r2) shows both task prioritisation regions and manual task assignment to direct it to the region of interest. Once Bear (r3) was the only functional agent within communications range, it was controlled using waypoints and later, in the time critical period, using teleoperation. The percentage of a robots run under each mode of operation is provided in Table \ref{table:final_run_modes}, showing the degree of operator intervention for each agent as a percentage of the robots total operational time.

\begin{table}[!ht]
\centering
\caption{Percentage of the prize run in each mode of operation for each platform, summarising data in \figref{fig:autonomy_mode_prize_run}. See \figref{fig:autonomy_mode_prize_run} for description of robot modes.}
\label{table:final_run_modes}

\begin{tabular}{l c c c c} 
 \toprule
  & Rat-r1 & Bear-r3 & Bluey-r2 & Bingo-r5 \\
    & (ATR) & (ATR) & (Spot) & (Spot) \\ 

 \midrule
 Waypoint Command       & 15.8\% &  54.1\%  & 11.5\% & 15.3\%\\
 Drop Comm Node Command &  1.7\% &   1.3\%  &  0.0\% &  0.0\%\\
 Teleoperation          & 12.5\% &  16.7\%  &  1.6\% &  1.3\%\\
 Prioritized Task       &  0.0\% &   0.2\%  & 31.3\% & 72.9\%\\
 Manual Task            &  4.1\% &  15.2\%  & 41.5\% &  0.0\%\\
 Default Task           & 65.9\% &  12.5\%  & 14.1\% & 10.5\%\\ 
 \bottomrule
\end{tabular}

\end{table}

The distribution of time that the ATRs spent using each planner, i.e., default hybrid A*, or the gaps planner of Section \ref{sss:gaps_planner}, is shown in Table \ref{table:final_run_planner}. The Spot robots are not shown as they only use hybrid A* (with the smaller agent dimension, the gaps planner was not found to be necessary). The table shows that the ATRs utilised the gaps planner for 21.3\% of the time, or 22.9\% of time where a plan was active. This was much larger than expected, and is due to the extensive narrow tunnels in the course. In most test environments, the gaps planner only activated in order to pass through narrow doorways; in this course, it enabled (slow) progress across parts of the course that would have been otherwise impassible.

\begin{table}[!ht]
\centering
\caption{Percentage of the prize run where ATR robots utilized each planner (out of time when path follow behaviour was active). ``None'' indicates that no planner was active, i.e., either the planner(s) are still generating a plan, or that no feasible plan was found.}
\label{table:final_run_planner}
\begin{tabular}{l c c c} 
 \toprule
  & Rat-r1 (ATR) & Bear-r3 (ATR) & Overall ATR\\ [0.5ex] 
 \midrule
 Hybrid A* & 78.0\% &  68.2\% &  71.6\% \\
 Gaps      & 15.1\% &  24.6\% &  21.3\% \\
 None      &  6.9\% &   7.2\% &   7.1\% \\
 \bottomrule
\end{tabular}

\end{table}

The distribution of time that the robots spent using each behaviour is shown in Table \ref{table:final_run_behaviour}, focusing on autonomous motion-based behaviours, i.e., excluding teleoperation and stopped behaviours. As expected, the decollide behaviour is utilized for a small proportion of the time, though its use reenables the path planning behaviour. Orientation correction exists to prevent robot tipping, and saw a single activation for a fraction of a second.

\begin{table}[!ht]
\centering
\caption{Percentage of the prize run where ATR robots utilized each behaviour (focusing on autonomous motion behaviours, excluding stopped and teleoperation-related behaviours). *Path following is achieved through different behaviours on ATR and Spot, utilizing the method of Section \ref{sss:path_follow} for ATR, Section \ref{sss:spot_local_nav} for Spot.}
\label{table:final_run_behaviour}
\begin{tabular}{lccccc} 
 \toprule
  & Rat-r1 & Bear-r3 & Bluey-r2 & Bingo-r5 & Overall \\ 
    & (ATR) & (ATR) & (Spot) & (Spot) &  \\ 

 \midrule
 Orient correction &    0\% & 0.002\% &    0\% &    0\% & 0.0008\% \\
 Path follow*      & 96.0\% &  93.3\% & 98.2\% & 97.4\% &   95.7\% \\
 Decollide         &  4.0\% &   6.7\% &  1.8\% &  2.6\% &    4.3\% \\
 \bottomrule
\end{tabular}

\end{table}

\begin{figure}[tb]
\centering
\includegraphics[width=0.75\columnwidth]{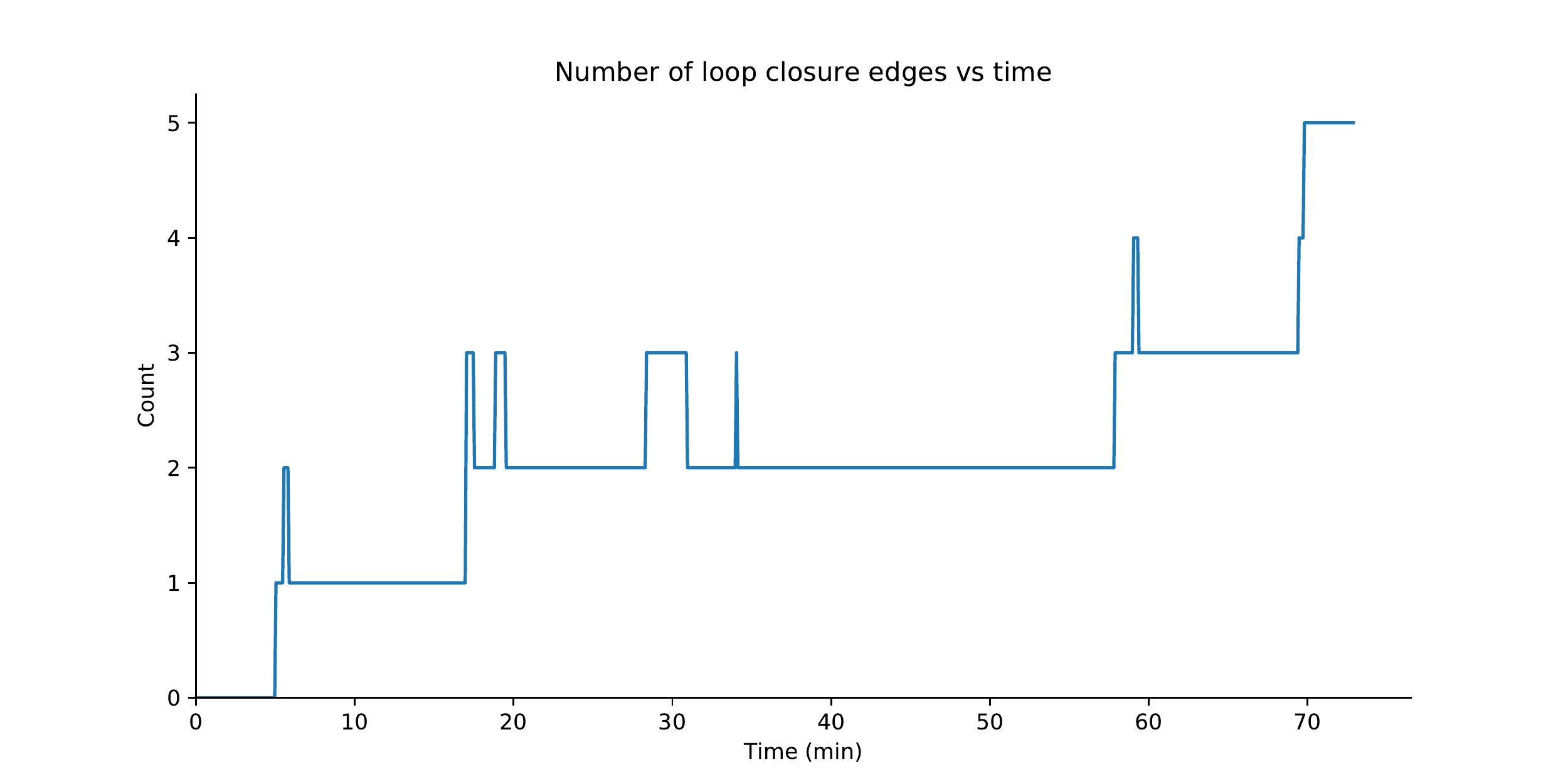}
\caption{Number of loop closure edges (i.e., edges not implied by odometry) as a function of time during the final prize run. Note that the run ends at 60\,min; loop closures after that point are due to the additional data subsequently relayed.}
\label{fig:bluey_loop_closure_time}
\end{figure}

The final pose graph for one agent (Bluey-r2) is shown in \figref{fig:ma_pose_graph}. The total number of loop closure edges in this graph is three. \figref{fig:bluey_loop_closure_time} shows the number of loop closure edges in the pose graph as a function of time (i.e., the number of edges that are not implied by odometry). When a loop closure occurs, there are often additional transient edges, which subsequently disappear as the graph is simplified, demoting some of the root nodes to become child nodes. The figure also shows that additional loop closures occurred after the end of the run (i.e., after the 60\,min mark); these occurred as data from Bear (r3) continued to be relayed through a slow (due to low SNR) communications link.

The agents sent a total of 203 artefact reports to the operator during the one hour mission (an average of 3.38 reports per minute). Of the 203 reports, 29 reports were of true positives and the remaining 174 were false positives. An analysis of the new multi-agent artefact tracking system showed that multiple agents had detected the same artefact (true positive or false positive) at 16 different locationsand had reduced the number of duplicate artefact observations seen by the operator by 27 reports. \figref{fig:ugv_detections} shows examples of object detections from the UGVs.

\begin{table}[th]
\centering
\caption{Summary of detection results in final prize run. *Cube counted as visual, cell phone as non-visual. $\dagger$Data not communicated to base in time. $\ddagger$Non-visual detection but too sparse/non-specific.}
\begin{tabular}{lcc}
\toprule
& Visual & Non-visual* \\
\midrule
Detected and scored & 20 & 3 \\
Detected but not reported/scored & 1$\dagger$ & 4$\ddagger$ \\
Passed but not detected & 3 & 0 \\
Did not pass & 9 & 0\\
\bottomrule
\end{tabular}
\label{tab:PrizeRunDetection}
\end{table}

\begin{table}[th]
\centering
\caption{Percentage of object detections and terrain coverage by robot type in the final prize run.}
\begin{tabular}{lccc}
\toprule
& Total Reports & True Positive & Terrain Coverage\\
\midrule
ATR & 43\% & 44\% & 49\% \\
Spot & 57\% & 55\% & 51\%\\
UAV & 0\% & 0\% & 0\%\\
\bottomrule
\end{tabular}
\label{tab:PrizeRunDetectionByPlatform}
\end{table}

\begin{figure}[tb]
\centering
\includegraphics[width=0.95\columnwidth]{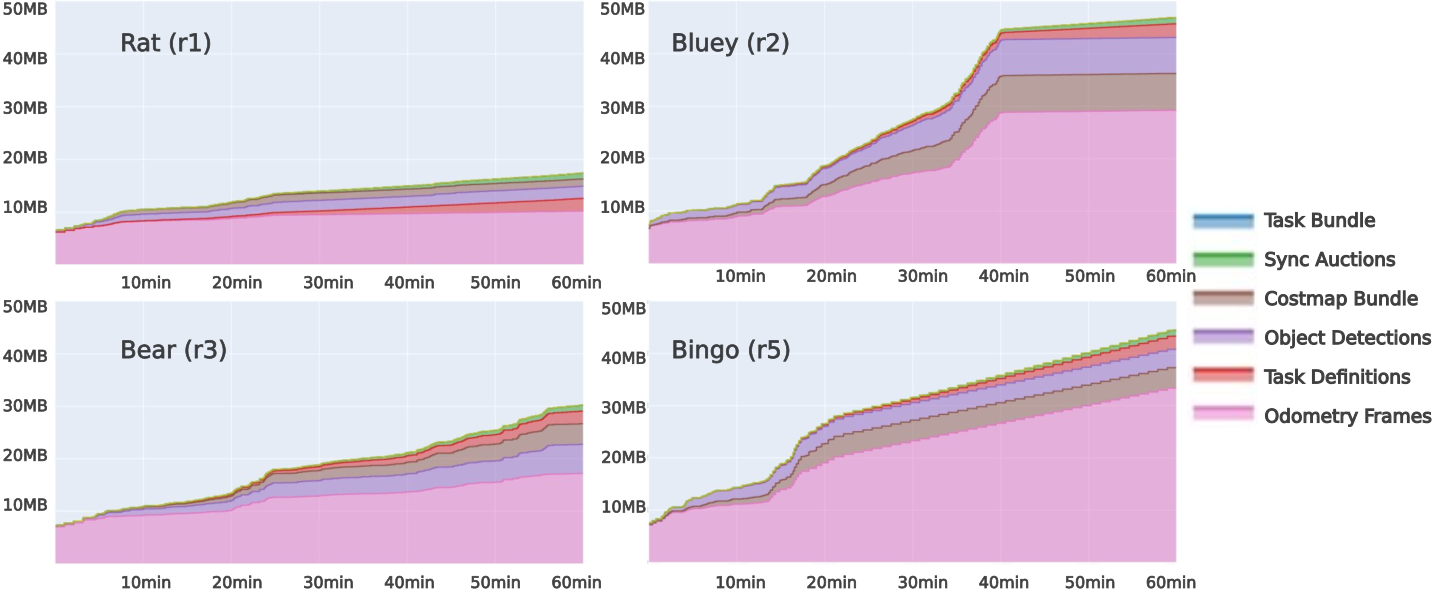}
\caption{Cumulative data load (megabytes) for each of the ground robots during the 60 minute final prize run.}
\label{fig:mule_load}
\end{figure}

\figref{fig:mule_load} shows the cumulative data generated by each of the autonomy processes of each ground agent during the final prize run. The figure shows that by far the largest contributor is the SLAM odometry frames required to permit each agent to build a unified map. Cost map bundles are the second contributor, which allow building of the unified traversability map on top of the SLAM solution. Object detections utilise a similar amount data, enabled by the tracking methods described in Section \ref{sec:perception}. Task definitions and bundles and sync auctions represent data used by the task allocator to achieve a common understanding of the task set and robot-task assignments. Note that Bingo (r5) fell at around 20\,min, but due to its resting position, odometry frames did not appear sufficiently similar to be suppressed, and thus the agent continued to generate significant odometry data. Conversely, Bluey (r5) fell at around 40\,min, and generated little odometry data thereafter.

Table~\ref{tab:PrizeRunAvgCPUPack} shows the average CPU usage for each of the processes on the perception pack, while Table~\ref{tab:PrizeRunAvgCPUAutonomy} shows the average CPU usage for processes on the autonomy computer. SLAM, image processing and object detection use similar CPU resources, although the latter two also use GPU resources. The heaviest CPU usage on the autonomy computer is OHM, which utilises the GPU for ray tracing, and the CPU for data pre-processing and height map generation. Differences between ATRs and Spots are caused by the Spot's different set of behaviours (e.g., passing a trajectory for spot to follow rather than using path follow), higher speed, greater use of autonomous exploration (and hence task allocation), and processor differences.

As reported by~\cite{chung_into_2023}, Team CSIRO Data61 excelled in the final prize run using alternate relevant evaluation metrics, achieving the lowest map deviation, greatest map coverage, highest report success rate (scored/submitted), and most accurate report with the smallest detection error (meters from ground truth). We were also the fastest team to enter the course with a robot (seconds from run start).

\begin{table}[th]
\centering
\caption{Average percentage of CPU usage on the perception pack in the final prize run (where 100\% denotes utilization of a full virtual core). Image processing encompasses image acquisition, rectification, and recording, and likewise lidar/IMU encompasses the respective signal acquisition and recording.}
\begin{tabular}{lcc}
\toprule
& CPU usage (ATR) & CPU usage (Spot) \\
\midrule
SLAM & 136.5\% & 158.5\% \\
Image processing & 135.5\% & 140.5\% \\
Object detection & 127\% & 133.5\% \\
Lidar/IMU & 73\% & 72.5\% \\
WiFi detection & 13\% & 14\%\\
\bottomrule
\end{tabular}
\label{tab:PrizeRunAvgCPUPack}
\end{table}


\begin{table}[th]
\centering
\caption{Average percentage of CPU usage on the autonomy computer in the final prize run (where 100\% denotes utilization of a full virtual core).}
\begin{tabular}{lcc}
\toprule
& CPU usage (ATR) & CPU usage (Spot) \\
\midrule
OHM (occupancy mapping) & 117.0\% & 112.4\% \\
Cost map generation & 24.3\% & 27.2\% \\
Behaviours & 14.4\% & 7.4\% \\
Hybrid A* planner & 6.2\% & 12.6\% \\
Gaps planner & 22.0\% & NA \\
Global mapping and planning & 26.3\% & 29.0\% \\
Exploration & 6.0\% & 14.9\% \\
Task allocation & 5.1\% & 10.8\% \\
Communications & 2.0\% & 1.8\% \\
External interfaces & 40.5\% & 64.0\% \\
Recording & 47.4\% & 20.2\% \\
\bottomrule
\end{tabular}
\label{tab:PrizeRunAvgCPUAutonomy}
\end{table}

\section{Lessons Learned}
\label{sec:Lessons}
The intensive development over the SubT program has provided us with a number of useful insights about how to conduct an activity of such scale. Some of these reflect things that our team did well, while others are hard-learned lessons where in hindsight we see the need to do things differently. In this section, we aim to chronicle some of these.

\subsection{Development and Testing}
Prior to the tunnel circuit, a synthetic tunnel environment was constructed on site, as illustrated in \figref{fig:QCAT_tunnel}. Over time, this was expanded to include stairs, mezzanines, and a terrain park. Testing initially focused on this environment, but grew to incorporate as many elements as we could recreate on site, for example, incorporating a long traverse from the tunnel to industrial warehouse regions. In the end, the team's strengths and weaknesses reflect the environments to which we had regular access. We did not have regular access to a representative underground communication environment, which made development of features such as autonomous communications node dropping difficult to validate (thus this task remained manual).

Throughout the program, we maintained a regular cadence with weekly integration testing. These test activities served several functions; providing the team with a holistic view of where each person's work fits into the capability as a whole, enabling clear prioritisation of work by regularly demonstrating the significance of different issues, and quickly revealing problems that arise when integrating work from multiple developers. While this process was invaluable in the lead up to challenge events, during other development periods, some team members found it limiting due to the time occupied by the test itself, as well as post-test analysis of results. Subsequently, in follow-on work, we have dropped back to fortnightly tests.

The weekly test regime greatly clarified the robustness requirements for agents. Platforms were run for many hundreds of hours, and the need to address issues arising from intermittent failures was highlighted by the impact they had on the overall test conduct and consequent team efficiency.

The team benefited from high standards in software development, including use of continuous integration servers, and peer review through enforced pull requests. The high quality Gazebo-based simulation environment was critical to development, and productivity was noticeably slower on features that were not adequately modelled in simulation but rather required extensive on-robot testing.

Due to the aggressive development schedule, it was regularly the case that hardware was not complete on the full robot fleet until shortly before each event. This last-minute scale up of the robot fleet led to a range of issues. Again, this can be viewed as being related to limitations in simulation. For example, communications were not well-modelled in simulation, so extensive difficulties were experienced when the fleet was scaled up towards the end of the campaign, providing higher traffic and more complications unique to each platform class. Similarly, computational limitations of the simulation environment in general did not support testing of the full fleet size on available hardware; work has since been conducted to enable the use of parallel computing environments in simulation, permitting greater scaling.

\subsection{Platforms}
Reflecting on the progression of our team's platforms, a key strength of our approach was a willingness to pivot rapidly, embracing opportunities to leverage developments in commercial offerings. Our original concept of operations centred around a bespoke hexapod design \citep{steindl_2020}, with a goal of providing the ability to navigate extreme terrain. This concept adapted based on two major learnings. Firstly, the surprising capability of the tracked BIA5 OzBot ATR platform on rough terrain (e.g., slopes up to 60\degree) significantly changed our view of the trade-off between platform types. Secondly, the engineering effort involved in developing a platform to the point where it has sufficient robustness to be a viable candidate in the challenge context was difficult to sustain under the resource constraints. As new, commercially available platforms emerged with the benefit of far greater engineering investments, the cost/benefit of bespoke development became less compelling.

Another aspect of our team's experience with platforms was the significant engineering effort required to adapt commercial platforms intended for teleoperation to robust autonomous operation. As described in \secref{sss:atr}, autonomous systems sometimes exerted control outside the designers' expectations, which led to outcomes such as motor burnout, and additionally, stock control systems made precise motion difficult. This was experienced with a range of wheeled and tracked platforms.

Through the duration of the program, the capability of commercial quadruped platforms has also increased greatly. As discussed above, our initial concept was on hexapod platforms based on the intuition that the additional legs would provide valuable improvements to stability on rough terrain. Again, the tremendous commercial investment in platforms such as Boston Dynamics Spot and ANYbotics ANYmal shows that the additional maturity of these quadruped platforms overcomes any advantage that an early prototype hexapod may hold. While these have come a long way, our own experience with Spot shows that falls on rough terrain are still an issue, and our original hypothesis regarding hexapod platforms may still stand, though the engineering investment necessary to test it would be large.

Finally, our approach of a common sensing pack and navigation stack paid large dividends throughout the program. All ground platforms utilised the same sensing solution and autonomy varied only through minor configuration parameter changes, and utilising outputs at different levels (e.g., sending trajectories to Spot vs low level control of tracked platforms).

\subsection{SLAM}
Our SLAM solution evolved significantly during the course of the program, especially in aspects relating to multi-agent systems. Our solution was quite sufficient in the environments tested, and rarely presented a limitation to overall performance. However, there are a variety of qualifications on that outcome:
\begin{itemize}
\item The sensor payload with the spinning lidar is both expensive and heavy. In many applications, it is desirable or essential to use smaller, lighter and cheaper sensor configurations. Understanding whether these sensor configurations can provide similarly adequate performance is a topic of further study.
\item SLAM performance appeared sufficient in the dust and smoke (e.g., fog machine) obscurants tested in SubT. This is specific to the obscurants encountered; each obscurant may react differently with the lidar signals.
\item Issues with place recognition for robot wake-up were addressed procedurally, in accordance with the competition rules. Robust and reliable wake-up location in more general problems remains a topic of interest. 
\item Due to the environment scale and excellent odometry performance, place recognition was not found to be necessary for loop closure. This problem is still open for larger scale (spatial or temporal) missions, or systems with poor odometry performance.
\item The regime of sharing frames between robots and solving the SLAM problem independently on each agent was shown to be highly effective, but there are limits to its scalability. True distributed computation that accommodates larger scale but maintains the ability to address problems with poor communication remains an interesting, open problem.
\end{itemize}

\subsection{Autonomy}
One key point of realisation for our team occurred in our local Cave Circuit event in September 2020. As described in \cite{hines_2020,hudson2021heterogeneous}, the ATR robots covered extreme terrain exceptionally well. They rolled on a number of occasions, each of which was a subject of close investigation. Some were due to subsidence of the terrain under the robot, a difficult problem that was unmodelled in the terrain analysis approach. Some were identified as unexpected conditions in the behaviour stack, which were easily addressed. Most, however, occurred under teleoperation when the operator intervened to force the robot to navigate to areas where autonomy was refusing to go. The conclusion from this point was that the local navigation capability was at the point where, subject to the situation awareness constraints and latency experienced by the operator, autonomous navigation performed better on this terrain than teleoperation. 

The conclusion on global navigation was somewhat different. Although performance has steadily improved throughout the development, it remains the case that the global maps have imperfections that benefit considerably from operator input. Most difficult is the trade-off between falsely clearing frontiers in narrow doorways and failing to correctly clear frontiers when visited, leading to revisits of the same space. This is exacerbated by the fact that it is often difficult to distinguish traversable and non-traversable openings without attempting them.

A related lesson was on the scoring used for selecting frontiers. As discussed in \secref{sec:ugv_global_autonomy}, it was found that size-based scoring as is common in next best view planners, often led to undesirable behaviour such as declining to enter a small opening. In the SubT context where such traversals are of prime importance, our conclusion was that, in the absence of semantic analysis covering cases such as doors, size-based scoring was unhelpful.

Finally, the human/robot interface concept has come on an interesting journey, starting from a highly manual waypoint-based interface in the tunnel circuit, to relying on fully autonomous explore without option for human input in the urban circuit, and finally arriving at a system which permits directed autonomy, with a complete set of tools for operator prioritisation at the Final Event. The emphasis on the human/robot team in this concept significantly contributed to our result.

\subsection{Perception}
The object detection capability had interesting lessons related to the generalisation error imposed by the competition structure. As much as we could collect test data in as wide a range of environments as possible, the unique and unpredictable nature of the environments presented in the challenge events inevitably led to significant model mismatch. Consequently, false detections were often a challenge, for which the only effective mitigation was temporal analysis (i.e., object tracking).

Another approach attempted but not deployed was to improve the operator's overall situational awareness via the use of a persistent coloured point cloud. The goal was to augment the 3D structure information computed by Wildcat from lidar data with colour information obtained from each agent's onboard cameras. The displayed point cloud would dynamically update as the agents explored the unknown environment and would remain visible for the entire duration of the mission. It was hoped that the persistence would allow the operator to virtually teleport to any point in the explored environment in order to look for artefacts and/or make more informed decisions on each agent's progress or current task \citep{Vechersky_2018}.

The feature was implemented to run in real-time on the agents with a resolution of one point per 30\,mm$^3$ voxel, and a rate of 4\,Hz per camera, limiting bandwidth by compressing and sending only new points to the base station for visualisation. The work required to reconstruct the point cloud on the base station was unable to be completed in time for the Final Event due to competing priorities. %

\begin{figure}[t]
  \centering
  \includegraphics[width=120mm]{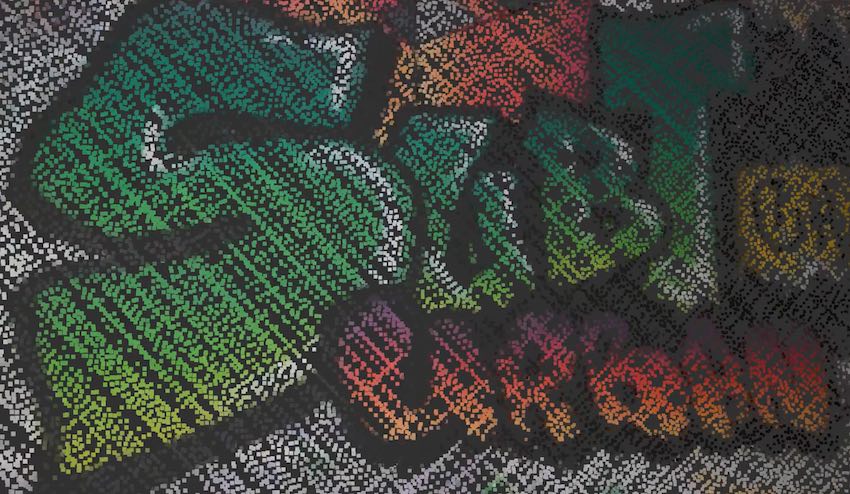}
  \caption{The raw coloured point cloud showing a mural in the circuit, calculated online by Team CSIRO Data61's Bluey platform during the Final Event.}
  \label{fig:coloured_point_cloud_subt_urban}
\end{figure}

A complete study of this functionality (including bandwidth impacts) will be pursued in the future; an example of the coloured point cloud is shown in \figref{fig:coloured_point_cloud_subt_urban}.

\begin{figure}[!b]
\centering
\subfloat[]{\includegraphics[height=3.5cm]{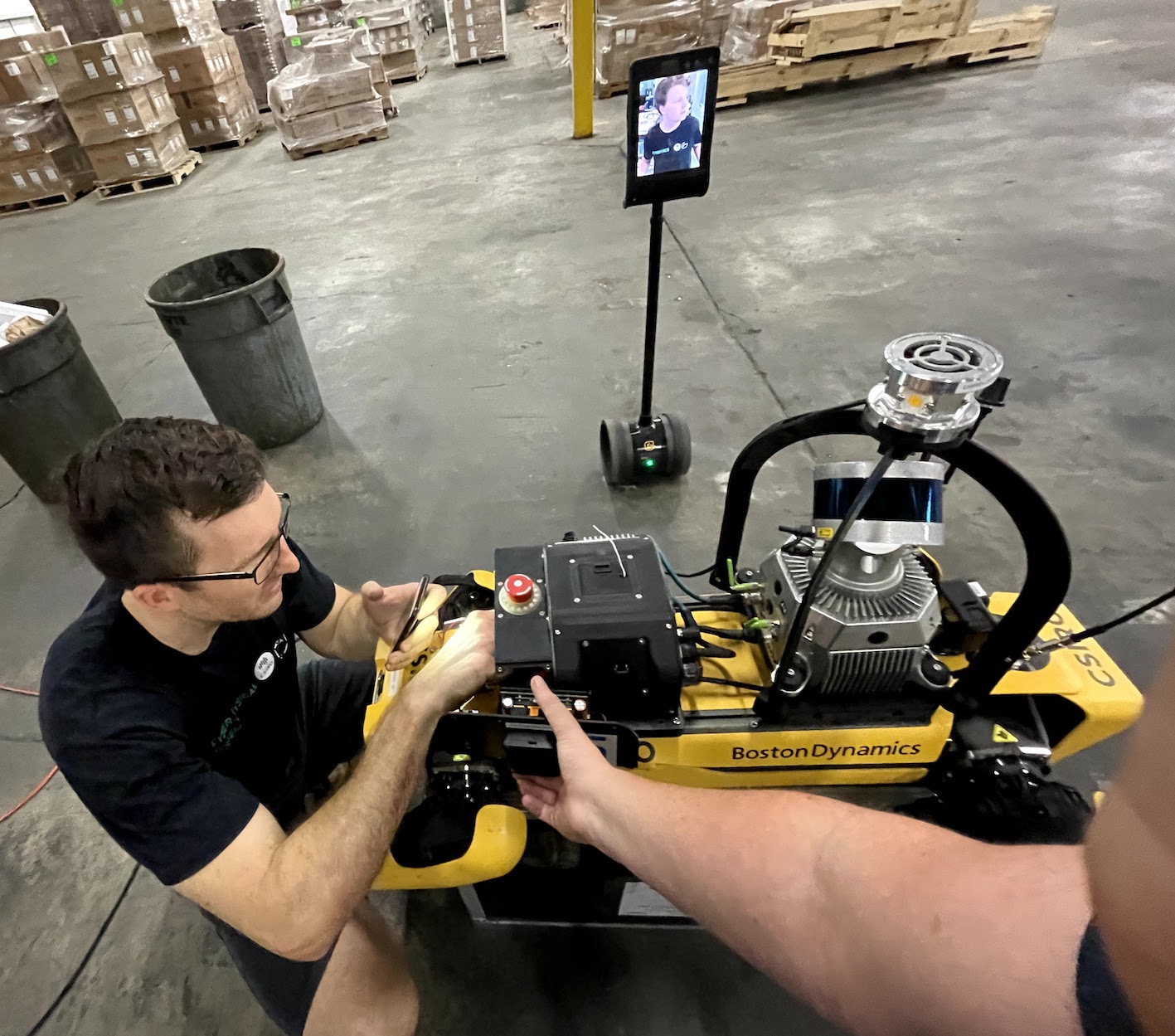}}~
\subfloat[]{\includegraphics[height=3.5cm]{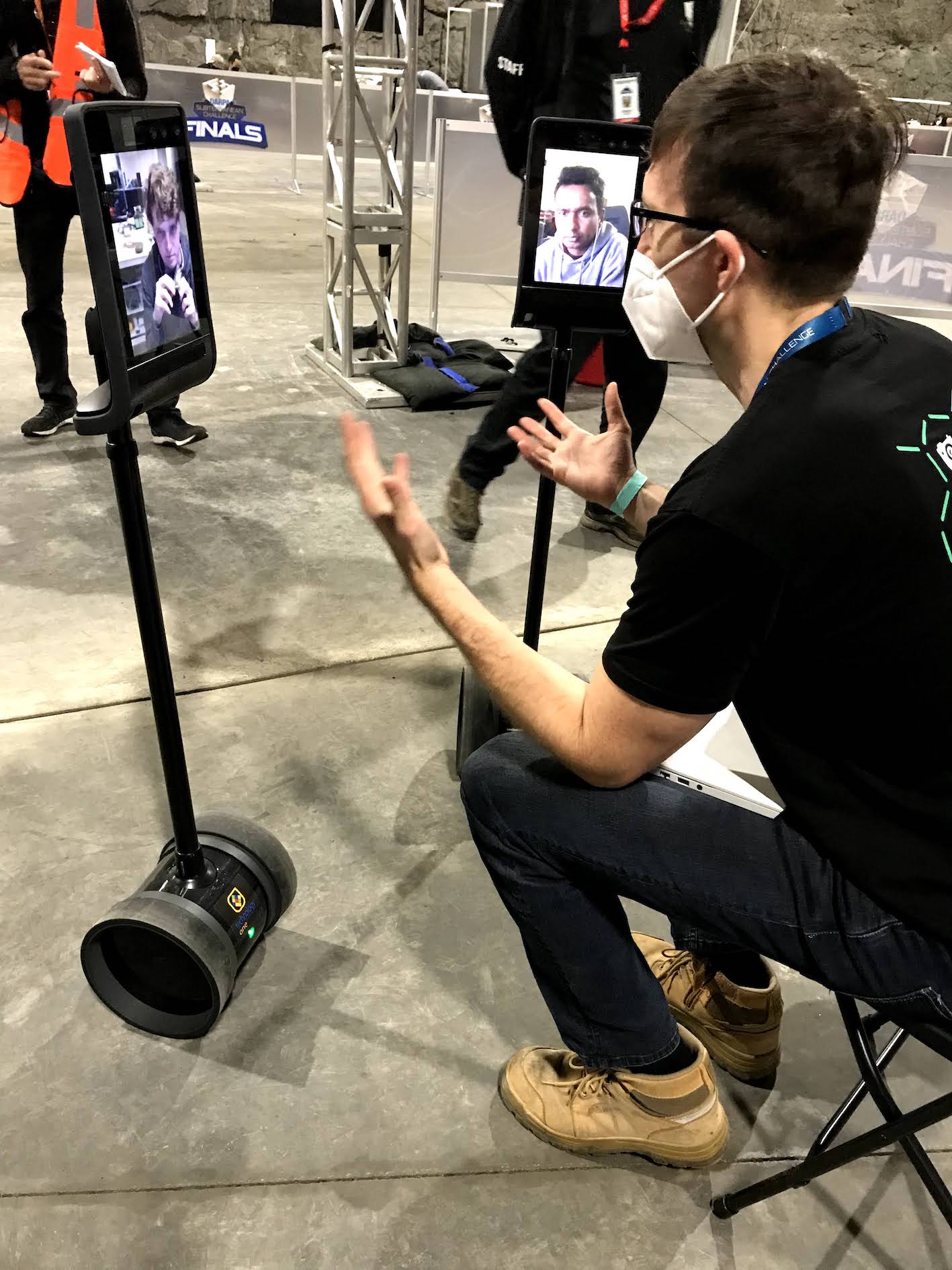}}~
\subfloat[]{\includegraphics[height=3.5cm]{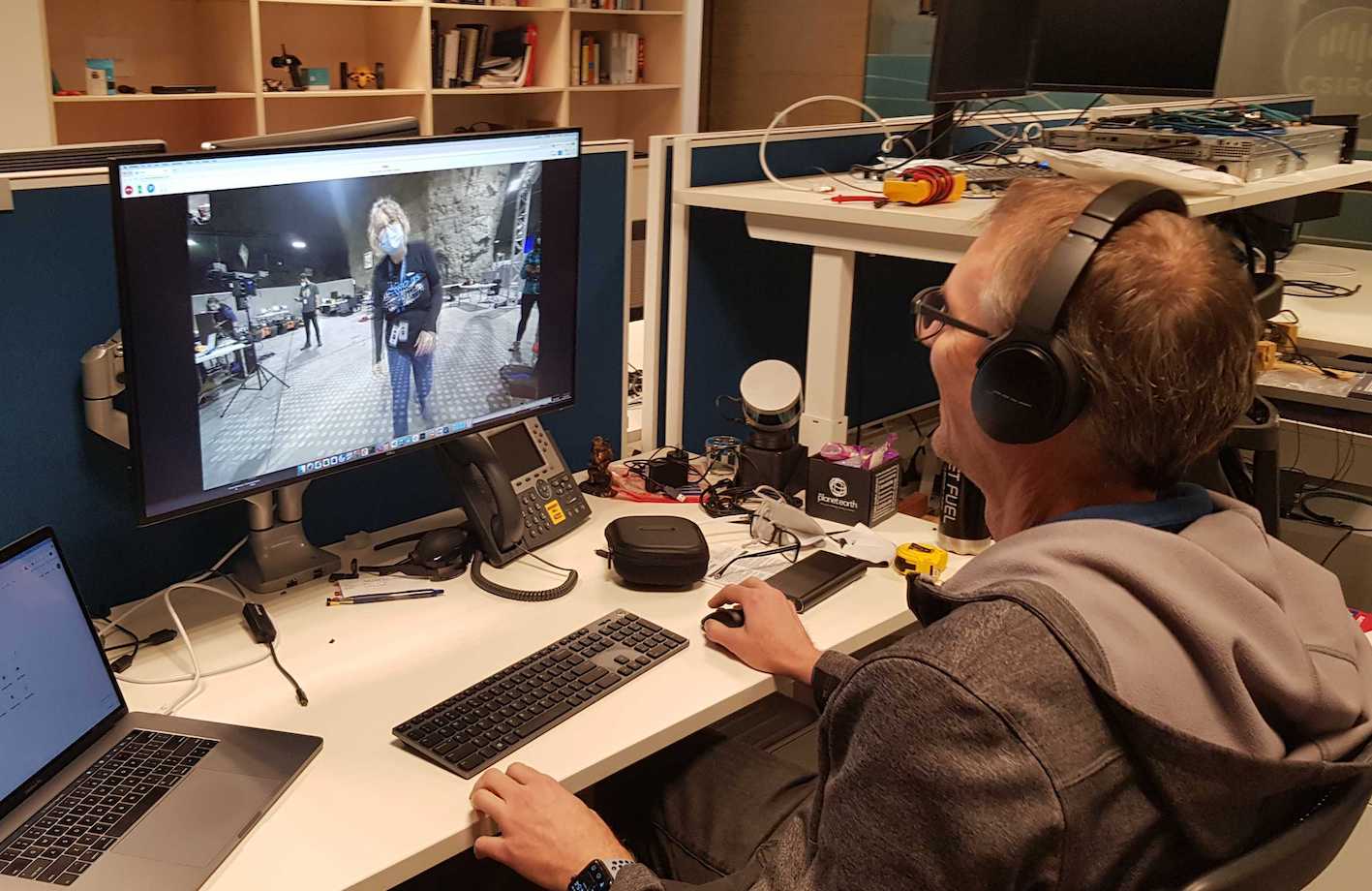}}

\caption{An engineer from the Australia based team providing remote advice via teleprecence robot to a US based team member to perform emergency electrical repairs on the Boston Dynamics Spot Quadruped's Autonomy payload (a), A US based team member providing run debrief to the Australia based team after a preliminary run (b) and an engineer from the Australia based team working US East coast hours from Brisbane to provide support during the Final Event (c).}
\label{fig:Remote_support}
\end{figure}

\subsection{Remote Support}
Finally, due to COVID-19 induced limitations on the composition of our deployed team for the Final Event led to a very challenging experience, where much of the development team provided remote support from the opposite side of the world. The Australian development team switched to the US East coast time zone for the period of the deployment (i.e., both the lead-up and the actual event), and provided support through telepresence robots and video conferencing (\figref{fig:Remote_support}). Sending robot recordings back to Australia was challenging due to the lack of high-bandwidth connections in the deployed location, and was generally only achievable after stripping out all but the most critical data from the log files.

Hardware work performed by the deployed team included replacing an ATR motor, fixing wiring faults with an autonomy computer on a Spot robot, and replacing broken drop node compute modules. In each case, with the help of duplicate hardware held back in Australia, the development team performed the procedure, capturing detailed step-by-step instructions and photographs, which were sent to the deployed team who executed the procedure under remote supervision.

It was also necessary to be realistic about the robot team composition that could be supported. For example, a decision was made to not ship the DTR as it was not considered feasible for the small deployed team to support a fourth platform type.

\section{Conclusions}
\label{sec:conclusion}
We have presented the system Team CSIRO Data61 deployed at the DARPA SubT Challenge finals. Special emphasis was given to the improvements and changes made to our systems and approach since Phase I (Tunnel Circuit) and Phase II (Urban and Cave Circuits) of the challenge. The reasons for these changes were also explained. Results from the Final Event were presented and analysed. We also provided insights and lessons learned over the overall campaign. The paradigm of using the same sensing and autonomy payloads on different robot platforms allowed us to effectively scale our fleet. This also allowed us to pivot to new platform types with minimum lead time as demonstrated by us fully integrating the Spot platform in to our fleet just a few months before the Final Event. In the final prize run, we had all but one of our robots being immobilised due to various challenge elements in the course leading up to the final minutes of the run. Despite this attrition of agents, the overall system performed as designed to provide resilience against this and we managed to create the most accurate maps of the environment as well as tie for the top score. Therefore, we have demonstrated how our paradigm of heterogeneous robot teams with unified perception and autonomy allowed Team CSIRO Data61 to achieve a remarkable outcome at the SubT Challenge finals, even without being able to send the full development team from Australia to the event in the US.

\subsubsection*{Acknowledgments}
The authors would like to thank Nicolas Hudson, Erin McColl, Nicola Cowen, William Docherty, Megan Croker, Julie Noblitt, Dawn Lillington, Margaret Donoghue and Alison Donnellan for their support leading up to and during the Final Event of the SubT Challenge. The authors also thank Tim Chung, Viktor Orekov and the DARPA team for the extraordinary efforts and support during the SubT program.

\bibliographystyle{apalike}
\bibliography{references.bib}

\end{document}